\documentclass[runningheads]{eccv_2026_template/llncs}

\usepackage{eccv_2026_template/eccv}

\usepackage{eccv_2026_template/eccvabbrv}

\usepackage{graphicx}
\usepackage{booktabs}
\usepackage[normalem]{ulem} 

\usepackage[accsupp]{axessibility}  

\usepackage[utf8]{inputenc} 
\usepackage[T1]{fontenc}
\usepackage{url}            
\usepackage{booktabs}       
\usepackage{amsfonts}       
\usepackage{nicefrac}       
\usepackage{microtype}      
\usepackage{xcolor,colortbl}         
\usepackage{amsmath}
\usepackage{amssymb}
\usepackage{bbm}
\usepackage{pifont}
\usepackage{multirow}
\usepackage{subcaption}
\usepackage{wrapfig}
\usepackage{minitoc}
\usepackage{xspace}
\usepackage{makecell}
\usepackage{capt-of}
\usepackage{sidecap}
\usepackage{array}
\usepackage{pgfplots}
\pgfplotsset{compat=1.18}
\usepgfplotslibrary{polar}
\usepackage{fontawesome5}
\usepackage{threeparttable}
\usepackage{standalone}
\usepackage{csquotes}
\usepackage{fix-cm}
\usepackage{placeins}  
\newcommand{\cmark}{\ding{51}}
\newcommand{\xmark}{\ding{55}}

\newcommand{\pretrained}{pre-trained\xspace}

\usepackage{hyperref}

\usepackage{orcidlink}
\def\modelname{HilDA\xspace}
\def\cls{\texttt{CLS}}
\def\Pretraining{Pre-training\xspace} 
\def\pretraining{pre-training\xspace} 
\def\minkunet{MinkUnet34\xspace} 

\begin{document}

\addtocontents{toc}{\protect\setcounter{tocdepth}{-10}}

\title{HilDA: \underline{Hi}erarchical Disti\underline{l}lation with \underline{D}iffusion for \underline{A}dvancing Self-Supervised LiDAR \Pretraining}

\titlerunning{HilDA: Hierarchical Distillation with Diffusion for LiDAR \Pretraining}

\author{
Maciej Wozniak$^{*,}$\inst{1}\orcidlink{0000-0002-3432-6151} \and
Jesper Ericsson$^{*}$\inst{1,3}\orcidlink{0009-0006-4836-9423} \and
Hariprasath Govindarajan\inst{2,4}\orcidlink{0000-0003-3428-6564} \and
Truls Nyberg\inst{1,3}\orcidlink{0000-0002-2069-6581} \and
Thomas Gustafsson\inst{3}\orcidlink{0009-0000-4549-1593} \and
Patric Jensfelt\inst{1}\orcidlink{0000-0002-1170-7162} \and
Olov Andersson\inst{1}\orcidlink{0000-0001-7248-1112}
}

\authorrunning{M. Wozniak et al. HilDA: Hierarchical Distillation with Diffusion}
\tocauthor{M. Wozniak et al.}

\institute{
KTH Royal Institute of Technology, Sweden\\
\email{\{maciejw, jesperik, patric, olovand\}@kth.se}
\and
Linköping University, Sweden\\
\email{hariprasath.govindarajan@liu.se}
\and
TRATON AB, Sweden\\
\email{\{truls.nyberg, thomas.gustafsson\}@scania.com}\\[0.5em]
\and
Qualcomm Auto Ltd Sweden Filial\\
$^{*}$Equal contribution. \quad
}

\let\oldaddcontentsline\addcontentsline
\let\oldaddtocontents\addtocontents

\renewcommand{\addcontentsline}[3]{}
\renewcommand{\addtocontents}[2]{}

\maketitle

\let\addcontentsline\oldaddcontentsline
\let\addtocontents\oldaddtocontents

\addtocontents{toc}{\protect\setcounter{tocdepth}{-10}}

\begin{abstract}

Leveraging Vision Foundation Models (VFMs) for camera-to-LiDAR knowledge distillation offers a promising solution to the scarcity of annotated data needed to represent the immense geometric and kinematic diversity of real-world autonomous driving (AD). However, current approaches typically treat VFMs as black-box teachers, relying exclusively on frame-wise feature similarity. Consequently, they do not fully exploit the teacher's layer-wise semantic structure and global context, as well as the rich spatiotemporal information inherent in LiDAR sequences.
We propose \textbf{\modelname}, a self-supervised \pretraining framework for LiDAR backbones that better captures the semantic \emph{what} and geometric \emph{where} needed for driving tasks. \modelname{} combines \emph{hierarchical distillation} comprising multi-layer distillation for progressive semantic alignment and global context distillation for scene-level semantics, with a \emph{temporal occupancy diffusion} objective promoting spatiotemporal consistency.
Models pre-trained with \modelname{} achieve state-of-the-art results on cross-modal distillation benchmarks and outperform models trained via prior distillation approaches on 3D object detection, scene flow, and semantic occupancy prediction. Code available at: \url{https://maxiuw.github.io/hilda}.


\end{abstract}

\section{Introduction}

Robust spatial perception in dynamic and challenging environments is essential for autonomous driving safety. While LiDAR provides high-fidelity spatial data, training 3D models typically demands massive, labeled datasets. However, the main limitation is not only the annotation cost but also the inability of manual labeling to cover the combinatorial diversity of real-world geometry and motion~\cite{fang2020augmented,meng2021towards}. To mitigate this challenge, focus has shifted towards Knowledge Distillation (KD) from Vision Foundation Models (VFMs) into self-supervised LiDAR backbones as a \pretraining step~\cite{seal,superflow,govindarajan2025cleverdistiller,lima,scalr}. By using  VFMs as a teacher, recent work has successfully equipped 3D networks with dense semantic features robust to domain shifts and adverse weather conditions~\cite{slidr,ppkt,seal}. 

\begin{figure}[t!]  
    \centering
    \includegraphics[width=0.75\linewidth, trim={11cm 22.7cm .3cm 0cm}, clip]{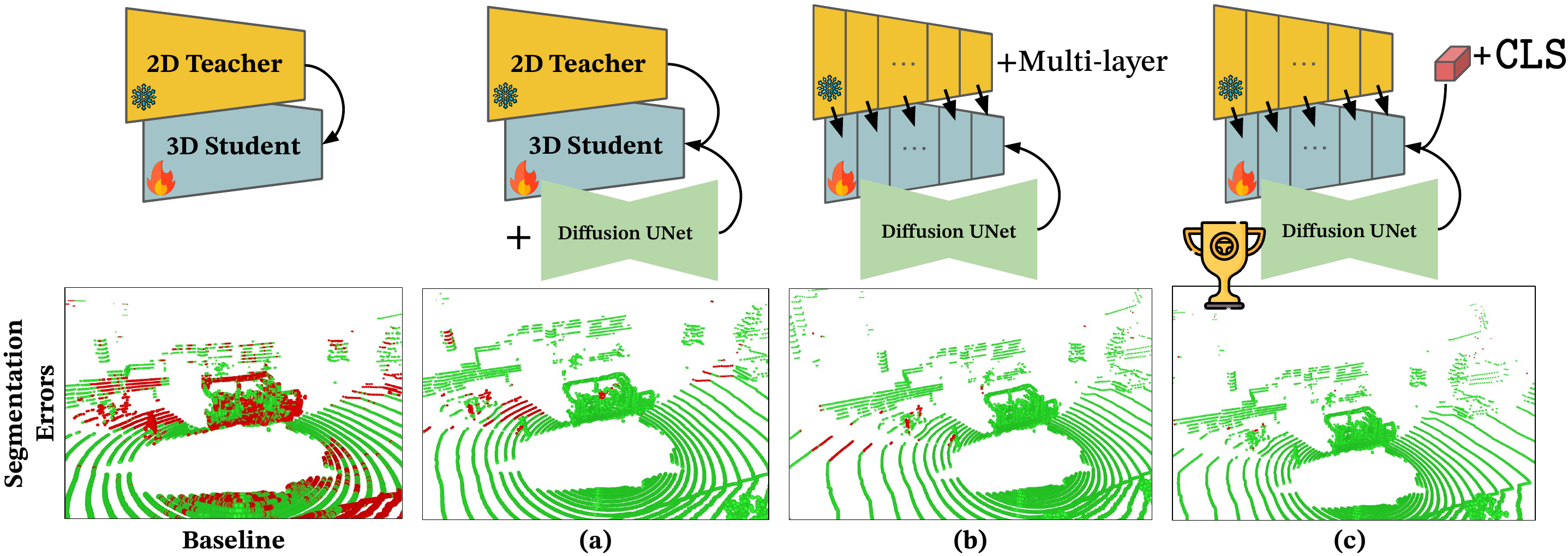}
    \caption{Effect of \modelname components on semantic accuracy. From baseline (left), we can see error reduction by adding (a) temporal occupancy diffusion, (b) multi-layer distillation, and (c) global context distillation. Best viewed zoomed in \faSearch.}
    \label{fig:first_fig}
\end{figure}

While these methods established a solid foundation for cross-modal distillation, we observe that they only distill the features from the final layer of the VFM~\cite{ppkt,slidr,seal,scalr,superflow,limoe,lima,govindarajan2025cleverdistiller}. This approach limits the ability to fully harness the rich representational knowledge that recent work suggests is distributed throughout the VFM’s layers~\cite{perception_encoder}. In vision encoders, the multi-layer progression of features across layers captures increasingly abstract and semantically rich information \cite{ahn2026adaptability}. Because of this, restricting distillation to the final layer may cause two important drawbacks. First, the progressive semantic refinement across layers is potentially lost, limiting the transfer of intermediate representations, which are often better suited for various downstream tasks \cite{el2024probing,yoon2025visual,keetha2023anyloc,perception_encoder}. Second, the final-layer approach overlooks the global context captured by the \cls{} token~\cite{vit}. Recent studies on image-based VFMs demonstrate that this token encodes scene-specific information, including the prominence of objects and their relational cues~\cite{ranzinger2024radio,heinrich2025radiov2,darcet2023vision}. Instead of distilling only the last layer features, we investigate distilling the hierarchy of how the VFM learns semantics by using both multi-layer features and the global $\cls{}$ token. We term this approach \textit{hierarchical distillation}.

Finally, while VFMs trained on images excel at semantic recognition, they inherently lack explicit information about 3D scene geometry and temporal cues, leaving these aspects underconstrained during \pretraining. This issue has been noted in prior cross-modal distillation and \pretraining works~\cite{govindarajan2025cleverdistiller,zhang2024contrastive,ljungbergh2025gasp}, which introduce an auxiliary occupancy prediction objective to improve downstream performance. However, none of these methods explores the pairing of discriminative and generative priors for spatiotemporal coherence. Recent findings in 2D image space have shown that diffusion-based denoising can serve as geometry-oriented supervision~\cite{mindthegap,he2024lotus,zhang2024telling,hudson2023soda,zheng2025diff}, encouraging models to internalize spatial manifolds and surface hierarchies through denoising. By combining the semantic \emph{what} from VFMs and the generative \emph{where} from diffusion, we aim to create a representation that is both semantically informed and structurally robust~\cite{taleoftwofeatures}, particularly in the dynamic and occluded environments typical of autonomous driving~\cite{chen2024e2edriving}. Concretely, we introduce a temporal occupancy prediction task with diffusion as an auxiliary \pretraining objective and show that this design yields representations that surpass prior methods on cross-modal distillation benchmarks and improve downstream performance on 3D tasks requiring semantic and spatiotemporal reasoning.

To address both the aspects of final-layer distillation and the lack of explicit 3D spatiotemporal supervision, we introduce \textbf{\modelname}, a self-supervised \pretraining framework that combines hierarchical distillation with temporal occupancy diffusion. Our main contributions are summarized as follows:

\begin{itemize}

    \item \textit{Hierarchical Distillation:} We introduce a novel cross-modal distillation approach that goes beyond standard final-layer distillation by capturing the semantic evolution of the VFM. This is achieved through a dual mechanism: first, a \textit{multi-layer distillation} strategy utilizes intermediate layers to teach the student LiDAR model how to progressively construct features; second, a \textit{global context distillation} aligns the VFM’s \cls{} token with a novel, learnable 3D global context token to promote holistic scene-level representations.


    \item \textit{Temporal Occupancy Diffusion}: We introduce an auxiliary self-supervised diffusion objective that casts future occupancy prediction as conditional generation, encouraging predictive geometric and motion encoding while addressing the spatiotemporal limits of pure semantic distillation.
   
\end{itemize}

\noindent
\Cref{fig:first_fig} visualizes the effect of progressively incorporating temporal occupancy diffusion and hierarchical distillation. Compared to standard final-layer distillation, adding consecutive components yields visibly reduced segmentation errors, indicating increasingly structured and semantically aligned LiDAR representations.
Extensive experiments on autonomous driving benchmarks in~\cref{sec:experiments} demonstrate that \modelname{} achieves state-of-the-art (SOTA) performance in camera-LiDAR cross-modal distillation. Whether evaluated via linear probing, few-shot transfer, or as an initialization for full supervision, our approach consistently outperforms existing methods. Furthermore, our distillation strategy produces features with enhanced multi-task transferability. Beyond 3D semantic segmentation, our model outperforms previous SOTA distillation approaches in 3D object detection, scene flow, and semantic occupancy prediction.

\section{Related Work}
\label{sec:related_work}

\paragraph{\textbf{Cross-Modal Distillation Methods.}}

In VFMs, the pre-training objective strongly shapes the learned representations. For example, global alignment enhances aggregated feature semantics~\cite{clip, byol, dino}, while dense prediction localizes features and promotes boundary adherence~\cite{sam, ibot}. When transferring such features, early camera--LiDAR distillation methods largely relied on contrastive objectives, which can be sensitive to noisy point--pixel correspondences, as highlighted by~\cite{scalr,govindarajan2025cleverdistiller}. Several methods~\cite{slidr,seal,olivine,superflow} therefore used grouping or semantic priors (\eg, SAM~\cite{sam} or OpenSeed~\cite{openseed} pseudo-masks). While helpful, these priors add overhead, and contrastive distillation can still exhibit \enquote{self-conflict} by sampling semantically similar regions as negatives~\cite{slidr,seal,olivine,superflow,scalr}. Recent work instead favors direct feature alignment~\cite{govindarajan2025cleverdistiller,scalr,lima,limoe}, using \eg, expressive 2D-to-3D transfer~\cite{govindarajan2025cleverdistiller} or scaling teacher/student models and data~\cite{scalr}. 

Because VFM layers differ in semantic abstraction~\cite{perception_encoder,touvron2021training,yang2022focal}, the progressively learned feature hierarchy is distributed across intermediate layers rather than confined to the final output~\cite{tian2025distillation,yang2024vitkd}. This hierarchy contains signals about \enquote{how} to learn the features. Yet both contrastive and direct-alignment pipelines supervise only the final-layer~\cite{ppkt,slidr,seal,scalr,superflow,limoe,lima,govindarajan2025cleverdistiller} and discard the teacher \texttt{CLS} token~\cite{vit}, limiting global alignment despite its shown benefit in the 2D setting~\cite{ranzinger2024radio,heinrich2025radiov2}. Rather than relying on computationally heavy grouping priors or data scaling, we leverage the VFM’s hierarchical progression via multi-layer distillation and global \texttt{CLS} token distillation to strengthen cross-modal alignment.

\paragraph{\textbf{Geometric and Temporal Auxiliaries.}}

Image-based VFMs transfer semantics but typically lack explicit 3D geometric or temporal supervision. Many camera–LiDAR distillation methods prioritize cross-modal feature alignment~\cite{limoe}, or semantic-temporal consistency objectives~\cite{superflow,lima}. Recent approaches \cite{govindarajan2025cleverdistiller,zhang2024contrastive} and representation learning works~\cite{agro2024uno,diehl2025dio,ljungbergh2025gasp} increasingly use occupancy prediction to impose spatiotemporal priors. While effective, discriminative occupancy prediction trained with per-query/voxel binary cross-entropy supervision models local marginals rather than the joint scene distribution. Shared receptive fields may implicitly promote consistency, but this objective provides no explicit incentive for globally coherent configurations, limiting its ability to capture complex 3D structure~\cite{Wang25DiffOcc}. In practice, occupancy is correlated across space and time, forming continuous structures~\cite{wei2023surroundocc} and maintaining object permanence~\cite{ma2024cam4docc}. This motivates the use of generative objectives that can model occupancy jointly and impose scene-level spatiotemporal structure through iterative refinement~\cite{gu2024dome}.

Diffusion models~\cite{ddpm,rombach2022high}, analogous to denoising autoencoders trained across noise levels~\cite{yang2023repfusion,ddae_iccv2023,deconstructing_ddm_ssl2025}, define a coarse-to-fine refinement process over dense structured targets~\cite{wang2024occgen}. Applied to spatiotemporal denoising, they can learn to represent stochastic future evolution by encoding global scene dynamics~\cite{voleti2022MCVD}. In autonomous driving, diffusion is used for spatiotemporal prediction and refinement, including LiDAR completion~\cite{nunes2025DBLP}, world-model rollouts~\cite{zhang2024bevworld,zhang2025epona,wang2023drivedreamer}, BEV/scene-flow refinement~\cite{ye2025bevdiffuser, liu2024difflow3d}, and as action decoders~\cite{wang2025alpamayo, liao2025diffusiondrive, fu2025orion}. While the diffusion objective alone can overemphasize local detail and weaken semantic discrimination, pairing it with robust discriminative teachers (\eg, DINOv2~\cite{dinov2}) is shown to preserve semantics while improving geometric priors~\cite{taleoftwofeatures, li2025cuao3d}. Motivated by this complementarity, we use diffusion as a \pretraining objective~\cite{hudson2023soda}, denoising future occupancy from past context across noise levels to inject multi-scale geometric and motion cues into the LiDAR encoder.

\section{Method}
\label{sec:method}

\begin{figure}[t]
    \centering
    \includegraphics[width=1.00\linewidth, trim={0 0 0 0}, clip]
    {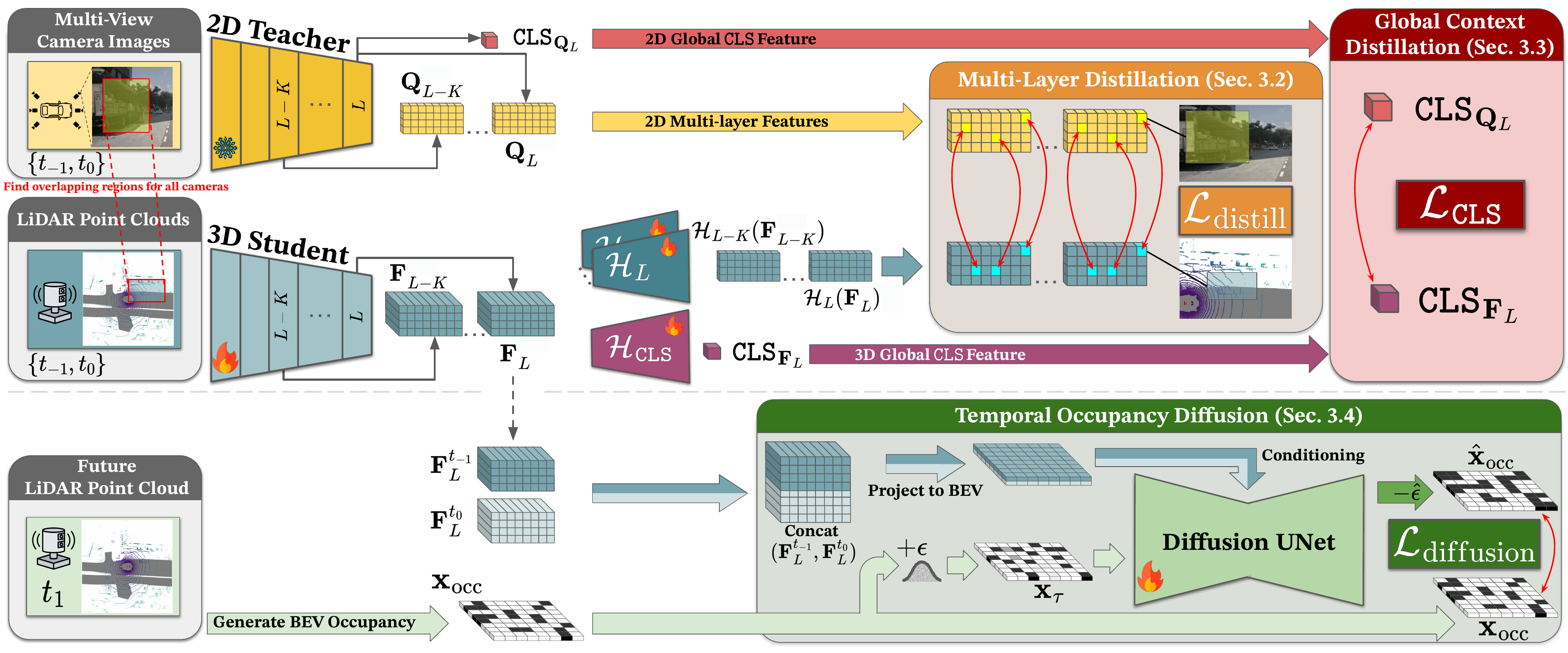}
    \caption{\textbf{Overview of \modelname{}.} With LiDAR sweeps and synchronized multi-view images, \modelname{} learns a 3D backbone using three self-supervised objectives: \textbf{(1)} multi-layer point--pixel distillation from a frozen VFM, \textbf{(2)} global context distillation via \cls{} tokens ($t_{-1}$ and $t_0$ are processed independently through the distillation modules), and \textbf{(3)} temporal occupancy diffusion conditioned on the student features $(\mathbf{F}_L^{t_{-1}}, \mathbf{F}_L^{t_0})$. Features are extracted from $t_{-1}$ and $t_0$, while $t_1$ constructs the future BEV occupancy target.}

    \label{fig:architecture}
\end{figure}

In this section, we define the \pretraining formulation and present the components comprising our framework \textbf{\modelname{}}. \Cref{fig:architecture} shows an architectural overview.

\subsection{Problem Definition and Motivation}

The primary objective of cross-modal knowledge distillation is to pre-train a LiDAR student network by transferring rich and dense semantic knowledge from a VFM while also learning geometric features exclusive to the LiDAR modality. For this task, let the input be a point cloud $\mathcal{P} = \{ \mathbf{p}_i \mid i = 1, \ldots, N \}$ of $N$ points, where each point $\mathbf{p}_i \in \mathbb{R}^{4}$ denotes the 3D spatial coordinates $(x, y, z)$ and intensity. Let $V$ be the number of synchronized cameras forming a surround view $\mathcal{I} = \{ \mathbf{I}_c \mid c = 1, \ldots, V \}$, where $\mathbf{I}_c \in \mathbb{R}^{H \times W \times 3}$ is the RGB-image from camera $c$.

The challenge lies in the fundamental discrepancy between modalities: LiDAR data is geometrically precise but semantically sparse and texture-less, whereas camera images are semantically dense but lack explicit depth or metric 3D structure. We aim to pre-train a LiDAR backbone $S_{\theta}(\cdot)$ in a self-supervised manner to inherit visual semantics without requiring manual annotations, while simultaneously learning spatiotemporal features through a generative auxiliary task. The outcome of our self-supervised \pretraining is a backbone $S_{\theta}$ that can produce robust feature representations useful for various perception tasks (note, for inference we remove all other components used in \pretraining). \modelname{} processes temporal sequences of three frames $\{t_{-1}, t_0, t_{1}\}$. Feature extraction is restricted to the past and present frames, $\{t_{-1}, t_0\}$, while the future frame $t_{1}$ serves exclusively as the target for occupancy prediction. We omit the temporal index in the two following~\cref{subsec:distillation,sec:method_cls} for notational simplicity.

\subsection{Multi-Layer Distillation}
\label{subsec:distillation}

To effectively transfer semantic knowledge, we employ a dense-to-sparse distillation strategy. This process relies strictly on geometric calibration, avoiding the need for semantic priors.  We find corresponding points and pixels by projecting the point coordinates $(x_i, y_i, z_i)$ onto the image plane $(u_i, v_i)_c$ of camera $c$ using calibration information. The following sensor calibration parameters are used: $\begin{bmatrix} u_i & v_i & 1 \end{bmatrix}^T = \rho(i) = \frac{1}{z_i} \times \Gamma_I \times \Gamma_{c \leftarrow \text{LiDAR}} \times \begin{bmatrix} x_i & y_i & z_i \end{bmatrix}^T$, where $\Gamma_I$ denotes the camera-intrinsic matrix and $\Gamma_{c \leftarrow \text{LiDAR}}$ is the transformation matrix from the LiDAR sensor to each camera. This enables a mapping between points and pixels for each camera image. Let $\mathcal{M}_c$ be the set of valid point--pixel correspondences for image $\mathbf{I}_c$. We define a pair $(i,j) \in \mathcal{M}_c$ if the projection of the 3D point $\mathbf{p}_i$ onto camera $c$ maps to the pixel with index $j$. Points outside $\mathcal{M}_c$ are excluded from multi-layer distillation but still contribute to the losses in~\cref{sec:method_cls,sec:diffusion}.

We distill features from multiple teacher layers into the corresponding student layers to capture not only the teacher’s final representation, but also how features form across the VFM hierarchy. By default, we use the last two layers. Let $L$ be the index of the final output layer for both teacher and student\footnote{\scriptsize \mbox{In the case of \minkunet}~\cite{choy20194minkunet}, it consists of \enquote{planes} that have multiple layers. We refer to these planes as \enquote{student layers}. For the teacher, we select adjacent ViT transformer blocks and refer to these as \enquote{teacher layers}. Thus, each selected feature layer from the student and teacher includes a non-trivial stage/block-level transformation rather than a single atomic layer.} (the two networks may have different total depths). We define the layer index $\ell\in\{L-K,\ldots,L\}$. We denote the student point-feature tensor at layer $\ell$ by $\mathbf{F}_{\ell}\in\mathbb{R}^{N\times C_{\ell}}$, and the feature of point $i$ by $\mathbf{f}_{i,\ell}=\mathbf{F}_{\ell}[i]\in\mathbb{R}^{C_{\ell}}$. Let $\mathbf{Q}_{\ell,c} \in \mathbb{R}^{H\times W \times D_{\ell}}$ be the feature map at the corresponding teacher layer for camera $c$, bi-linearly up-sampled to the image resolution $H \times W$. From this, we can obtain the teacher feature at pixel $j$ as $\mathbf{q}_{j,\ell,c}=\mathbf{Q}_{\ell,c}[j]\in\mathbb{R}^{D_{\ell}}$. We apply a lightweight, layer-specific MLP $\mathcal{H}_{\ell}:\mathbb{R}^{C_{\ell}}\!\to\!\mathbb{R}^{D_{\ell}}$ to align the teacher and student feature dimensions (see~\cref{fig:architecture}). Cosine distance is then minimized over point-pixel pairs across layers and images:
\begin{equation}
\mathcal{L}_{\mathrm{distill}} = \frac{1}{V(K+1)} \sum_{\ell=L-K}^L \sum_{c=1}^{V} \frac{1}{|\mathcal{M}_{c}|}\sum_{(i,j) \in \mathcal{M}_c} \left( 1 - \frac{\mathcal{H}_{\ell}(\mathbf{f}_{i,\ell}) \cdot \mathbf{q}_{j,\ell,c}}{\|\mathcal{H}_{\ell}(\mathbf{f}_{i,\ell})\|_2 \, \|\mathbf{q}_{j,\ell,c}\|_2} \right).
\label{eq:distill}
\end{equation}
By maximizing similarity across layers, the 3D student learns to progressively encode visual concepts using an information hierarchy similar to the teacher.  

\subsection{Global Context Distillation}
\label{sec:method_cls}
Local point-pixel distillation ensures fine-grained alignment but may miss the holistic context of a scene (\eg, distinguishing a highway environment from a residential area). ViT-based VFMs inherently encode global context in their \cls{} token. Thus, as part of our novel hierarchical distillation strategy, we complement the local multi-layer distillation loss in~\cref{eq:distill} with global context distillation.

To construct a unified visual scene descriptor $\cls{}_{\mathbf{Q}_L}$, we extract the \cls{} tokens from the teacher's final layer $L$ for all $V$ camera images and apply a max-pooling across them. Correspondingly, for the LiDAR scene $\mathcal{P}$, we design a global context \enquote{token} $\cls{}_{\mathbf{F}_L}$ by applying a dedicated MLP projection head $\mathcal{H}_\cls{}(\cdot)$ to the features from the final layer $L$ of the student network, followed by global max-pooling over all point features in the scene. Max pooling highlights the most active features. We then align the global representation of the student with the teacher by minimizing the Mean Squared Error (MSE):
\begin{equation}
    \mathcal{L}_{\cls{}} = \|\cls{}_{\mathbf{F}_L} - \cls{}_{\mathbf{Q}_L}\|_2^2.
\label{eq:cls}
\end{equation}

This objective acts as a scene-level regularizer, encouraging the 3D backbone to aggregate global context by matching the student’s pooled representation to the VFM’s \cls{} embedding. This emphasizes salient view- and point-level activations to align global semantic cues between the student and the teacher.~\Cref{sec:additional_results,tab:ablation_design_components_compact} ablates different loss formulations for~\cref{eq:cls} as well as different pooling methods for constructing $\cls{}_{\mathbf{Q}_L}$ and $\cls{}_{\mathbf{F}_L}$.

\subsection{Spatiotemporal Self-Supervised Learning as an Auxiliary Task}
\label{sec:diffusion}

Distillation provides semantic context but does not explicitly model spatiotemporal scene dynamics. We therefore add a label-free auxiliary task that predicts future BEV occupancy with a conditional diffusion model, encouraging the 3D student $S_\theta$ to learn predictive spatial representations~\cite{hudson2023soda,diffae2022}. Given two LiDAR sweeps $\mathcal{P}^{t_{-1}}$ and $\mathcal{P}^{t_0}$, $S_\theta$ encodes them into features $\mathbf{F}^{t_{-1}}_L$ and $\mathbf{F}^{t_0}_L$, which are collapsed into a shared dense BEV history feature map $\mathbf{C}_{\mathrm{history}}$. The future sweep $\mathcal{P}^{t_1}$ is transformed into the $t_0$ frame and projected to a ground-removed BEV occupancy target $\mathbf{x}_{\mathrm{occ}}\in\{0,1\}^{H_{\mathrm{BEV}}\times W_{\mathrm{BEV}}}$.  We formulate future occupancy prediction as conditional DDPM denoising~\cite{ddpm}. At diffusion step $\tau$, the target occupancy is corrupted as $\mathbf{x}_\tau = \sqrt{\bar{\alpha}_\tau}\mathbf{x}_{\mathrm{occ}} + \sqrt{1-\bar{\alpha}_\tau}\epsilon$, with $\epsilon\sim\mathcal{N}(0,\mathbf{I})$ and $\bar{\alpha}_\tau=\prod_{s=1}^{\tau}(1-\beta_s)$. A denoising network (2D UNet~\cite{ronneberger2015unet}) then predicts $\epsilon_\theta(\mathbf{x}_\tau,\tau,\mathbf{C}_{\mathrm{history}})$, as shown in~\cref{fig:architecture}. Conditioning is implemented via channel-wise concatenation of $\mathbf{x}_\tau$ and $\mathbf{C}_{\mathrm{history}}$, following~\cite{Li_2025_CVPR}. We train it with the hybrid objective 
\begin{equation} 
\mathcal{L}_{\mathrm{diffusion}} = \underbrace{ \mathbb{E}_{\tau,\epsilon,\mathbf{C}_{\mathrm{history}}} \Big[ \big\| \epsilon - \epsilon_\theta(\mathbf{x}_\tau, \tau, \mathbf{C}_{\mathrm{history}}) \big\|_2^2 \Big] }_{\text{Noise Prediction}} \;+\; \lambda \underbrace{ \big\| \mathbf{x}_{\mathrm{occ}} - \hat{\mathbf{x}}_{\mathrm{occ}} \big\|_2^2 }_{\text{Occ. Reconstr.}} \label{eq:diffusion_loss} \end{equation} 
where the first term supervises noise prediction, while the second, weighted by $\lambda$, provides a complementary reconstruction signal~\cite{nichol2021improved} which improves conditioning at low $\tau$ (see~\cref{sec:add_ablations} for ablation). The reconstructed occupancy $\hat{\mathbf{x}}_{\mathrm{occ}}$ is recovered from the noisy sample by inverting the forward process using the predicted noise. Training across varying $\tau$ exposes the model to denoising tasks at different noise levels, encouraging a coarse-to-fine behavior from global structure to local detail. Full implementation details and target construction are provided in~\cref{sec:implementationdetails}.

\subsection{Final Pre-Training Objective}
\modelname{} is trained end-to-end on three objectives, encouraging the 3D student to learn detailed semantics via distillation, while jointly learning scene geometry and dynamics via diffusion-based temporal occupancy prediction:
\begin{equation}
    \mathcal{L}_{\mathrm{total}} = \omega_{ds}\mathcal{L}_{\mathrm{distill}} + \omega_{gl}\mathcal{L}_{\cls{}} + \omega_{df}\mathcal{L}_{\mathrm{diffusion}}. 
\label{eq:loss_total}
\end{equation}

\section{Experiments}
\label{sec:experiments}
\subsection{Experimental Setup}
We distill from DINOv2~\cite{dinov2} to \mbox{\minkunet}~\cite{choy20194minkunet}, the standard backbone in prior camera--LiDAR distillation work. Our method also supports other 3D backbones, \eg, PTv3~\cite{ptv3} (supplementary~\cref{sec:additional_results}). Each teacher--student configuration is \pretrained \textit{once} on the nuScenes training split~\cite{caesar2020nuscenes} using synchronized RGB--LiDAR data and calibration only; no task labels are used. The resulting backbone is transferred to \textit{all} downstream benchmarks \textit{without} target-dataset re-\pretraining. \Cref{tab:minkunet_main_table} varies the DINOv2 teacher size, while~\cref{sec:multi-task} uses the same ViT-B/\minkunet checkpoint for all transfer tasks. We evaluate segmentation on nuScenes~\cite{caesar2020nuscenes}, SemanticKITTI~\cite{behley2019semantickitti}, Waymo~\cite{sun2020waymo}, ScribbleKITTI~\cite{unal2022scribble}, RELLIS-3D~\cite{jiang2021rellis}, SemanticSTF~\cite{xiao20233d}, and DAPS-3D~\cite{klokov2023daps3d}; robustness on nuScenes-C~\cite{kong2023robo3d}; 3D detection on KITTI~\cite{kitti} and nuScenes; semantic occupancy on nuScenes; and scene flow on Argoverse~2~\cite{Argoverse2}. Linear probing (LP) freezes the \pretrained backbone and trains only a linear head, whereas data-scarce/full fine-tuning updates the full network using the indicated label fraction. For semantic occupancy, we freeze the backbone and train a lightweight decoder. Metrics are mIoU for segmentation/occupancy, mAP for detection, EPE-based metrics for scene flow, and mCE/mRR for robustness. We use author-reported results unless stated otherwise and provide details in supplementary~\cref{sec:implementationdetails}. \modelname{}$^\dagger$ denotes \modelname{} without the diffusion auxiliary \pretraining loss; it is excluded from \textbf{best}/\underline{$2^{nd}$-best} ranking to emphasize the comparison of \modelname{} against prior SOTA.

\subsection{Cross-Modal Distillation Benchmarks}
\label{sec:main-cross-modal-results}

In the next three tables, we compare our method with previous SOTA approaches on standard cross-modal distillation benchmarks for 3D semantic segmentation. 

\begin{table*}[t!]
\centering
\small
\caption{\small 3D semantic segmentation results for cross-modal distillation methods \pretrained on nuScenes~\cite{caesar2020nuscenes}, and fine-tuned on nuScenes, SemanticKITTI~\cite{behley2019semantickitti}, and Waymo~\cite{sun2020waymo}. All methods use the \minkunet~\cite{choy20194minkunet} backbone. LP denotes linear probing with a frozen backbone. \enquote{S.P.} refers to usage of semantic priors.
}
\label{tab:minkunet_main_table}
\renewcommand{\arraystretch}{0.7} 
\setlength{\tabcolsep}{4.5pt}
\resizebox{\linewidth}{!}{
\begin{tabular}{lccccccccccc}

\toprule
\multirow{2}{*}{{\textbf{Method}}} &  \multirow{2}{*}{{\textbf{Venue}}} &  \multirow{2}{*}{{\textbf{Teacher}}} &  \multirow{2}{*}{{\textbf{\begin{tabular}[c]{@{}c@{}}S.P.\end{tabular}}}} & \multicolumn{6}{c}{\textbf{nuScenes}} &  {\textbf{SKITTI}} & \multicolumn{1}{c}{\textbf{Waymo}} \\

\cmidrule(lr){5-10} \cmidrule(lr){11-11} \cmidrule(lr){12-12} 

 {} &  {} &  {} &  {} & \textbf{LP} & \textbf{1\%} & \textbf{5\%} & \textbf{10\%} & \textbf{25\%} &  \textbf{Full}  &  \textbf{1\%} & \textbf{1\%} \\ 

\hline

Random & - & - & {\textcolor{red}{\xmark}} & 8.10 & 30.30 & 47.84 & 56.15 & 65.48 & 74.66 & 39.50 & 39.41 \\ 

\arrayrulecolor{black} \midrule \midrule \arrayrulecolor{lightgray}

 {SLidR~\cite{slidr}} &  {CVPR’22} &  {{\color[HTML]{2FC061} ViT-S}} &  {\textcolor{green}{\cmark}} &  44.70 &  41.16 &  53.65 &  61.47 &  66.71 &  74.20 &  44.67 & 47.57 \\ 

\hline

 {Seal~\cite{seal}} &  {NIPS’23} &  {{\color[HTML]{2FC061} ViT-S}} &  {\textcolor{green}{\cmark}} &  45.16 &  44.27   &  55.13 &  62.46 &  67.64 &  75.58 &  46.51 & 48.67 \\ 

\hline

 {SuperFlow~\cite{superflow}} &  {ECCV'24} &  {{\color[HTML]{2FC061} ViT-S}} &  {\textcolor{green}{\cmark}} &  46.44 &  47.81 &  59.44 &  64.47 &  69.20 &  76.54 &  47.97 & 49.94 \\ 

\hline

 {LiMoE~\cite{xu2025limoe}} &  {CVPR'25} &  {\color[HTML]{2FC061}ViT-S} &  {\textcolor{green}{\cmark}} &  48.20 &  49.60 &  60.54 &  65.65 &  \cellcolor{white!80!SpringGreen}\underline{71.39} &  77.27 &  49.53 & \cellcolor{white!80!SpringGreen}\underline{51.42} \\ \hline

CleverDistiller~\cite{govindarajan2025cleverdistiller} &  {BMVC'25} &  {{\color[HTML]{2FC061} ViT-S}} &  {\textcolor{red}{\xmark}} &  49.81 &  \cellcolor{white!80!SpringGreen}\underline{56.90} & \cellcolor{white!80!SpringGreen}\underline{64.55} &  \cellcolor{white!80!SpringGreen}\underline{65.92} &  70.11 &  \cellcolor{white!80!SpringGreen}\underline{77.61} &  \cellcolor{white!80!SpringGreen}\underline{50.59} & 50.99 \\  \hline

 {SuperFlow++~\cite{superflow++}} &  {TPAMI'25} &  {{\color[HTML]{2FC061} ViT-S}} &  {\textcolor{green}{\cmark}} &  48.57 &  49.07 &  60.57 &  65.21 &  70.05 &  76.92 &  49.27 & 51.25 \\ 

\hline

LiMA~\cite{lima}& ICCV'25 &  {{\color[HTML]{2FC061} ViT-S}} & {\textcolor{red}{\xmark}} & \cellcolor{white!80!SpringGreen}\underline{54.76} & 48.75 & 60.83 & 65.41 & 69.31 & 76.94 & 49.28&  50.23 \\ \hline

\textbf{\modelname$^\dagger$} & - & {{\color[HTML]{2FC061} ViT-S}}   & {\textcolor{red}{\xmark}}  & 55.13 & 57.81 & 66.26 & 67.32 & 70.54   & 77.17 & 50.80  &  50.87\\ 

\textbf{\modelname} & - & {{\color[HTML]{2FC061} ViT-S}}   & {\textcolor{red}{\xmark}}  & \cellcolor{white!40!SpringGreen}\textbf{56.29}  & \cellcolor{white!40!SpringGreen}\textbf{59.46} & \cellcolor{white!40!SpringGreen}\textbf{68.57} & \cellcolor{white!40!SpringGreen}\textbf{70.22} & \cellcolor{white!40!SpringGreen}\textbf{73.94}  & \cellcolor{white!40!SpringGreen}\textbf{78.15} & \cellcolor{white!40!SpringGreen}\textbf{52.88}  & \cellcolor{white!40!SpringGreen}\textbf{52.16}  \\

\arrayrulecolor{black} 
\midrule
\midrule 
\arrayrulecolor{lightgray}


 {SLidR~\cite{slidr}} &  {CVPR’22} &  {{\color{Apricot} ViT-B}} &  {\textcolor{green}{\cmark}} &  45.35 &  41.64 &  55.83 &  62.68 &  67.61 &  74.98 &  45.50 & 48.32 \\ 

\hline

 {Seal~\cite{seal}} &  {NIPS’23} &  {{\color{Apricot} ViT-B}} &  {\textcolor{green}{\cmark}} &  46.59 &  45.98   &  57.15 &  62.79 &  68.18 &  75.41 &  47.24 & 48.91 \\ 

\hline

 {SuperFlow~\cite{superflow}} &  {ECCV'24} &  {{\color{Apricot} ViT-B}} &  {\textcolor{green}{\cmark}} &  47.66 &  48.09   &  59.66 &  64.52 &  69.79 &  76.57 &  48.40 & 50.20 \\ 

\hline

 {ScaLR~\cite{scalr}} &  {CVPR'24} &  {{\color{Apricot} ViT-B}} &  {\textcolor{red}{\xmark}} &  41.80 &  55.83 &  63.46 &  65.24 &  68.70 &  74.76 &  45.59 & 49.60 \\ 

\hline

 {LiMoE~\cite{xu2025limoe}} &  {CVPR'25}         &  {{\color{Apricot} ViT-B}} &  {\textcolor{green}{\cmark}} &  49.07 &  50.23 &  61.51 &  66.17 &  \cellcolor{white!80!Apricot}\underline{71.56} &  77.81 &  50.30 & 51.77 \\ \hline

CleverDistiller~\cite{govindarajan2025cleverdistiller} &  {BMVC'25} &  {{\color{Apricot} ViT-B}} &  {\textcolor{red}{\xmark}} &  51.89 &  \cellcolor{white!80!Apricot}\underline{59.80} &  \cellcolor{white!80!Apricot}\underline{66.44} &  \cellcolor{white!80!Apricot}\underline{67.65} &  69.53 &  \cellcolor{white!80!Apricot}\underline{78.49} &  \cellcolor{white!80!Apricot}\underline{51.48} & \cellcolor{white!80!Apricot}\underline{53.56} \\ \hline 
SuperFlow++~\cite{superflow++}& TPAMI'25&  {{\color{Apricot} ViT-B}} &  {\textcolor{green}{\cmark}}& 48.86& 49.56& 60.75& 65.46& 70.19& 77.29& 49.90& 51.65 \\ \hline

LiMA~\cite{lima}& ICCV'25 &  {{\color{Apricot} ViT-B}}  & {\textcolor{red}{\xmark}} & \cellcolor{white!80!Apricot}\underline{56.65} & 51.29& 61.11& 65.62& 70.43& 76.91& 50.44& 51.35 \\ \hline

\textbf{\modelname$^\dagger$} & - & {{\color{Apricot} ViT-B}}   & {\textcolor{red}{\xmark}}  & 56.51 & 61.01 & 67.17 & 70.13  & 72.83   & 77.53 & 51.58 & 52.11 \\ 

\textbf{\modelname} & - & {{\color{Apricot} ViT-B}}   & {\textcolor{red}{\xmark}}  & \cellcolor{white!51!Apricot}\textbf{58.95} & \cellcolor{white!51!Apricot}\textbf{62.71} & \cellcolor{white!51!Apricot}\textbf{70.19} & \cellcolor{white!51!Apricot}\textbf{71.00} & \cellcolor{white!51!Apricot}\textbf{73.68} & \cellcolor{white!51!Apricot}\textbf{79.12} & \cellcolor{white!51!Apricot}\textbf{53.44} & \cellcolor{white!51!Apricot}\textbf{53.89}  \\

\arrayrulecolor{black} \midrule \midrule \arrayrulecolor{lightgray}


 {SLidR~\cite{slidr}} &  {CVPR’22} &  {{\color[HTML]{6434FC} ViT-L}} &  {\textcolor{green}{\cmark}} &  45.70 &  42.77 &  57.45 &  63.20 &  68.13 &  75.51 &  47.01 & 48.60 \\ 

\hline

 {Seal~\cite{seal}} &  {NIPS’23} &  {{\color[HTML]{6434FC} ViT-L}} &  {\textcolor{green}{\cmark}} &  46.81 &  46.27 &  58.14 &  63.27 &  68.67 &  75.66 &  47.55 & 50.02 \\ 

\hline

 {SuperFlow~\cite{superflow}} &  {ECCV'24} &  {{\color[HTML]{6434FC} ViT-L}} &  {\textcolor{green}{\cmark}} &  48.01 &  49.95   &  60.72 &  65.09 &  70.01 &  77.19 &  49.07 & 50.67 \\ 

\hline

 {ScaLR~\cite{scalr}} &  {CVPR'24} &  {{\color[HTML]{6434FC} ViT-L}} &  {\textcolor{red}{\xmark}} &  40.12 &  55.78 &  63.28 &  64.76 &  68.19 &  75.09 &  44.85 & 50.34 \\ 

\hline

 {LiMoE~\cite{xu2025limoe}} &  {CVPR'25} &  {{\color[HTML]{6434FC} ViT-L}}  &  {\textcolor{green}{\cmark}} &  49.35 &  51.41 &  62.07 &  66.64 &  \cellcolor{white!85!Periwinkle}\underline{71.59} &  77.85 &  50.69 & 51.93 \\ 

\hline

CleverDistiller~\cite{govindarajan2025cleverdistiller} & BMVC'25 & {\color[HTML]{6434FC} ViT-L} & \textcolor{red}{\xmark} & 52.45 & \cellcolor{white!85!Periwinkle}\underline{60.64} & \cellcolor{white!85!Periwinkle}\underline{67.03} & \cellcolor{white!85!Periwinkle}\underline{67.29} & 70.45 & \cellcolor{white!85!Periwinkle}\underline{78.29} & 52.28 & \cellcolor{white!55!Periwinkle}\textbf{54.83} \\ \hline

SuperFlow++~\cite{superflow++}& TPAMI'25&  {{\color[HTML]{6434FC} ViT-L}}  &  {\textcolor{green}{\cmark}} & 49.78& 50.92& 61.83& 66.30& 71.07& 77.63& 50.33& 52.12 \\ \hline

LiMA~\cite{lima} & ICCV'25 &  {\color[HTML]{6434FC} ViT-L}  & {\textcolor{red}{\xmark}} & \cellcolor{white!85!Periwinkle}\underline{56.67} & 53.22 & 62.46 & 66.00 & 70.59 & 77.23 & \cellcolor{white!85!Periwinkle}\underline{52.29} & 51.19 \\ \hline

\textbf{\modelname$^\dagger$} & - & {\color[HTML]{6434FC} ViT-L}   & {\textcolor{red}{\xmark}}  & 56.91 & 61.60  & 67.57 &  70.71 &  72.68 & 78.47 & 51.99 & 52.64  \\ 

\textbf{\modelname} & - &  {\color[HTML]{6434FC} ViT-L}  & {\textcolor{red}{\xmark}}  & \cellcolor{white!55!Periwinkle}\textbf{60.06}  & \cellcolor{white!55!Periwinkle}\textbf{62.82} & \cellcolor{white!55!Periwinkle}\textbf{70.58} & \cellcolor{white!55!Periwinkle}\textbf{72.55}  & \cellcolor{white!55!Periwinkle}\textbf{75.06} & \cellcolor{white!55!Periwinkle}\textbf{79.44} & \cellcolor{white!55!Periwinkle}\textbf{54.48} & \cellcolor{white!85!Periwinkle}\underline{54.41} \\

\arrayrulecolor{black} \bottomrule

\end{tabular}
}

\end{table*}

\begin{figure}
    \centering
    \includegraphics[width=0.9\linewidth, trim={0 18cm 1cm 0}, clip]{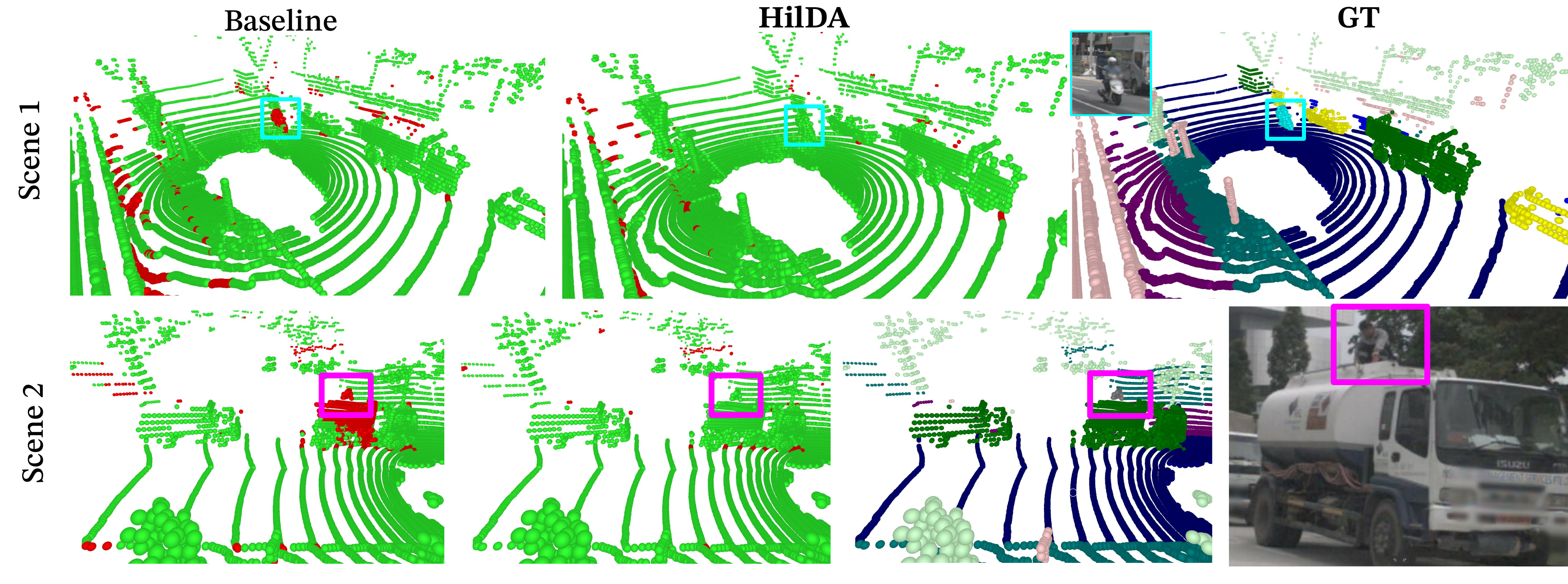}
    \caption{Qualitative comparison of segmentation performance of \modelname{} and ScaLR. \modelname{} makes fewer errors and correctly segments rare cases like a scooter driver (Scene 1) or a person on top of a truck (Scene 2).}
    \label{fig:four_col_comparison}
\end{figure}

\paragraph{\textbf{Main Results.}} As shown in~\cref{tab:minkunet_main_table},\textbf{ \modelname{}} achieves SOTA performance across all scenarios (except for one - second best). The most significant gains are observed in the data-scarce scenarios (1\%--10\%) where our method outperforms previous SOTA by a significant margin, including CleverDistiller~\cite{govindarajan2025cleverdistiller}, LiMoE~\cite{limoe}, and ScaLR~\cite{scalr}. While LiMA~\cite{lima} previously set a clear margin on linear probing (LP), our method surpasses it while also delivering stronger fine-tuning results. This suggests that our hierarchical distillation with a temporal diffusion strategy yields a richer representation than prior methods relying on \eg, global contrastive learning or semantic priors, thereby reducing dependence on large-scale annotated data. Additionally, it substantially improves transfer to new domains like Semantic-KITTI~\cite{behley2019semantickitti} or the Waymo Open Dataset~\cite{sun2020waymo}. This pattern holds across different teacher sizes (ViT-S, ViT-B, and ViT-L). We also qualitatively observe this in~\cref{fig:four_col_comparison}, where our features yield notably lower segmentation error and work much better for long-tail cases (\eg, a person on the top of a truck).
\paragraph{\textbf{Domain Generalization.}}
In~\cref{tab:generalization}, we evaluate the transferability of our model, \pretrained on nuScenes, to four diverse datasets. The results show that our learned representations are transferable to various sensor configurations and urban layouts. On ScriKITTI~\cite{unal2022scribble}, despite extremely sparse scribble-level annotations, our method maintains top performance, demonstrating its ability to extract meaningful features from minimal and spatially disconnected supervision. On Rellis-3D~\cite{jiang2021rellis}, which is recorded in an unstructured off-road environment, our model outperforms strong baselines such as LiMoE~\cite{limoe}. This indicates that our approach captures fundamental geometric and semantic properties that generalize across drastically different terrains. Consistent gains on SemSTF~\cite{xiao20233d} and DAPS-3D~\cite{klokov2023daps3d} further confirm that our \pretraining produces robust, dataset-agnostic features serving as strong universal initializations. 

\begin{table}[ht] 
\centering
\scriptsize
\caption{Domain generalization benchmarks. Models \pretrained on nuScenes and fine-tuned on listed datasets. Metrics in mIoU.}
\label{tab:generalization}
\begin{tabular}{lcccccccc}
\toprule
\multirow{2}{*}{\textbf{Method}} & \multicolumn{2}{c}{\textbf{ScriKITTI}} & \multicolumn{2}{c}{\textbf{Rellis-3D}} & \multicolumn{2}{c}{\textbf{SemSTF}} & \multicolumn{2}{c}{\textbf{DAPS-3D}} \\
\cmidrule(lr){2-3} \cmidrule(lr){4-5} \cmidrule(lr){6-7} \cmidrule(lr){8-9}
& \multicolumn{1}{l}{\textbf{1\%}} & {\textbf{10\%}} & \multicolumn{1}{l}{\textbf{1\%}} & {\textbf{10\%}} & \multicolumn{1}{l}{\textbf{50\%}} & {\textbf{100\%}} & \multicolumn{1}{l}{\textbf{50\%}} & \textbf{100\%} \\ 
\midrule
Random & 23.81 & 47.60 & 38.46 & 53.60 & 48.03 & 48.15 & 74.32 & 79.38 \\ 
\arrayrulecolor{lightgray} \hline

SLidR& 39.60 & 50.45 & 49.75 & 54.57 & 52.01 & 54.35 & 81.00 & 85.40 \\ 
\hline
Seal & 40.64 & 52.77 & 51.09 & 55.03 & 53.46 & 55.36 & 81.88 & 85.90 \\ 
\hline

ScaLR & 36.45 & 49.16 & 47.91 & 48.86 & 52.10 & 54.40 & 81.92 & 85.58 \\ 
\hline
SuperFlow & 42.70 & 54.00 & 52.83 & 55.71 & 54.72 & 56.57 & 82.43 & 86.21 \\ 
\hline
LiMoE & 43.95 & 55.96 & 53.74 & 56.67 & \cellcolor{white!80!Apricot}\underline{55.60} & \cellcolor{white!80!Apricot}\underline{57.31} & \cellcolor{white!80!Apricot}\underline{83.24} & 86.68 \\ \hline

CleverD & 44.03 & \cellcolor{white!80!Apricot}\underline{56.70} & \cellcolor{white!80!Apricot}\underline{58.35} & \cellcolor{white!80!Apricot}\underline{60.92} & 53.99 & 55.66 & 83.06 & \cellcolor{white!80!Apricot}\underline{87.95} \\ \hline

LiMA & \cellcolor{white!80!Apricot}\underline{45.90} & 55.13 & 55.62 & 57.15& 55.45 & 56.70 & 83.11 & 86.63 \\ \hline

\textbf{\modelname{}$^\dagger$} & 48.26 & 59.02 & 56.77 & 59.04 & 55.86 & 58.11 & 84.26 & 87.18 \\ 

\textbf{\modelname{}} & \cellcolor{white!51!Apricot}\textbf{48.39} & \cellcolor{white!51!Apricot}\textbf{58.22} & \cellcolor{white!51!Apricot}\textbf{59.49} & \cellcolor{white!51!Apricot}\textbf{62.88} & \cellcolor{white!51!Apricot}\textbf{56.72} & \cellcolor{white!51!Apricot}\textbf{57.83} & \cellcolor{white!51!Apricot}\textbf{85.93} & \cellcolor{white!51!Apricot}\textbf{89.08} \\ 
\arrayrulecolor{black} \bottomrule
\end{tabular}
\end{table}

\paragraph{\textbf{Robustness.}}

In~\cref{tab:robust_lp} (and supplementary~\cref{sec:additional_results} for full fine-tuning), we evaluate feature robustness on the corrupted \mbox{nuScenes-C}~\cite{kong2023robo3d} benchmark. \modelname{} exhibits superior resilience compared to previous methods, achieving the lowest mean Corruption Error (mCE) and highest mean Relative Robustness (mRR).

It performs particularly well under \textit{Cross-Sensor} and \textit{Snowy} corruptions, which involve heavy interference and sparsity, clearly outperforming previous SOTA methods in these scenarios. These results confirm that our combination of hierarchical distillation and temporal occupancy prediction produces reliable 3D representations even under severe sensor degradation. 
\begin{table*}[t]
\renewcommand{\arraystretch}{1} 
\setlength{\tabcolsep}{1.8pt}
\centering
\scriptsize
\caption{Comparison on robustness under weather corruption and sensor failure on nuScenes-C~\cite{kong2023robo3d} benchmark using linear probing. Metrics given in percentage (\%).}
\label{tab:robust_lp}


\begin{tabular}{l cc ccccccccc}
\toprule

\multirow{2}{*}{Method} & 
\multirow{2}{*}{\textbf{mCE $\downarrow$}} & 
\multirow{2}{*}{\textbf{mRR $\uparrow$}} & 
\multicolumn{9}{c}{mIoU $\uparrow$} \\

\cmidrule(lr){4-12}

& & & {Fog} & {Rain} & {Snow} & {Blur} & {Beam} & {Cross} & {Echo} & {Sensor} & \textbf{Mean} \\ 
\midrule

{SLidR~\cite{slidr}} & 
\cellcolor{gray!10} 179.38 & \cellcolor{gray!10} 77.18 & 
34.88 & 38.09 & 32.64 & 26.44 & 33.73 & 20.81 & 31.54 & 21.44 & 
\cellcolor{gray!10} 29.95 \\ \hline

{Seal~\cite{seal}} & 
\cellcolor{gray!10} 166.18 & \cellcolor{gray!10} 75.38 & 
37.33 & 42.77 & 29.93 & 37.73 & 40.32 & 20.31 & 37.73 & 24.94 & 
\cellcolor{gray!10} 33.88 \\ \hline 

{SuperFlow~\cite{superflow}} & 
\cellcolor{gray!10} 161.78 & \cellcolor{gray!10} 75.52 & 
37.59 & 43.42 & 37.60 & 39.57 & 41.40 & 23.64 & 38.03 & 26.69 & 
\cellcolor{gray!10} 35.99 \\ \hline

{Scalr~\cite{scalr}} & 
\cellcolor{gray!10} 173.18 & \cellcolor{gray!10} 78.91 & 
37.55 & 37.96 & 36.29 & 33.64 & 33.06 & 23.01 & 33.62 & 23.70 & 
\cellcolor{gray!10} 33.59 \\ \hline

{LiMoE~\cite{limoe}} & 
\cellcolor{gray!10} 155.77 & \cellcolor{gray!10} 78.23 & 
40.35 & 45.28 & 39.14 & 42.10 & 44.21 & 27.33 & 39.20 & 29.49 & 
\cellcolor{gray!10} 38.39 \\ \hline

{CleverD \cite{govindarajan2025cleverdistiller}} & 
\cellcolor{gray!10} 151.21 & \cellcolor{white!90!Apricot}\underline{79.76} & 
43.96 & 46.91 & 41.20 & 41.05 & 42.15 & \cellcolor{white!90!Apricot}\underline{45.67} & 41.30 & 28.85 & 
\cellcolor{gray!10} 41.39 \\ \hline

{LiMA~\cite{lima}} & 
\cellcolor{white!90!Apricot}\underline{137.23} & \cellcolor{gray!10} 79.30 & 
\cellcolor{white!90!Apricot}\underline{51.52} & \cellcolor{white!90!Apricot}\underline{54.90} & \cellcolor{white!90!Apricot}\underline{45.63} & \cellcolor{white!90!Apricot}\underline{50.55} & \cellcolor{white!90!Apricot}\underline{49.67} & 27.24 & \cellcolor{white!90!Apricot}\underline{45.76} & \cellcolor{white!90!Apricot}\underline{34.09} & 
\cellcolor{white!90!Apricot}\underline{44.92} \\ \hline 

{\textbf{\modelname{}$^\dagger$}} & 
\cellcolor{gray!10} 134.63 & \cellcolor{gray!10} 80.70 & 50.41 & 55.86 & 49.63 & 49.95 & 45.25 & 50.18 & 44.09 & 34.44 & \cellcolor{gray!10} 47.58\\ 

{\textbf{\modelname{}}} & 
\cellcolor{white!51!Apricot}\textbf{124.27} & 
\cellcolor{white!51!Apricot}\textbf{88.20} & 
\cellcolor{white!51!Apricot}\textbf{55.99} & \cellcolor{white!51!Apricot}\textbf{57.97} & \cellcolor{white!51!Apricot}\textbf{54.29} & \cellcolor{white!51!Apricot}\textbf{51.16} & \cellcolor{white!51!Apricot}\textbf{51.30} & \cellcolor{white!51!Apricot}\textbf{56.29} & \cellcolor{white!51!Apricot}\textbf{47.91} & \cellcolor{white!51!Apricot}\textbf{40.31} & 
\cellcolor{white!51!Apricot}\textbf{52.00}\\ 

\arrayrulecolor{black}
\bottomrule
\end{tabular}
\end{table*}

\subsection{Transferability to Spatiotemporal Tasks}
\label{sec:multi-task}
We further assess transfer to tasks requiring spatial and temporal reasoning: 3D object detection (spatial), semantic occupancy (spatiotemporal), and scene flow (spatiotemporal). All experiments use the same ViT-B/\minkunet{} backbone after \pretraining. We re-trained ScaLR, SuperFlow, and CleverDistiller (SF/CD) using author settings and verified that their segmentation performance matches the reported results\footnote{\scriptsize Checkpoints/complete code for LiMoE/LiMA is unavailable, authors reported lost access. CD (code provided by CD's authors)/ScaLR/SF code was used for obtaining the checkpoints.}; details are in supplementary~\cref{sec:implementationdetails}.

\begin{figure}[htbp]
    \centering
    
    \begin{minipage}[t]{0.4\textwidth}
        \vspace{0pt} 
        \centering
        \scriptsize
        \renewcommand{\arraystretch}{1} 
        \setlength{\tabcolsep}{0.9pt}
        \captionof{table}{Evaluation on 3D Object Detection (3DOD) benchmarks.}
        \begin{tabular}{lcccccc}
            \toprule
            \multirow{2}{*}{Method} & \multicolumn{3}{c}{KITTI (mAP)} & \multicolumn{3}{c}{nuSc (mAP)} \\
            \cmidrule(lr){2-4} \cmidrule(lr){5-7}
            & 5\% & 10\% & 20\% & 5\% & 10\% & 20\% \\
            \midrule
            Random & 56.1 & 59.1 & 61.6 & 38.1 & 43.5 & 45.2 \\
            PPKT & 57.8 & 60.1 & 61.2 & - & - & -\\
            SLidR & 57.8 & 61.4 & 62.4 & - & - & -\\
            ScaLR & 56.2 & 62.3 & 66.5 & 46.1 & 50.3 & \cellcolor{white!80!Apricot}\underline{55.1} \\
            SF & 59.3 & 62.7 & 64.2 & 45.9 & 51.1 & 54.5 \\

            CD & 
            \cellcolor{white!80!Apricot}\underline{59.8} & 
            \cellcolor{white!80!Apricot}\underline{66.6} & 
            \cellcolor{white!80!Apricot}\underline{67.1} & 
            \cellcolor{white!80!Apricot}\underline{47.9} & 
            \cellcolor{white!80!Apricot}\underline{51.9} & 
            54.3 \\
            
            \textbf{\modelname{}}$^\dagger$ & 
            60.1 & 
            66.3 & 
            69.4 &
            47.0 &  
            52.2 & 
            55.7 \\
            
            \textbf{\modelname{}} & 
            \cellcolor{white!51!Apricot}\textbf{61.4} & 
            \cellcolor{white!51!Apricot}\textbf{67.3} & 
            \cellcolor{white!51!Apricot}\textbf{71.0} & 
            \cellcolor{white!51!Apricot}\textbf{50.4} & 
            \cellcolor{white!51!Apricot}\textbf{54.9} & 
            \cellcolor{white!51!Apricot}\textbf{57.9} \\
            \bottomrule
        \end{tabular}
        \label{table:3dod}
    \end{minipage}\hfill
    \begin{minipage}[t]{0.57\textwidth}
        \vspace{0pt} 
        \centering
        \includegraphics[width=\linewidth, trim={0 17.7cm 15cm 0}, clip]{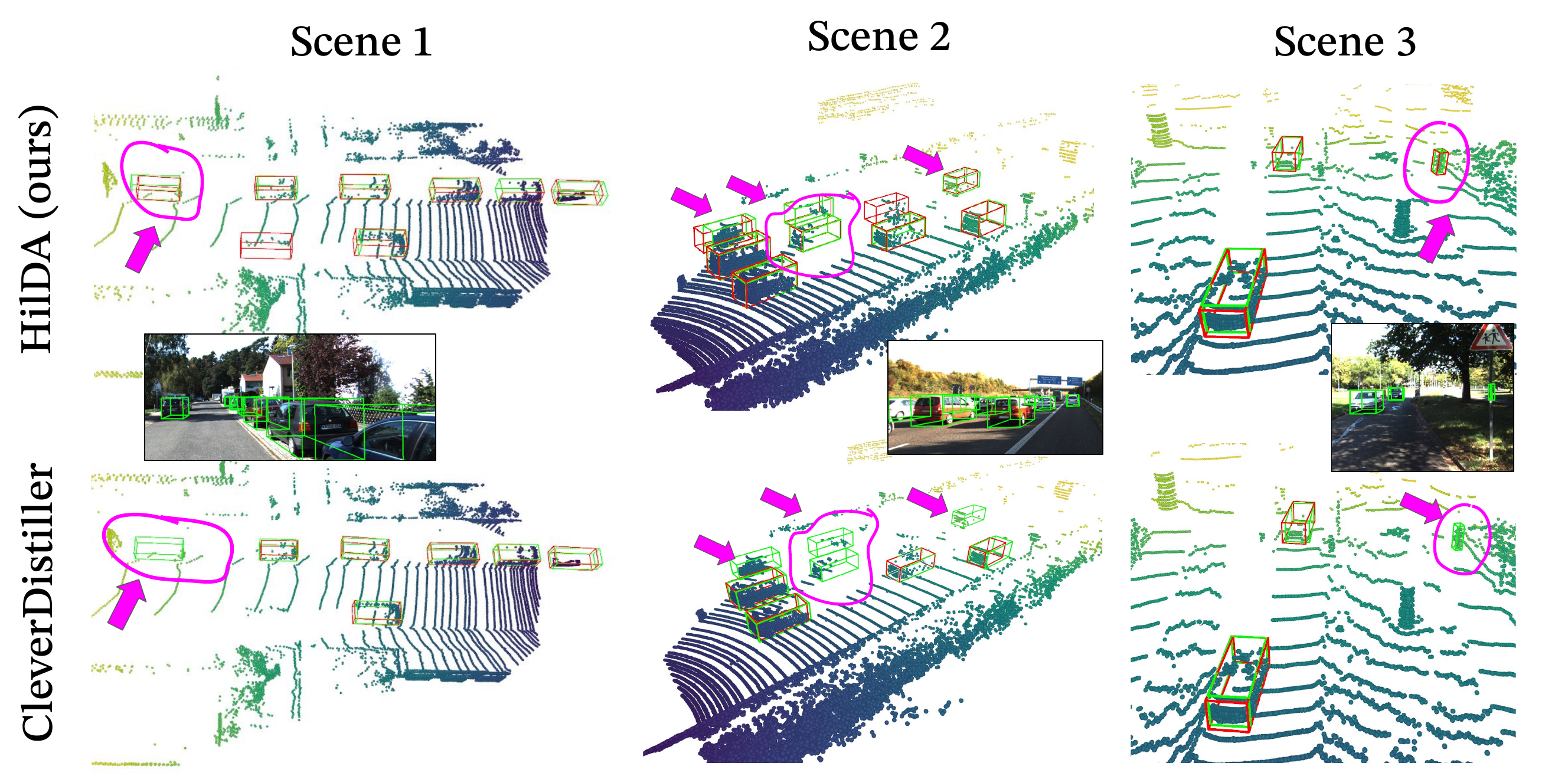}
        \captionof{figure}{Qualitative comparison of 3DOD.  Green boxes are GT, and red are predictions.}
        \label{fig:3dod_nofp}
    \end{minipage}
    
\end{figure}

\paragraph{\textbf{3D Object Detection.}}
We evaluate PointRCNN~\cite{shi2019pointrcnn} with \modelname{} on KITTI~\cite{kitti} and nuScenes~\cite{caesar2020nuscenes}. As shown in~\cref{table:3dod}, \modelname{} consistently outperforms prior distillation baselines across datasets. Notably, we find that our model demonstrates robust detections in complex scenarios, such as at long-range and under heavy occlusion (see~\cref{fig:3dod_nofp}). This suggests that hierarchical distillation with temporal diffusion yields semantically consistent and geometrically grounded features. 

%

\paragraph{\textbf{Semantic Occupancy.}}
We evaluate 3D/4D semantic occupancy following Occ4cast~\cite{Liu2023occ4cast}, with added class labels. The \pretrained backbone is frozen and only a lightweight decoder is trained. Inputs are $t_{-1}$ and $t_0$ ($\Delta t=0.5$s). We predict occupancy at $t_0$ for 3D, and at $t_{0}$ and $t_1$ for 4D, and report frame-averaged metrics. \Cref{tab:4docc_sem_agg_agg} aggregates IoU into \textit{dynamic}, \textit{static}, and \textit{surface} classes (see~\cref{sec:additional_results} for a class-wise breakdown across time horizons and a detailed discussion on the effects of including the diffusion auxiliary task. \Cref{sec:add_qualitative} shows qualitative results). \modelname{} outperforms all baselines in 3D and 4D. The largest gains from including the diffusion-based loss during \pretraining are on dynamic/object-centric classes where future occupancy directly supervises object permanence and short-term motion. Surface gains are smaller because the ground-removed \pretraining BEV target emphasizes free/occupied boundaries and removes vertical cues useful for within-region semantic distinctions (\eg, terrain \vs sidewalk). \Cref{fig:sem_occ_combo} shows that \modelname{} maintains the highest mIoU across a horizon of up to $5$ seconds.

%
\pgfplotscreateplotcyclelist{methodcycle}{
  {blue,   mark=*,         line width=0.9pt},
  {orange,    mark=square*,    line width=0.9pt},
  {green!60!black, mark=triangle*, line width=0.9pt},
  {violet, mark=diamond*,   line width=0.9pt},
  {red, mark=o,          line width=0.9pt},
}

\begin{figure}[h!]
    \centering
    \begin{minipage}[c]{0.5\linewidth}
        \makeatletter\def\@captype{table}\makeatother 
        \centering
        \scriptsize
        \setlength{\tabcolsep}{2pt}
        \renewcommand{\arraystretch}{0.8}
        \caption{Comparison of frozen pre-trained backbones on nuScenes semantic occupancy prediction. Frame-averaged metrics.}
        \begin{tabular}{c l|r|rrr}
            \toprule
             & \textbf{Method} &
            \textbf{mIoU} &
            \textbf{Dyna.} & \textbf{Stat.} & \textbf{Surf.} \\
            \midrule
            
            \multirow{6}{*}{\rotatebox{90}{\textbf{3D}}} & Random & 10.6 & 3.7  & 10.6 & 24.4 \\
             & ScaLR & 16.2 & 9.5 & 16.4 & \cellcolor{white!80!Apricot}\underline{29.4} \\
             & SF & 12.5 & 5.2 & 12.1 & 25.7 \\
             & CD & \cellcolor{white!80!Apricot}\underline{16.5} & \cellcolor{white!80!Apricot}\underline{10.0} & \cellcolor{white!80!Apricot}\underline{16.5} & \cellcolor{white!80!Apricot}\underline{29.4} \\
             
             & \textbf{\modelname{}$^\dagger$}   & 
             19.0 & 
             13.3 & 
             19.0 & 
             30.2 \\
             
             & \textbf{\modelname{}} & 
             \cellcolor{white!51!Apricot}\textbf{20.0} & 
             \cellcolor{white!51!Apricot}\textbf{14.5} & 
             \cellcolor{white!51!Apricot}\textbf{20.2} & 
             \cellcolor{white!51!Apricot}\textbf{30.4} \\
            \midrule
            
            \multirow{6}{*}{\rotatebox{90}{\textbf{4D}}} & Random & 10.3 & 3.1 & 10.7 & 24.4 \\
             & ScaLR & 14.9 & 7.1 & 15.9 & 29.5 \\
             & SF & 11.9 & 4.9 & 12.2 & 25.6 \\
             & CD & \cellcolor{white!80!Apricot}\underline{15.5} & \cellcolor{white!80!Apricot}\underline{8.0} & \cellcolor{white!80!Apricot}\underline{16.5} & \cellcolor{white!80!Apricot}\underline{29.6} \\
             
             & \textbf{\modelname{}$^\dagger$} & 
             17.3 & 
             10.1 & 
             18.5 & 
             30.4 \\
             
             & \textbf{\modelname{}} & 
             \cellcolor{white!51!Apricot}\textbf{18.4} & 
             \cellcolor{white!51!Apricot}\textbf{11.4} & 
             \cellcolor{white!51!Apricot}\textbf{20.2} & 
             \cellcolor{white!51!Apricot}\textbf{30.5} \\
            \addlinespace[2pt]
            
            \bottomrule
        \end{tabular}
            
            
            
        \label{tab:4docc_sem_agg_agg}
    \end{minipage}
    \hfill
    \begin{minipage}[c]{0.47\linewidth}
        \raggedright
        \begin{tikzpicture} 
            \begin{axis}[
                width=1\linewidth,
                height=0.9\linewidth,
                xlabel={Forecast frame at $t_k$},
                ylabel={mIoU (\%)},
                ylabel style={yshift=-6pt},
                xmin=0, xmax=10,
                xtick={0,1,2,3,4,5,6,7,8,9,10},
                xticklabels={$t_0$,$t_1$,$t_2$,$t_3$,$t_4$,$t_5$,$t_6$,$t_7$,$t_8$,$t_9$,$t_{10}$},
                ymax=20,
                ytick={10,12,14,16,18},
                grid=both,
                major grid style={opacity=0.25},
                minor grid style={opacity=0.10},
                cycle list name=methodcycle,
                legend columns=2,
                legend style={
                    font=\tiny,
                    at={(0.14,1.0)}, anchor=north west,   
                    draw=none,
                    fill=white, fill opacity=0.75,         
                    text opacity=1,
                    inner sep=1pt,
                },
                mark size=1.6pt,
                line width=0.9pt,
            ]
            
            \addplot+ coordinates {
            (0,17.7) (1,16.5) (2,16.0) (3,15.6) (4,15.4) (5,15.1) (6,14.9) (7,14.7) (8,14.5) (9,14.4) (10,14.3)
            };
            \addlegendentry{\modelname{}}
            
            \addplot+ coordinates {
            (0,16.9) (1,15.7) (2,15.2) (3,15.0) (4,14.7) (5,14.5) (6,14.4) (7,14.2) (8,14.1) (9,14.0) (10,13.8)
            };
            \addlegendentry{\modelname{}$^\dagger$}
            

            \addplot+ coordinates {
            (0,15.2) (1,14.3) (2,14.0) (3,13.8) (4,13.5) (5,13.4) (6,13.3) (7,13.1) (8,13.0) (9,12.9) (10,12.7)
            };
            \addlegendentry{CleverDistiller}
            
            \addplot+ coordinates {
            (0,14.6) (1,13.9) (2,13.6) (3,13.5) (4,13.3) (5,13.1) (6,13.0) (7,12.9) (8,12.8) (9,12.7) (10,12.6)
            };
            \addlegendentry{ScaLR}
            
            \addplot+ coordinates {
            (0,11.8) (1,11.3) (2,11.0) (3,10.9) (4,10.8) (5,10.7) (6,10.6) (7,10.5) (8,10.5) (9,10.4) (10,10.4)
            };
            \addlegendentry{SF}

            \addplot+ coordinates {
            (0,10.0) (1,9.6) (2,9.5) (3,9.4) (4,9.3) (5,9.3) (6,9.2) (7,9.2) (8,9.2) (9,9.2) (10,9.1)
            };
            \addlegendentry{Random}
    
            \end{axis}
        \end{tikzpicture} 
        \caption{Semantic mIoU over forecast horizon. Decoder trained from $t_0$ through $t_{10}$.}
        \label{fig:sem_occ_combo}
    \end{minipage}
\end{figure}
%
\paragraph{\textbf{Scene Flow.}}
\begin{figure}[t]
    \centering
    \begin{minipage}{0.56\textwidth}
        \centering

        \scriptsize
        \setlength{\tabcolsep}{2pt} 
        \renewcommand{\arraystretch}{1.5}
        \captionof{table}{Scene flow on ArgoverseV2~\cite{Argoverse2}. Ego Motion refers to the apparent motion of points caused solely by the movement of ego itself.}
        \resizebox{\textwidth}{!}{
            \begin{tabular}{l c c c c c c c}
                \toprule
                \multirow{2}{*}{\textbf{Methods}} &  & \multicolumn{4}{c}{\textbf{3-way EPE}} & \multicolumn{2}{c}{\textbf{Norm. EPE}} \\
                \cmidrule(lr){3-6} \cmidrule(lr){7-8}
                 & & \textbf{Mean} $\downarrow$ & \textbf{FD}$\downarrow$ & \textbf{FS}$\downarrow$ & \textbf{BS}$\downarrow$ & \textbf{Dyn}$\downarrow$ & \textbf{Stat}$\downarrow$ \\
                \midrule
                Ego Motion & & 0.181 & 0.534 & 0.010 & 0.000 & 1.000 & 0.007 \\
                \hline
                
                SSF~\cite{khoche2025ssf} & & 0.028 & 0.058 & 0.018 & 0.009 & 0.267 & 0.013 \\
                
                +ScaLR~\cite{scalr} & & 
                0.026 & 
                0.056 & 
                0.016 & 
                \cellcolor{white!80!Apricot}\underline{0.007} & 
                0.234 & 
                \cellcolor{white!80!Apricot}\underline{0.012}\\
                
                +SF~\cite{superflow++} & & 
                0.025 & 
                0.053 & 
                \cellcolor{white!80!Apricot}\underline{0.015} & 
                \cellcolor{white!80!Apricot}\underline{0.007} & 
                \cellcolor{white!80!Apricot}\underline{0.198} & 
                \cellcolor{white!80!Apricot}\underline{0.012} \\
                
                +CD~\cite{govindarajan2025cleverdistiller} & & 
                \cellcolor{white!80!Apricot}\underline{0.024} &
                \cellcolor{white!80!Apricot}\underline{0.049}  & 
                0.016 & 
                0.008 & 
                0.210 & 
                \cellcolor{white!51!Apricot}\textbf{0.011}\\
                
                \textbf{+\modelname{}$^\dagger$} & &
                0.024  &
                0.051 &
                0.015 &
                0.007   & 
                0.181 & 
                0.011  \\
                
                \textbf{+\modelname{}} & &
                \cellcolor{white!51!Apricot}\textbf{0.021}  & 
                \cellcolor{white!51!Apricot}\textbf{0.044} & 
                \cellcolor{white!51!Apricot}\textbf{0.014} & 
                \cellcolor{white!51!Apricot}\textbf{0.006}  & 
                \cellcolor{white!51!Apricot}\textbf{0.146} &
                \cellcolor{white!51!Apricot}\textbf{0.011} \\
                \bottomrule
            \end{tabular}
            
        }
        \label{tab:av2_comparison}
    \end{minipage}
    \hfill
    \begin{minipage}{0.4\textwidth}
        \centering
        \includegraphics[width=\linewidth, trim={0.5cm 12.75cm 32.5cm 0.6cm}, clip]{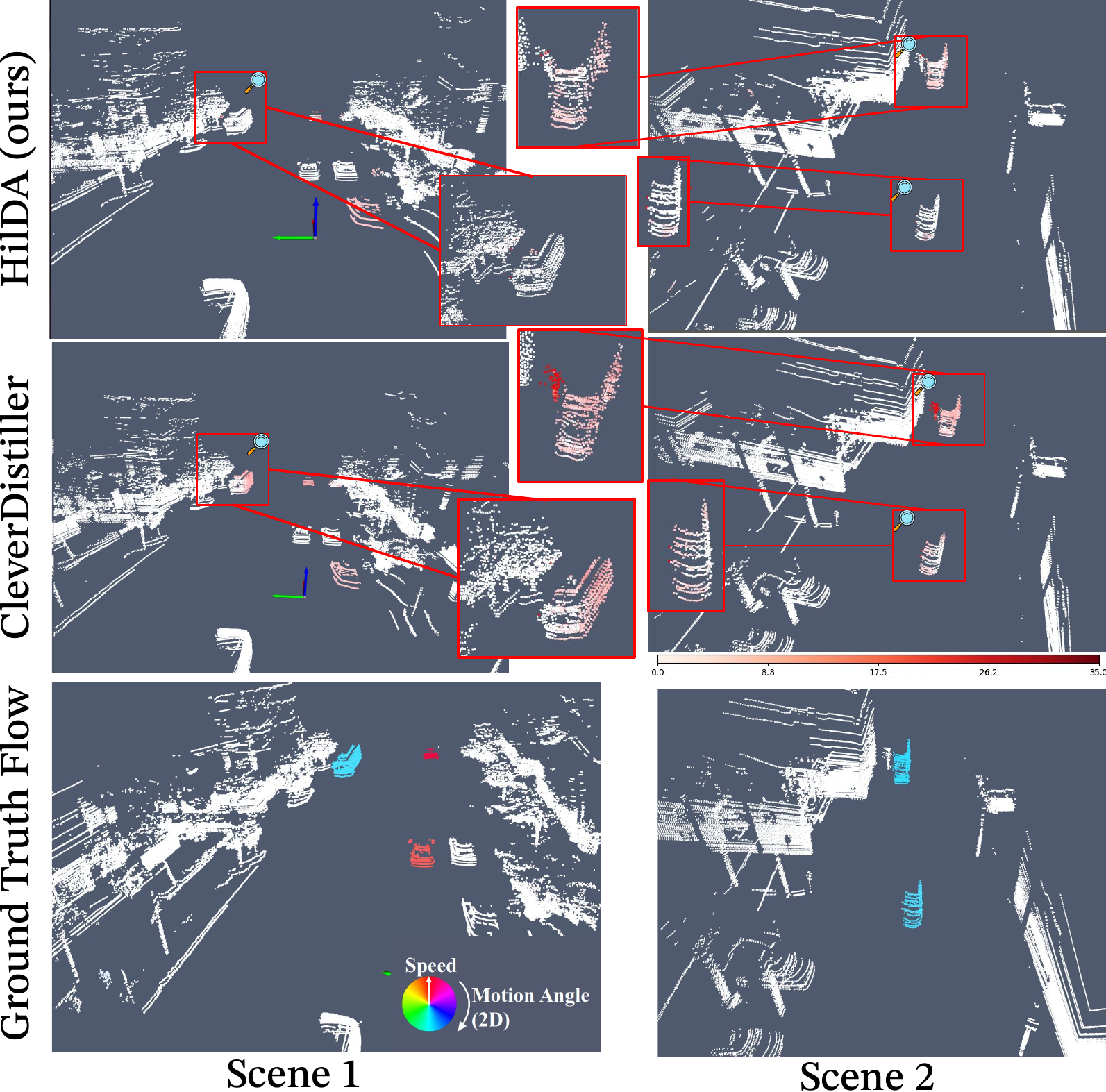}
        \caption{Qualitative scene flow comparison.}
        \label{fig:scene_flow_vis}
    \end{minipage}
\end{figure}
%
We integrate \mbox{\minkunet} into SSF~\cite{khoche2025ssf} and evaluate scene flow on Argoverse~2~\cite{Argoverse2}, jointly testing task transfer and domain generalization. Replacing the randomly initialized backbone with \modelname{} \pretrained weights improves all metrics in~\cref{tab:av2_comparison}, especially dynamic object estimation. \Cref{fig:scene_flow_vis} shows lower speed/direction errors (indicated by fewer red regions) and cleaner motion maps compared to baseline, indicating that \modelname{} features better capture temporal structure and enable better motion estimation where baselines fail.

\subsection{Ablations}
\label{sec:ablations}
We ablate the different components and design choices of our method, using 3D semantic segmentation as the evaluation task. See the supplementary material in~\cref{sec:additional_results} for an extensive set of additional ablations.


\paragraph{\textbf{Component Analysis.}} In~\cref{table:ablation_components_main}, ablation of our proposed components are presented. Adding diffusion (a) or hierarchical distillation (c) both significantly improves the baseline, while applying both components (e) yields the highest overall performance, demonstrating the synergy between hierarchical distillation and temporal occupancy diffusion.

\begin{table}[!t]
  \centering
  \scriptsize
  \setlength{\tabcolsep}{3pt}
  \caption{Ablation of \pretraining loss components. ScaLR~\cite{scalr} with MLP projection head is (base).}
  \label{table:ablation_components_main}
  \begin{tabular}{c|ccc|cc|c|c}
    \toprule
    \multirow{2}{*}{\#} &
    \multirow{2}{*}{\textbf{Diff}} &
    \multirow{2}{*}{\textbf{Distill}} &
    \multirow{2}{*}{\textbf{\cls{}}} &
    \multicolumn{2}{c|}{nuScenes} & SK & Waymo \\
    & & & & LP & 1\% & 1\% & 1\% \\
    \midrule
    (base) & \textcolor{red}{\xmark} & \textcolor{red}{\xmark} & \textcolor{red}{\xmark} & 46.36 & 55.01 & 50.15 & 49.08 \\
    (a) & \textcolor{green}{\cmark} & \textcolor{red}{\xmark} & \textcolor{red}{\xmark} & 50.43 & 56.97 & 49.59 & 49.84 \\
    (b) & \textcolor{red}{\xmark} & \textcolor{green}{\cmark} & \textcolor{red}{\xmark} & 53.77 & 57.03 & 50.63 & 50.71 \\
    (c) & \textcolor{red}{\xmark} & \textcolor{green}{\cmark} & \textcolor{green}{\cmark} & 55.13 & 57.81 & 50.80 & 50.87 \\
    (d) & \textcolor{green}{\cmark} & \textcolor{green}{\cmark} & \textcolor{red}{\xmark} & \underline{55.53} & \underline{59.04} & \underline{52.41} & \underline{51.89} \\
    \rowcolor{white!40!SpringGreen}
    \cellcolor{white!40}(e) & \cellcolor{white!40}\textcolor{green}{\cmark} & \cellcolor{white!40}\textcolor{green}{\cmark} & \cellcolor{white!40}\textcolor{green}{\cmark} & \textbf{56.29} & \textbf{59.46} & \textbf{52.88} & \textbf{52.16} \\
    \bottomrule
  \end{tabular}
\end{table}


\paragraph{\textbf{Multi-Layer Distillation Design.}}
We ablate how to transfer the VFM hierarchy for calibrated point--pixel supervision. While multi-layer distillation has been studied in 2D~\cite{chen2021distilling}, camera-to-LiDAR distillation introduces an additional geometric point--pixel correspondence constraint. Earlier \minkunet{} student layers aggregate coarser, irregular 3D neighborhoods whose sparse features contain voxels that may project to different semantic regions. Additionally, overly early VFM teacher layers may not match the abstraction level of late LiDAR features. \modelname{} therefore uses what we term \textit{Separate} late-layer matching, aligning neighboring blocks (\eg, $\mathbf{Q}_{L}{\rightarrow}\mathbf{F}_{L}$ and $\mathbf{Q}_{L-1}{\rightarrow}\mathbf{F}_{L-1}$). 

\Cref{tab:ablation_multi_layer_design} shows that this design outperforms both final- and penultimate-layer distillation. Extending it to earlier layers reduces performance, but still surpasses single-layer. This supports the view that intermediate teacher features are useful when their abstraction level and geometric support remain compatible with the student features. The Separate strategy also outperforms \textit{KR-style} aggregation~\cite{chen2021distilling}, \textit{Aggregate} matching, and \textit{non-uniform} matching. In KR-style aggregation, sliding windows of teacher layers supervise successive student layers,
$\{\mathbf{Q}_{L-k}\}_{k=j}^{K-1}{\rightarrow}\mathbf{F}_{L-j}$ with $j=0,1$ for a $K$-layer aggregation. Aggregate matching instead concatenates the selected teacher-layer features into a single target,
$\{\mathbf{Q}_{L-k}\}_{k=0}^{K-1}{\rightarrow}\mathbf{F}_{L}$, whereas non-uniform matching skips intermediate layer pairs, \eg,
$(\mathbf{Q}_{L}{\rightarrow}\mathbf{F}_L, \mathbf{Q}_{L-2}{\rightarrow}\mathbf{F}_{L-2})$. Thus, the gain comes from preserving compatible late-layer correspondences rather than simply adding more teacher features. \Cref{tab:ablation_design_components_compact} further shows that cosine distance is the strongest point--pixel alignment loss. Since ViT teacher and \minkunet{} student features have different architectures and feature-norm distributions, cosine supervises semantic direction rather than cross-architecture scale matching. For a more extensive analysis, see~\cref{sec:add_ablations}.

\begin{table*}[t]
    \centering
    \scriptsize

    \begin{minipage}[t]{0.47\textwidth}
        \centering
        \setlength{\tabcolsep}{4pt}
        \renewcommand{\arraystretch}{0.85}

        \caption{Ablation of multi-layer distillation designs. All rows use only $\mathcal{L}_{\mathrm{distill}}$ during \pretraining. \modelname{} corresponds to \enquote{Separate, Last 2}.}
        \label{tab:ablation_multi_layer_design}
        \resizebox{\linewidth}{!}{
        \begin{tabular}{lc}
            \toprule
            \textbf{Design} & \textbf{nuSc LP} \\
            \midrule
            Final layer only & 46.4 \\
            Penultimate only & 49.9 \\
            \midrule
            Separate, Last 2 & \cellcolor{white!40!SpringGreen}\textbf{53.8} \\
            Separate, Last 3 & 52.0 \\
            Separate, Last 4 & 51.6 \\
            \midrule
            Aggregate, Last 3 & 53.0 \\
            3-layer KR-style~\cite{chen2021distilling} & 51.4 \\
            Non-uniform matching & 52.4 \\
            \bottomrule
        \end{tabular}}
    \end{minipage}\hfill
    \begin{minipage}[t]{0.49\textwidth}
        \centering
        \setlength{\tabcolsep}{4pt}
        \renewcommand{\arraystretch}{0.85}

        \caption{Ablation of several design choices. Each block changes one component while keeping the \modelname{} setups (cosine, max pooling, diffusion) fixed.}
        \label{tab:ablation_design_components_compact}
        \resizebox{\linewidth}{!}{
        \begin{tabular}{llc}
            \toprule
            \textbf{Comp.} & \textbf{Variant} & \textbf{nuSc LP} \\
            \midrule
            \multirow{3}{*}{$\mathcal{L}_{\mathrm{distill}}$}
                & $\ell_2$ & 53.8 \\
                & Kullback-Leibler & 52.1 \\
                & Cosine distance & \cellcolor{white!40!SpringGreen}\textbf{56.3} \\
            \midrule
            \multirow{3}{*}{\cls{} pool.}
                & Learnable pool. & 55.9 \\
                & Per-img. $\cls_{\mathbf{F}_L}$ & 55.5 \\
                & Max pooling & \cellcolor{white!40!SpringGreen}\textbf{56.3} \\
            \midrule
            \multirow{3}{*}{Occ. dec.}
                & Simple decoder & 49.1 \\
                & ALSO & 54.7 \\
                & Diffusion & \cellcolor{white!40!SpringGreen}\textbf{56.3} \\
            \bottomrule
        \end{tabular}}
    \end{minipage}
\end{table*}

\paragraph{\textbf{Global Context Distillation Design.}}

\Cref{tab:ablation_design_components_compact} shows that max pooling is the strongest aggregation strategy for constructing $\cls{}_{\mathbf{F}_L}$, outperforming learnable pooling and per-view/frustum student $\cls$ tokens. This suggests that scene-level alignment benefits from surfacing the most salient responses, and that \textit{global} $\cls$ distillation best complements the local multi-layer distillation, making the hierarchical distillation design of \modelname{} the most effective configuration. A more extensive analysis is found in~\cref{sec:add_ablations}.


\paragraph{\textbf{Diffusion \vs Occupancy Decoders.}}
To isolate temporal occupancy diffusion, we replace it with deterministic occupancy decoders: a simple decoder~\cite{Liu2023occ4cast} and ALSO~\cite{also}. \Cref{tab:ablation_design_components_compact} shows that diffusion performs best, supporting our use of a joint generative scene objective rather than independent per-voxel occupancy prediction which may miss global structural coherence. Through iterative denoising, diffusion provides a label-free spatiotemporal pre-training signal that better captures complex 3D-scene structure and multi-modal evolution through coarse-to-fine refinement. As shown in~\cref{sec:main-cross-modal-results,sec:multi-task}, this auxiliary yields gains across all evaluated tasks, complementing hierarchical distillation.

\section{Conclusion}
We introduced \modelname{}, a self-supervised LiDAR \pretraining framework that leverages the hierarchical information in VFMs to address limitations of prior camera-to-LiDAR distillation. With hierarchical distillation, combining multi-layer and global context distillation, \modelname{} captures how semantic features evolve across VFM layers, yielding stronger representations. \modelname{} also uses temporal occupancy diffusion as an auxiliary task, encouraging features to encode geometric structure and temporal dynamics. The result is an informative 3D representation that captures both the semantic \emph{what} and the geometric \emph{where}. Across segmentation, detection, semantic occupancy, robustness, and scene flow, \modelname{} improves over prior camera--LiDAR distillation methods. A limitation of point--pixel distillation is sensitivity to LiDAR--camera misalignment. \modelname{} however partly mitigates this with alignment-free global context distillation and temporal diffusion. Future work could explore how to optimize layer-to-layer distillation through learned matching strategies and dynamic loss weighting, and extend diffusion beyond BEV (\eg, via latent diffusion~\cite{rombach2022high}) to better preserve 3D geometry and temporal evolution.

\section*{Acknowledgements}
We are grateful to Mohammad Nazari, Ajinkya Khoche, Qingwen Zhang, and John Folkesson for insightful discussions and helpful feedback on the proposed method. This work was partially supported by the Wallenberg AI, Autonomous Systems and Software Program (WASP) funded by the Knut and Alice Wallenberg Foundation. This work has been performed with support from Sweden's Innovation Agency (VINNOVA) through the Strategic Vehicle Research and Innovation Programme (FFI), grant no. 2024-03640, and from TRATON AB. The computations and data handling were enabled by the Berzelius resource provided by the Knut and Alice Wallenberg Foundation at the National Supercomputer Centre.


{\scriptsize
\bibliographystyle{eccv_2026_template/splncs04}
\bibliography{references_fixed}
}
\clearpage

\appendix
\setcounter{page}{1}
\counterwithin*{table}{section}
\counterwithin*{figure}{section}
\counterwithin*{equation}{section}
\renewcommand{\thetable}{\thesection\arabic{table}}
\renewcommand{\thefigure}{\thesection\arabic{figure}}
\renewcommand{\theequation}{\thesection\arabic{equation}}
\section*{Supplementary Material}
\addtocontents{toc}{\protect\setcounter{tocdepth}{2}}
\makeatletter
\renewcommand\tableofcontents{\@starttoc{toc}}
\makeatother
\tableofcontents

\section{Additional Quantitative Results}
\label{sec:additional_results}

The baseline results displayed in this section correspond to the ones reported by the respective authors unless indicated otherwise. We also show the results for \textbf{\modelname{}$^\dagger$}, our method excluding the diffusion auxiliary task during \pretraining. In all the tables, the best/$2^{nd}$ best results are \textbf{bold}/\underline{underline}. We exclude \modelname{}$^{\dagger}$ from this ranking to emphasize the difference between \modelname{} (our best model) and previous SOTA.

\subsection{Robustness Benchmark using Fully Fine-Tuned Backbones}

The main paper reports linear-probing results to isolate representation quality. In~\cref{tab:robust_full}, we complement these with robustness benchmarks under full fine-tuning (after \pretraining). We observe state-of-the-art performance in most configurations and the best average performance overall.

\begin{table*}[t]
\renewcommand{\arraystretch}{1} 
\setlength{\tabcolsep}{1.8pt}
\centering
\scriptsize
\caption{Performance comparison on robustness under weather corruption and sensor failure on nuScenes-C~\cite{kong2023robo3d} benchmark using \emph{full} fine-tuning with 100\% labeled data. All mCE ($\downarrow$), mRR ($\uparrow$), and mIoU ($\uparrow$) metrics are given in percentage (\%).}
\label{tab:robust_full}
\begin{tabular}{l cc cccccccccc}
\toprule
\multirow{2}{*}{Method} &  
{\multirow{2}{*}{mCE $\downarrow$}} & 
{\multirow{2}{*}{mRR $\uparrow$}} & 
\multicolumn{9}{c}{mIoU $\uparrow$} \\
\cmidrule(lr){4-12}
& & & {Fog}   & {Rain} & {Snow}  & {Blur} & {Beam} & {Cross} & {Echo} & {Sensor} & \textbf{Mean} \\ 
\midrule

{Random} &  112.20 & 72.57 & {62.96} & {70.65} & {55.48} & {51.71} & {62.01} & {31.56} & {59.64} & {39.41} & 54.18 \\  \hline

{SLidR~\cite{slidr}} &  106.08 & 75.99 & {65.41} & {72.31} & {56.01} & {56.07} & {62.87} & {41.94} & {61.16} & {38.90}  & 56.83 \\ \hline

{Seal~\cite{seal}}  &  92.63 & 83.08 & {72.66} & {74.31} & 66.22 & \cellcolor{white!90!ForestGreen}{\underline{66.14}} & {65.96} & {57.44} & {59.87} & {39.85}  & 62.81 \\ \hline

{SuperFlow~\cite{superflow}} &  91.67 & 83.17 & {70.32} & \cellcolor{white!90!ForestGreen}\underline{75.77} & {65.41} & {61.05} & \cellcolor{white!90!ForestGreen}\underline{68.09} & {60.02} & {58.36} & {50.41}  & 63.68 \\ \hline

{Scalr~\cite{scalr}} &  99.86& 81.35& {69.23} & {72.47} & {60.67} & {55.11} & {64.07} & {56.93} & {60.34} & {46.11} & 60.62 \\ \hline

{LiMoE~\cite{limoe}} & \cellcolor{white!90!ForestGreen}\underline{88.43} & 83.28 & {71.10} & \cellcolor{white!65!ForestGreen}{\textbf{75.92}} & {65.66} & {63.86} & \cellcolor{white!65!ForestGreen}{\textbf{68.52}} & {60.78} & \cellcolor{white!90!ForestGreen}\underline{61.91} & {50.66} & 64.80 \\ \hline

{CleverDistiller~\cite{govindarajan2025cleverdistiller}} &  91.28 & \cellcolor{white!65!ForestGreen}\textbf{87.42} & \cellcolor{white!90!ForestGreen}\underline{72.83} & {70.91} & {66.03} & {60.18} & {66.85} & \cellcolor{white!90!ForestGreen}{\underline{68.04}} & {60.17} & \cellcolor{white!65!ForestGreen}{\textbf{53.63}} & \cellcolor{white!90!ForestGreen}\underline{64.83} \\ \hline

{LiMA~\cite{lima}} &  91.43 &82.57 &71.24 &73.38 & \cellcolor{white!90!ForestGreen}\underline{67.33} &\cellcolor{white!65!ForestGreen}\textbf{66.73} &66.71 &47.66 &61.72 &48.65 &62.93 \\ \hline

{\textbf{\modelname{}$^\dagger$}} & 90.05 & 82.27 & 73.48 & 72.78 & 68.97 & 59.48 & 65.39 &  66.97 & 62.60 &51.12& 65.10 \\

{\textbf{\modelname{}}} & \cellcolor{white!65!ForestGreen}\textbf{85.08} & \cellcolor{white!90!ForestGreen}\underline{85.13} &\cellcolor{white!65!ForestGreen}\textbf{74.77}  & 74.75 &\cellcolor{white!65!ForestGreen}\textbf{68.06} & 63.45&  66.69&  \cellcolor{white!65!ForestGreen}\textbf{69.17}&\cellcolor{white!65!ForestGreen}\textbf{63.25} &\cellcolor{white!90!ForestGreen}\underline{51.48} & \cellcolor{white!65!ForestGreen}\textbf{67.36} \\

\arrayrulecolor{black}
\bottomrule
\end{tabular}
\end{table*}

\subsection{3D Semantic Segmentation}

As observed in Table~\ref{tab:classwise}, \modelname{} achieves the highest mean Intersection over Union (mIoU) across all evaluated Vision Transformer backbones (ViT-S, ViT-B, and ViT-L). The improvements over the second-best, CleverDistiller~\cite{govindarajan2025cleverdistiller}, are primarily concentrated in geometrically diverse classes and static road features. Notably, \modelname{} yields substantial gains in the classes \textit{traffic cone}, \textit{truck}, and \textit{other flat} surfaces. Furthermore, both the ViT-B and ViT-L architectures overcome the $0.0$ IoU failure state for the \textit{bicycle} class.



\begin{table*}[!ht]
\setlength{\tabcolsep}{3.0pt}
\renewcommand{\arraystretch}{1.15}
\centering
\caption{3D semantic segmentation. Per-class IoU scores (\%) of state-of-the-art \pretraining methods \pretrained and fine-tuned on nuScenes with 1\% annotations.}

\newlength{\clsheadheight}
\setlength{\clsheadheight}{2.0cm}

\newcommand{\clshead}[2]{%
    \parbox[b][\clsheadheight][b]{1.75em}{%
        \centering
        \rotatebox{90}{#1}\par
        \colorbox{#2}{\textcolor{#2}{\rule{1.25ex}{1.25ex}}}%
    }%
}
\resizebox{\textwidth}{!}{%
\begin{tabular}{l | c | r*{16}{>{\centering\arraybackslash}p{1.75em}}}

\toprule
\textbf{Method} &
\textbf{mIoU} &
\clshead{barrier}{red!80} &
\clshead{bicycle}{green!70} &
\clshead{bus}{yellow!85} &
\clshead{car}{blue!80} &
\clshead{const.\ veh.}{orange!85} &
\clshead{motorcycle}{purple!70} &
\clshead{pedestrian}{pink!55} &
\clshead{traffic cone}{orange!90!red} &
\clshead{trailer}{gray!45} &
\clshead{truck}{brown!70} &
\clshead{drive surf.}{cyan!45} &
\clshead{other flat}{brown!45} &
\clshead{sidewalk}{gray!25} &
\clshead{terrain}{olive!70} &
\clshead{manmade}{blue!15!black!35} &
\clshead{vegetation}{green!60!black} \\
\midrule

\multicolumn{18}{c}{{\color[HTML]{2FC061} \textbf{ViT-S}}} \\

SLidR & 41.2 & 0.0 & 0.0 & 26.6 & 72.0 & 12.4 & 15.8 & 51.4 & 22.9 & 11.7 & 35.3 & 92.9 & 36.3 & 58.7 & 63.6 & 81.2 & 82.3 \\

Seal  & 44.3 & 20.0 & 0.0 & 19.4 & 74.7 & 10.6 & \cellcolor{white!80!SpringGreen}\underline{45.7} & 60.3 & \cellcolor{white!80!SpringGreen}\underline{29.2} & 17.4 & 38.1 & 93.2 & 26.0 & 58.8 & 64.5 & 81.9 & 81.9 \\ 

ScaLR & 53.0 & 58.4 & 0.0 & 64.9 & \cellcolor{white!80!SpringGreen}\underline{83.3} & 10.5 & 20.7 & 59.6 & 24.6 & 26.4 & 55.8 & 93.9 & 51.2 & 63.4 & 68.5 & 83.7 & 83.1 \\ 

SuperFlow & 47.8 & 38.2 & \cellcolor{white!80!SpringGreen}\underline{1.8} & 25.8 & 79.0 & 15.3 & 43.6 & 60.3 & 0.0 & 28.4 & 55.4 & 93.7 & 28.8 & 59.1 & 59.9 & 83.5 & 83.1 \\

LiMoE& 49.6 & 39.9 & \cellcolor{white!40!SpringGreen}\textbf{4.6} & 27.3 & 80.2 & \cellcolor{white!80!SpringGreen}\underline{17.1} & 45.4 & 61.2 & 6.2 & 29.5 & 58.4 & 94.0 & 34.2 & 62.3 & 64.6 & 84.1 & \cellcolor{white!40!SpringGreen}\textbf{84.5} \\ 

CleverD &  \cellcolor{white!80!SpringGreen}\underline{56.9} & \cellcolor{white!80!SpringGreen}\underline{60.2} & 0.0 & \cellcolor{white!80!SpringGreen}\underline{71.2} & \cellcolor{white!40!SpringGreen}\textbf{84.9} & 11.8 & \cellcolor{white!40!SpringGreen}\textbf{47.1} & \cellcolor{white!40!SpringGreen}\textbf{65.1} & 26.6 & \cellcolor{white!80!SpringGreen}\underline{31.4} & \cellcolor{white!80!SpringGreen}\underline{61.7} & \cellcolor{white!80!SpringGreen}\underline{94.2} & \cellcolor{white!80!SpringGreen}\underline{54.1} & \cellcolor{white!80!SpringGreen}\underline{64.9} & \cellcolor{white!80!SpringGreen}\underline{69.5} & \cellcolor{white!80!SpringGreen}\underline{84.2} & 83.5 \\

\textbf{\modelname{}$^\dagger$} & 57.8 & 65.0 & 0.5 & 73.0 & 77.0 & 11.4 & 47.1 & 61.9 & 33.0 & 34.4 & 61.2 & 94.9 & 60.5 & 66.9 & 70.1 & 84.6 & 83.4 \\

\textbf{\modelname{}} & \cellcolor{white!40!SpringGreen}\textbf{59.4} & \cellcolor{white!40!SpringGreen}\textbf{69.2} & 0.0 & \cellcolor{white!40!SpringGreen}\textbf{76.6} & 81.9 & \cellcolor{white!40!SpringGreen}\textbf{30.4} & 12.1 & \cellcolor{white!80!SpringGreen}\underline{62.4} & \cellcolor{white!40!SpringGreen}\textbf{42.2} & \cellcolor{white!40!SpringGreen}\textbf{39.0} & \cellcolor{white!40!SpringGreen}\textbf{68.7} & \cellcolor{white!40!SpringGreen}\textbf{95.6} & \cellcolor{white!40!SpringGreen}\textbf{62.3} & \cellcolor{white!40!SpringGreen}\textbf{68.1} & \cellcolor{white!40!SpringGreen}\textbf{71.7} & \cellcolor{white!40!SpringGreen}\textbf{85.5} & \cellcolor{white!80!SpringGreen}\underline{84.1} \\

\midrule
\multicolumn{18}{c}{{\color{Apricot} \textbf{ViT-B}}} \\

PPKT & 40.9 & 0.0 & 0.0 & 24.5 & 73.5 & 12.2 & 7.0 & 51.0 & 13.5 & 15.4 & 36.3 & 93.1 & 40.4 & 59.2 & 63.5 & 81.7 & 82.2 \\ 

SLidR & 41.6 & 0.0 & 0.0 & 26.7 & 73.4 & 10.3 & 16.9 & 51.3 & 23.3 & 12.7 & 38.1 & 93.0 & 37.7 & 58.8 & 63.4 & 81.6 & 82.7 \\ 

Seal  & 46.0 & 43.0 & 0.0 & 26.7 & 81.3 & 9.9 & 41.3 & 56.2 & 0.0 & 21.7 & 51.6 & 93.6 & 42.3 & 62.8 & 64.7 & 82.6 & 82.7 \\ 

SuperFlow & 48.1 & 39.1 & 0.9 & 30.0 & 80.7 & 10.3 & 47.1 & 59.5 & 5.1 & 27.6 & 55.4 & 93.7 & 29.1 & 61.1 & 63.5 & 82.7 & 83.6 \\ 

ScaLR & 55.8 & 60.8 & 0.0 & 69.7 & \cellcolor{white!80!Apricot}\underline{85.1} & 20.4 & 30.6 & 60.8 & 27.8 & 29.5 & 58.1 & 94.2 & 55.6 & 64.3 & 69.3 & 84.1 & 83.1 \\ 

LiMoE& 50.2 & 41.5 & \cellcolor{white!80!Apricot}\underline{3.8} & 32.2 & 81.7 & 12.9 & \cellcolor{white!80!Apricot}\underline{49.3} & 61.1 & 7.3 & 29.3 & 57.8 & 94.2 & 35.1 & 62.9 & 65.4 & 84.0 & \cellcolor{white!80!Apricot}\underline{84.8} \\

CleverD& \cellcolor{white!80!Apricot}\underline{59.8} & \cellcolor{white!80!Apricot}\underline{61.8} & 0.0 & \cellcolor{white!80!Apricot}\underline{72.8} & \cellcolor{white!51!Apricot}\textbf{85.7} & \cellcolor{white!80!Apricot}\underline{35.2} & \cellcolor{white!51!Apricot}\textbf{49.6} & \cellcolor{white!80!Apricot}\underline{65.4} & \cellcolor{white!80!Apricot}\underline{33.6} & \cellcolor{white!80!Apricot}\underline{32.9} & \cellcolor{white!80!Apricot}\underline{62.6} & \cellcolor{white!80!Apricot}\underline{94.5} & \cellcolor{white!80!Apricot}\underline{59.5} & \cellcolor{white!80!Apricot}\underline{65.6} & \cellcolor{white!80!Apricot}\underline{69.5} & \cellcolor{white!80!Apricot}\underline{84.7} & 83.5\\

\textbf{\modelname{}$^\dagger$} & 61.0 & 69.4 & 4.9 & 74.0 & 81.9 & 32.0 & 25.6 & 67.4 & 44.6 & 38.4 & 68.3 & 95.2 & 64.4 & 69.2 & 72.0 & 84.7 & 84.1 \\

\textbf{\modelname{}} & \cellcolor{white!51!Apricot}\textbf{62.7} & \cellcolor{white!51!Apricot}\textbf{69.7} & \cellcolor{white!51!Apricot}\textbf{7.4} & \cellcolor{white!51!Apricot}\textbf{78.3} & \cellcolor{white!51!Apricot}\textbf{85.7} & \cellcolor{white!51!Apricot}\textbf{41.6} & 11.1 & \cellcolor{white!51!Apricot}\textbf{67.9} & \cellcolor{white!51!Apricot}\textbf{47.5} & \cellcolor{white!51!Apricot}\textbf{39.3} & \cellcolor{white!51!Apricot}\textbf{75.5} & \cellcolor{white!51!Apricot}\textbf{95.9} & \cellcolor{white!51!Apricot}\textbf{67.5} & \cellcolor{white!51!Apricot}\textbf{71.6} & \cellcolor{white!51!Apricot}\textbf{73.3} & \cellcolor{white!51!Apricot}\textbf{86.2} & \cellcolor{white!51!Apricot}\textbf{85.0} \\

\midrule
\multicolumn{18}{c}{{\color[HTML]{6434FC} \textbf{ViT-L}}} \\

PPKT & 42.1 & 0.0 & 0.0 & 24.4 & 78.8 & 15.1 & 9.2 & 54.2 & 14.3 & 12.9 & 39.1 & 92.9 & 37.8 & 59.8 & 64.9 & 82.3 & 83.6 \\

SLidR  & 42.8 & 0.0 & 0.0 & 23.9 & 78.8 & 15.2 & 20.9 & 55.0 & 28.0 & 17.4 & 41.4 & 92.2 & 41.2 & 58.0 & 64.0 & 81.8 & 82.7 \\

Seal  & 46.3 & 41.8 & 0.0 & 23.8 & 81.4 & 17.7 & 46.3 & 58.6 & 0.0 & 23.4 & 54.7 & 93.8 & 41.4 & 62.5 & 65.0 & 83.8 & 83.8 \\

SuperFlow & 50.0 & 44.5 & 0.9 & 22.4 & 80.8 & 17.1 & \cellcolor{white!85!Periwinkle}\underline{50.2} & 60.9 & 21.0 & 25.1 & 55.1 & 93.9 & 35.8 & 61.5 & 62.6 & 83.7 & 83.7 \\

LiMoE & 51.4 & 45.3 & \cellcolor{white!85!Periwinkle}\underline{4.1} & 25.3 & 82.2 & 18.4 & \cellcolor{white!55!Periwinkle}\textbf{52.5} & 61.8 & 22.3 & 26.4 & 56.2 & 94.3 & 37.6 & 63.3 & 63.9 & 84.4 & \cellcolor{white!55!Periwinkle}\textbf{85.0}\\

ScaLR& 55.8 & 65.0 & 0.0 & \cellcolor{white!85!Periwinkle}\underline{68.7} & 85.3 & 15.9 & 27.7 & 61.4 & 28.3 & 31.0 & 61.8 & 94.2 & 51.6 & 64.6 & 70.0 & 84.1 & 83.0  \\

CleverD& \cellcolor{white!85!Periwinkle}\underline{60.6} & \cellcolor{white!85!Periwinkle}\underline{65.5} & 0.0 & \cellcolor{white!55!Periwinkle}\textbf{72.3} & \cellcolor{white!55!Periwinkle}\textbf{88.3} & \cellcolor{white!85!Periwinkle}\underline{37.2} & 37.4 & \cellcolor{white!85!Periwinkle}\underline{65.1} & \cellcolor{white!85!Periwinkle}\underline{35.3} & \cellcolor{white!55!Periwinkle}\textbf{40.9} & \cellcolor{white!85!Periwinkle}\underline{71.9} & \cellcolor{white!85!Periwinkle}\underline{94.5} & \cellcolor{white!85!Periwinkle}\underline{55.8} & \cellcolor{white!85!Periwinkle}\underline{66.5} & \cellcolor{white!85!Periwinkle}\underline{70.6} & \cellcolor{white!85!Periwinkle}\underline{85.0} & 83.8\\

\textbf{\modelname{}$^\dagger$} & 61.6 & 70.4 & 5.8 & 69.0 & 82.9 & 27.2 & 58.7 & 52.0 & 46.0 & 36.2 & 66.8 & 95.5 & 64.8 & 69.6 & 72.4 & 85.1 & 83.2 \\

\textbf{\modelname{}} & \cellcolor{white!55!Periwinkle}\textbf{62.8} & \cellcolor{white!55!Periwinkle}\textbf{68.3} & \cellcolor{white!55!Periwinkle}\textbf{7.6} & 67.9 & \cellcolor{white!85!Periwinkle}\underline{88.0} & \cellcolor{white!55!Periwinkle}\textbf{38.6} & 35.9 & \cellcolor{white!55!Periwinkle}\textbf{66.8} & \cellcolor{white!55!Periwinkle}\textbf{44.5} & \cellcolor{white!85!Periwinkle}\underline{39.3} & \cellcolor{white!55!Periwinkle}\textbf{72.6} & \cellcolor{white!55!Periwinkle}\textbf{95.8} & \cellcolor{white!55!Periwinkle}\textbf{64.3} & \cellcolor{white!55!Periwinkle}\textbf{70.8} & \cellcolor{white!55!Periwinkle}\textbf{73.4} & \cellcolor{white!55!Periwinkle}\textbf{85.9} & \cellcolor{white!85!Periwinkle}\underline{84.6} \\

\bottomrule
\end{tabular}
}
\label{tab:classwise}
\end{table*}


\subsection{Semantic Occupancy Prediction}

\begin{table*}[!ht]
\centering
\caption{3D/4D semantic occupancy IoU results with a frozen \pretrained backbone and a trainable occupancy decoder head. We compare cross-modal distillation \pretraining baselines against \modelname{} using occupancy targets and inputs produced by the Occ4Cast~\cite{Liu2023occ4cast} pre-processing pipeline. The results are shown for different time horizons. The models input up to two sweeps ($t_{-1}$, $t_{0}$) and predict occupancy up to and including $t_T$ ($t_0$, ..., $t_T$). Notation -1/T indicates two input frames starting from $t_{-1}$ and a prediction horizon up to and including $t_T$. Notation 0/0 indicates input and target are both at frame $t_0$. The 4D metrics are averaged over frame-wise metrics. The aggregate metric corresponds to the semantic class averaged mIoU. Best/second-best are in \textbf{bold}/\underline{underlined}.}
\label{tab:4docc_sem_temp}

\setlength{\tabcolsep}{3.0pt}
\renewcommand{\arraystretch}{1.15}

\setlength{\clsheadheight}{2.25cm}

\newcommand{\clshead}[2]{%
    \parbox[b][\clsheadheight][b]{1.75em}{%
        \centering
        \rotatebox{90}{#1}\par
        \colorbox{#2}{\textcolor{#2}{\rule{1.25ex}{1.25ex}}}%
    }%
}

\resizebox{\textwidth}{!}{%
\begin{tabular}{c l|r|*{16}{>{\centering\arraybackslash}p{1.75em}}}
\toprule
 & \textbf{Method} &
\textbf{mIoU} &
\clshead{barrier}{red!80} &
\clshead{bicycle}{green!70} &
\clshead{bus}{yellow!85} &
\clshead{car}{blue!80} &
\clshead{const.\ veh.}{orange!85} &
\clshead{motorcycle}{purple!70} &
\clshead{pedestrian}{pink!55} &
\clshead{traffic cone}{orange!90!red} &
\clshead{trailer}{gray!45} &
\clshead{truck}{brown!70} &
\clshead{drive surf.}{cyan!45} &
\clshead{other flat}{brown!45} &
\clshead{sidewalk}{gray!25} &
\clshead{terrain}{olive!70} &
\clshead{manmade}{blue!15!black!35} &
\clshead{vegetation}{green!60!black} \\
\midrule
\multirow{6}{*}{\rotatebox{90}{\textbf{3D (0/0)}}} 

& Random 
& 10.0 & 1.7 & 0.0 & 0.0 & 16.8 & 0.0 & 0.0 & 0.0 & 0.0 & 8.3 & 7.4 & 34.2 & 6.9 & 11.8 & 17.7 & 31.8 & 23.3 \\

& ScaLR 
& 14.9 & 11.1 & 0.0 & 1.9 & 24.7 & 1.0 & \cellcolor{white!80!Apricot}\underline{0.6} & 10.6 & \cellcolor{white!80!Apricot}\underline{1.4} & \cellcolor{white!80!Apricot}\underline{13.9} & 16.1 & \cellcolor{white!80!Apricot}\underline{36.9} & \cellcolor{white!80!Apricot}\underline{16.2} & 18.9 & 22.2 & 36.0 & \cellcolor{white!80!Apricot}\underline{26.1} \\

& SuperFlow & 11.8 & 2.2 & 0.0 & 0.3 & 21.3 & 0.2 & 0.0 & 0.6 & 0.5 & 12.7 & 12.7 & 35.5 & 11.1 & 13.3 & 18.6 & 34.2 & 25.0 \\

& CleverDistiller 
& \cellcolor{white!80!Apricot}\underline{15.7} & \cellcolor{white!80!Apricot}\underline{14.4} & 0.0 & \cellcolor{white!80!Apricot}\underline{7.8} & \cellcolor{white!80!Apricot}\underline{25.0} & \cellcolor{white!80!Apricot}\underline{2.5} & 0.5 & \cellcolor{white!80!Apricot}\underline{14.2} & 1.2 & 13.0 & \cellcolor{white!80!Apricot}\underline{16.2} & 36.6 & 15.6 & \cellcolor{white!80!Apricot}\underline{19.0} & \cellcolor{white!80!Apricot}\underline{22.5} & \cellcolor{white!80!Apricot}\underline{36.2} & \cellcolor{white!80!Apricot}\underline{26.1} \\

& \textbf{\modelname{}$^\dagger$}
& 17.9 & 18.1 & 0.0 & 16.5 & 25.8 & 5.5 & 4.5 & 18.4 & 3.0 & 15.1 & 18.9 & 36.9 & 16.8 & 20.2 & 23.4 & 37.0 & 26.6 \\
\multirow{-6}{*}{\rotatebox{90}{\textbf{3D (0/0)}}} &



\textbf{\modelname{}} 
& \cellcolor{white!51!Apricot}\textbf{18.8} & \cellcolor{white!51!Apricot}\textbf{19.9} & 0.0 & \cellcolor{white!51!Apricot}\textbf{18.1} & \cellcolor{white!51!Apricot}\textbf{26.3} & \cellcolor{white!51!Apricot}\textbf{11.2} & \cellcolor{white!51!Apricot}\textbf{6.4} & \cellcolor{white!80!Apricot}\underline{18.2} & \cellcolor{white!51!Apricot}\textbf{5.3} & 13.8 & \cellcolor{white!51!Apricot}\textbf{20.6} & \cellcolor{white!51!Apricot}\textbf{37.1} & 15.9 & \cellcolor{white!51!Apricot}\textbf{20.5} & \cellcolor{white!51!Apricot}\textbf{23.5} & \cellcolor{white!51!Apricot}\textbf{37.1} & \cellcolor{white!51!Apricot}\textbf{26.6} \\
\midrule
\multirow{6}{*}{\rotatebox{90}{\textbf{3D (-1/0)}}} 

& Random 
& 10.6 & 1.4 & 0.0 & 0.0 & 16.0 & 0.0 & 0.0 & 0.0 & 0.0 & 6.7 & 6.9 & 38.8 & 6.4 & 13.5 & 20.0 & 34.5 & 25.4 \\

& ScaLR 
& 16.2 & 9.4 & 0.0 & 4.8 & 25.3 & \cellcolor{white!80!Apricot}\underline{2.3} & 0.2 & 12.0 & \cellcolor{white!80!Apricot}\underline{3.0} & \cellcolor{white!80!Apricot}\underline{14.1} & \cellcolor{white!80!Apricot}\underline{16.9} & 41.8 & \cellcolor{white!80!Apricot}\underline{14.2} & \cellcolor{white!80!Apricot}\underline{20.9} & \cellcolor{white!80!Apricot}\underline{26.1} & 38.9 & 28.9 \\

& SuperFlow & 12.5 & 3.0 & 0.0 & 0.5 & 20.7 & 0.2 & 0.0 & 1.6 & 0.7 & 12.4 & 11.7 & 39.5 & 9.5 & 14.4 & 21.3 & 37.0 & 27.5 \\

& CleverDistiller 
& \cellcolor{white!80!Apricot}\underline{16.5} & \cellcolor{white!80!Apricot}\underline{11.5} & 0.0 & \cellcolor{white!80!Apricot}\underline{5.4} & \cellcolor{white!80!Apricot}\underline{25.7} & 1.7 & \cellcolor{white!80!Apricot}\underline{0.6} & \cellcolor{white!80!Apricot}\underline{16.0} & 2.3 & 13.9 & 16.8 & \cellcolor{white!80!Apricot}\underline{41.9} & 12.9 & 20.8 & 26.0 & \cellcolor{white!80!Apricot}\underline{39.2} & \cellcolor{white!80!Apricot}\underline{29.0} \\

& \textbf{\modelname{}$^\dagger$} 
& 19.0 & 17.2 & 0.0 & 16.7 & 26.6 & 7.2 & 2.1 & 19.7 & 3.6 & 15.3 & 19.1 & 42.1 & 15.5 & 22.6 & 26.7 & 39.8 & 29.2 \\
\multirow{-6}{*}{\rotatebox{90}{\textbf{3D (-1/0)}}} &



\textbf{\modelname{}} 
& \cellcolor{white!51!Apricot}\textbf{20.0} & \cellcolor{white!51!Apricot}\textbf{18.8} & 0.0 & \cellcolor{white!51!Apricot}\textbf{17.3} & \cellcolor{white!51!Apricot}\textbf{27.3} & \cellcolor{white!51!Apricot}\textbf{10.9} & \cellcolor{white!51!Apricot}\textbf{6.0} & \cellcolor{white!51!Apricot}\textbf{19.8} & \cellcolor{white!51!Apricot}\textbf{5.3} & \cellcolor{white!80!Apricot}\underline{14.5} & \cellcolor{white!51!Apricot}\textbf{20.7} & \cellcolor{white!51!Apricot}\textbf{42.1} & \cellcolor{white!51!Apricot}\textbf{17.0} & \cellcolor{white!51!Apricot}\textbf{23.0} & \cellcolor{white!51!Apricot}\textbf{27.3} & \cellcolor{white!51!Apricot}\textbf{39.9} & \cellcolor{white!51!Apricot}\textbf{29.3} \\
\midrule
\multirow{6}{*}{\rotatebox{90}{\textbf{4D (-1/1)}}} 

& Random 
& 10.3 & 1.1 & 0.0 & 0.0 & 12.6 & 0.0 & 0.0 & 0.0 & 0.0 & 6.4 & 5.6 & 38.6 & 7.2 & 13.2 & 20.3 & 34.5 & 25.4 \\

& ScaLR 
& 14.9 & 9.6 & 0.0 & 1.8 & 21.3 & 1.2 & 0.0 & 5.1 & 1.0 & \cellcolor{white!80!Apricot}\underline{14.2} & 13.3 & \cellcolor{white!80!Apricot}\underline{42.0} & \cellcolor{white!80!Apricot}\underline{14.1} & 20.8 & \cellcolor{white!80!Apricot}\underline{26.2} & 39.0 & 28.9 \\

& SuperFlow & 11.9 & 2.6 & 0.0 & 0.5 & 16.8 & 0.2 & 0.0 & 0.2 & 0.3 & 11.2 & 10.0 & 39.7 & 9.0 & 14.2 & 21.0 & 36.8 & 27.3 \\

& CleverDistiller 
& \cellcolor{white!80!Apricot}\underline{15.5} & \cellcolor{white!80!Apricot}\underline{11.4} & 0.0 & \cellcolor{white!80!Apricot}\underline{4.1} & \cellcolor{white!80!Apricot}\underline{21.8} & \cellcolor{white!80!Apricot}\underline{2.0} & \cellcolor{white!80!Apricot}\underline{0.2} & \cellcolor{white!80!Apricot}\underline{8.0} & \cellcolor{white!80!Apricot}\underline{1.9} & 13.8 & \cellcolor{white!80!Apricot}\underline{14.0} & 41.9 & 13.6 & \cellcolor{white!80!Apricot}\underline{21.1} & \cellcolor{white!80!Apricot}\underline{26.2} & \cellcolor{white!80!Apricot}\underline{39.1} & \cellcolor{white!80!Apricot}\underline{29.0} \\

& \textbf{\modelname{}$^\dagger$} 
& 17.3 & 14.8 & 0.0 & 10.9 & 22.4 & 5.0 & 1.7 & 10.2 & 2.9 & 14.5 & 16.0 & 42.1 & 16.6 & 23.1 & 26.9 & 39.7 & 29.3 \\
\multirow{-6}{*}{\rotatebox{90}{\textbf{4D (-1/1)}}} &



\textbf{\modelname{}} 
& \cellcolor{white!51!Apricot}\textbf{18.4} & \cellcolor{white!51!Apricot}\textbf{19.0} & 0.0 & \cellcolor{white!51!Apricot}\textbf{15.7} & \cellcolor{white!51!Apricot}\textbf{23.3} & \cellcolor{white!51!Apricot}\textbf{9.6} & \cellcolor{white!80!Apricot}\underline{1.1} & \cellcolor{white!80!Apricot}\underline{10.1} & \cellcolor{white!51!Apricot}\textbf{5.1} & 13.8 & \cellcolor{white!51!Apricot}\textbf{18.0} & \cellcolor{white!51!Apricot}\textbf{42.4} & \cellcolor{white!51!Apricot}\textbf{16.8} & \cellcolor{white!51!Apricot}\textbf{23.2} & \cellcolor{white!51!Apricot}\textbf{27.2} & \cellcolor{white!51!Apricot}\textbf{39.9} & \cellcolor{white!80!Apricot}\underline{29.2} \\
\midrule
\multirow{6}{*}{\rotatebox{90}{\textbf{4D (-1/5)}}} 

& Random 
& 9.6 & 0.8 & 0.0 & 0.0 & 9.9 & 0.0 & 0.0 & 0.0 & 0.0 & 5.4 & 3.5 & 38.1 & 3.1 & 12.7 & 20.2 & 34.1 & 25.4 \\

& ScaLR 
& 13.5 & 7.2 & 0.0 & 0.2 & 17.3 & 0.4 & 0.0 & 1.1 & 0.7 & \cellcolor{white!80!Apricot}\underline{11.5} & 11.0 & \cellcolor{white!80!Apricot}\underline{41.8} & 11.7 & \cellcolor{white!80!Apricot}\underline{20.6} & \cellcolor{white!80!Apricot}\underline{25.8} & 38.6 & 28.6 \\

& SuperFlow & 11.2 & 2.5 & 0.0 & 0.0 & 13.7 & 0.1 & 0.0 & 0.1 & 0.1 & 9.6 & 7.5 & 39.6 & 6.9 & 14.2 & 21.3 & 36.7 & 27.2 \\

& CleverDistiller 
& \cellcolor{white!80!Apricot}\underline{14.0} & \cellcolor{white!80!Apricot}\underline{9.2} & 0.0 & \cellcolor{white!80!Apricot}\underline{1.8} & \cellcolor{white!80!Apricot}\underline{17.6} & \cellcolor{white!80!Apricot}\underline{0.7} & 0.0 & \cellcolor{white!80!Apricot}\underline{1.6} & \cellcolor{white!80!Apricot}\underline{1.3} & \cellcolor{white!80!Apricot}\underline{11.5} & \cellcolor{white!80!Apricot}\underline{11.8} & 41.5 & \cellcolor{white!80!Apricot}\underline{13.2} & 20.2 & \cellcolor{white!80!Apricot}\underline{25.8} & \cellcolor{white!80!Apricot}\underline{39.1} & \cellcolor{white!80!Apricot}\underline{28.8} \\

& \textbf{\modelname{}$^\dagger$} 
& 15.4 & 12.4 & 0.0 & 4.7 & 18.6 & 2.9 & 0.1 & 3.6 & 2.1 & 12.5 & 13.1 & 42.2 & 14.6 & 22.9 & 27.0 & 39.8 & 29.3 \\
\multirow{-6}{*}{\rotatebox{90}{\textbf{4D (-1/5)}}} &



\textbf{\modelname{}} 
& \cellcolor{white!51!Apricot}\textbf{16.3} & \cellcolor{white!51!Apricot}\textbf{16.1} & 0.0 & \cellcolor{white!51!Apricot}\textbf{6.4} & \cellcolor{white!51!Apricot}\textbf{18.6} & \cellcolor{white!51!Apricot}\textbf{8.1} & \cellcolor{white!51!Apricot}\textbf{2.0} & \cellcolor{white!80!Apricot}\underline{3.3} & \cellcolor{white!51!Apricot}\textbf{2.9} & \cellcolor{white!80!Apricot}\underline{12.2} & \cellcolor{white!51!Apricot}\textbf{14.3} & \cellcolor{white!80!Apricot}\underline{42.1} & \cellcolor{white!51!Apricot}\textbf{16.4} & \cellcolor{white!80!Apricot}\underline{22.7} & \cellcolor{white!51!Apricot}\textbf{27.0} & \cellcolor{white!51!Apricot}\textbf{39.9} & \cellcolor{white!80!Apricot}\underline{29.2} \\
\midrule
\multirow{6}{*}{\rotatebox{90}{\textbf{4D (-1/10)}}} 

& Random 
& 9.4 & 0.4 & 0.0 & 0.0 & 8.6 & 0.0 & 0.0 & 0.0 & 0.0 & 4.9 & 2.8 & 38.5 & 3.9 & 12.2 & 19.6 & 34.3 & 25.4 \\

& ScaLR 
& \cellcolor{white!80!Apricot}\underline{13.5} & \cellcolor{white!80!Apricot}\underline{7.4} & 0.0 & 0.1 & 15.4 & 0.8 & 0.0 & 0.6 & \cellcolor{white!80!Apricot}\underline{0.8} & \cellcolor{white!80!Apricot}\underline{11.3} & 9.9 & \cellcolor{white!51!Apricot}\textbf{42.0} & \cellcolor{white!80!Apricot}\underline{12.5} & 20.8 & \cellcolor{white!80!Apricot}\underline{26.1} & 38.7 & 28.7 \\

& SuperFlow & 10.8 & 1.6 & 0.0 & 0.0 & 11.8 & 0.0 & 0.0 & 0.0 & 0.0 & 7.6 & 6.7 & 39.6 & 7.0 & 14.2 & 20.9 & 36.6 & 27.2 \\

& CleverDistiller 
& 13.4 & 6.7 & 0.0 & \cellcolor{white!80!Apricot}\underline{1.5} & \cellcolor{white!80!Apricot}\underline{15.6} & \cellcolor{white!80!Apricot}\underline{1.0} & 0.0 & \cellcolor{white!80!Apricot}\underline{0.7} & 0.2 & 9.9 & \cellcolor{white!80!Apricot}\underline{10.8} & \cellcolor{white!80!Apricot}\underline{41.8} & 11.7 & \cellcolor{white!80!Apricot}\underline{21.0} & \cellcolor{white!80!Apricot}\underline{26.1} & \cellcolor{white!80!Apricot}\underline{38.9} & \cellcolor{white!80!Apricot}\underline{28.8} \\

& \textbf{\modelname{}$^\dagger$} 
& 14.8 & 11.7 & 0.0 & 3.6 & 16.5 & 2.5 & 0.0 & 2.1 & 0.3 & 10.7 & 11.8 & 42.4 & 16.5 & 23.2 & 26.8 & 39.8 & 29.2 \\
\multirow{-6}{*}{\rotatebox{90}{\textbf{4D (-1/10)}}} &



\textbf{\modelname{}} 
& \cellcolor{white!51!Apricot}\textbf{15.4} & \cellcolor{white!51!Apricot}\textbf{14.8} & 0.0 & \cellcolor{white!51!Apricot}\textbf{4.4} & \cellcolor{white!80!Apricot}\underline{16.4} & \cellcolor{white!51!Apricot}\textbf{6.8} & 0.0 & \cellcolor{white!80!Apricot}\underline{1.7} & \cellcolor{white!51!Apricot}\textbf{1.2} & \cellcolor{white!51!Apricot}\textbf{11.4} & \cellcolor{white!51!Apricot}\textbf{13.2} & \cellcolor{white!51!Apricot}\textbf{42.0} & \cellcolor{white!51!Apricot}\textbf{16.0} & \cellcolor{white!51!Apricot}\textbf{22.5} & \cellcolor{white!51!Apricot}\textbf{27.1} & \cellcolor{white!51!Apricot}\textbf{39.9} & \cellcolor{white!51!Apricot}\textbf{29.3} \\
\bottomrule
\end{tabular}%
}
\end{table*}

Table~\ref{tab:4docc_sem_temp} shows 3D and 4D semantic occupancy prediction with a frozen \pretrained backbone and a trainable decoder (details on model, data, training, and metrics can be found in~\cref{sec:implementationdetails}). For this task, we inherit the data pre-processing and target generation pipeline of Occ4cast~\cite{Liu2023occ4cast} to produce ground truth semantic occupancy in 3D and 4D using the nuScenes \cite{caesar2020nuscenes} lidarseg dataset. \modelname{} consistently achieves the best mIoU across all settings, and its margin over cross-modal distillation baselines persists as the prediction horizon increases. While performance naturally degrades with longer forecasting horizons, \modelname{} maintains a stable lead even at 10-step prediction, indicating stronger long-term spatiotemporal priors in the learned representation. 



Ablating the future BEV occupancy diffusion component (\modelname{} \vs \modelname{}$\dagger$) shows that it provides a consistent benefit beyond the 2D teacher distillation signal. Notably, during the LiDAR backbone \pretraining, the temporal diffusion objective conditions on LiDAR data from frames at time $(t_{-1}, t_0)$ and predicts occupancy at $t_1$, with $\Delta t = 0.5\,\mathrm{s}$. This supervision is directly aligned with the 4D (-1/1) setting in~\cref{tab:4docc_sem_temp} (corresponding to predicting $t_1$ conditioned on $(t_{-1}, t_0)$), where we also observe the largest ablation gain ($+1.1$ mIoU). Notation -1/T indicates two input frames starting from $t_{-1}$ and a prediction horizon up to and including $t_T$. The observed pattern suggests that the diffusion-based occupancy supervision complements the knowledge distillation by injecting training signals that promote learning temporal cues.


Importantly, the diffusion component disproportionately improves hard and rare occupied categories, such as \emph{construction vehicle}, \emph{traffic cone}, and \emph{motorcycle}, with additional gains on \emph{truck} and \emph{barrier}. At the longest horizon 4D (-1/10), the same effect remains visible. These gains are consistent with the intuition that diffusion supervision encourages geometric and temporal structure, which is particularly valuable for compact, occluded, or low-frequency objects. We also observe a mild trade-off where diffusion improves most difficult foreground classes but can slightly reduce performance on a small set of more ambiguous or boundary-heavy categories. For example, we observe a decrease for the classes \emph{trailer} and \emph{other flat} in 3D (0/0). In 4D (-1/10), \emph{other flat} and \emph{sidewalk} decreases. These drops are modest and do not affect the overall trend (mIoU increases across all horizons). Consistent with our discussion in~\cref{sec:multi-task}, we hypothesize that this reflects a shift toward geometry-aligned discrimination where our future BEV occupancy diffusion objective provides supervision for free--occupied boundaries and spatial extent, but it does not directly supervise semantic distinctions within occupied regions (\eg, differentiating surface categories such as \emph{sidewalk} \vs \emph{drivable surface}). While diffusion may capture scene-layout priors, it is less aligned with the \emph{semantic} boundaries required by IoU, as highlighted by~\cite{taleoftwofeatures}. Moreover, the BEV projection removes vertical cues that can be informative for certain semantic separations.

\paragraph{\textbf{OpenOccupancy Benchmark.}}

\modelname{} is also evaluated on the 3D semantic occupancy task introduced by OpenOccupancy~\cite{wang2023openocc} to compare against dedicated occupancy predictors. To this end, we apply our frozen backbone together with the same occupancy decoder head used in the Occ4cast setting. We use the standard OpenOccupancy target and output shape for a direct comparison with prior work. The frozen and \pretrained features of \modelname{}, coupled with a lightweight decoder head, perform competitively with 3D semantic occupancy frameworks~\cite{wang2023openocc, wang2024occgen}, as shown in~\cref{tab:openocc}. Notably, the LiDAR-only baselines are purpose-built occupancy predictors trained end-to-end for the task, with reported model sizes of 63M parameters for L-OpenOccupancy~\cite{wang2023openocc} and L-CONet~\cite{wang2023openocc}, and 62M for L-OccGen~\cite{wang2024occgen}. In contrast, \modelname{} relies on a frozen \pretrained sparse LiDAR backbone with 39.1M parameters and only a 0.4M \emph{trainable} occupancy-specific decoder head. The comparison therefore highlights that competitive occupancy performance can be obtained from a substantially lighter task-specific adaptation of a self-supervised \pretrained 3D representation.


\begin{table*}[!ht]
\setlength{\tabcolsep}{3.0pt}
\renewcommand{\arraystretch}{1.15}
\centering
\caption{3D semantic occupancy using the OpenOccupancy~\cite{wang2023openocc} benchmark. The table shows geometric IoU (binary occupancy) scores, mean IoU averaged over the semantic class scores, and the per-class IoU scores. Results from baselines are obtained directly from the published papers. Compared to dedicated occupancy networks~\cite{wang2023openocc,wang2024occgen}, frozen \modelname{} with a lightweight occupancy decoder head is able to compare competitively.}

\setlength{\clsheadheight}{2.0cm}

\newcommand{\clshead}[2]{%
    \parbox[b][\clsheadheight][b]{1.75em}{%
        \centering
        \rotatebox{90}{#1}\par
        \colorbox{#2}{\textcolor{#2}{\rule{1.25ex}{1.25ex}}}%
    }%
}
\resizebox{\textwidth}{!}{%
\begin{tabular}{l | c c | r*{16}{>{\centering\arraybackslash}p{1.75em}}}

\toprule
\textbf{Method} &
\textbf{IoU} &
\textbf{mIoU} &
\clshead{barrier}{red!80} &
\clshead{bicycle}{green!70} &
\clshead{bus}{yellow!85} &
\clshead{car}{blue!80} &
\clshead{const.\ veh.}{orange!85} &
\clshead{motorcycle}{purple!70} &
\clshead{pedestrian}{pink!55} &
\clshead{traffic cone}{orange!90!red} &
\clshead{trailer}{gray!45} &
\clshead{truck}{brown!70} &
\clshead{drive surf.}{cyan!45} &
\clshead{other flat}{brown!45} &
\clshead{sidewalk}{gray!25} &
\clshead{terrain}{olive!70} &
\clshead{manmade}{blue!15!black!35} &
\clshead{vegetation}{green!60!black} \\
\midrule

L-OpenOccupancy~\cite{wang2023openocc} & 
30.8 & 11.7 & 12.2 & 4.2 & 11.0 & 12.2 & \cellcolor{white!65!gray}\textbf{8.3} & 
4.4 & 8.7 & 4.0 & 8.4 & 10.3 & 23.5 & 16.0 & 14.9 & 15.7 & 15.0 & 17.9 \\

L-CONet~\cite{wang2023openocc} & 
\cellcolor{white!80!gray}\underline{30.9} & 
\cellcolor{white!80!gray}\underline{15.8} & 
\cellcolor{white!80!gray}\underline{17.5} & \cellcolor{white!65!gray}\textbf{5.2} & 
\cellcolor{white!80!gray}\underline{13.3} & 
\cellcolor{white!80!gray}\underline{18.1} & \cellcolor{white!80!gray}\underline{7.8} & 
\cellcolor{white!80!gray}\underline{5.4} & 9.6 & 5.6 & 
\cellcolor{white!80!gray}\underline{13.2} & 
\cellcolor{white!80!gray}\underline{13.6} & 
\cellcolor{white!80!gray}\underline{34.9} & 
\cellcolor{white!80!gray}\underline{21.5} & 
\cellcolor{white!80!gray}\underline{22.4} & 
\cellcolor{white!80!gray}\underline{21.7} & 19.2 & 23.5 \\

L-OccGen~\cite{wang2024occgen} & 
\cellcolor{white!65!gray}\textbf{31.6} & 
\cellcolor{white!65!gray}\textbf{16.8} & 
\cellcolor{white!65!gray}\textbf{18.8} & 
\cellcolor{white!80!gray}\underline{5.1} & 
\cellcolor{white!65!gray}\textbf{14.8} & 
\cellcolor{white!65!gray}\textbf{19.6} & 
7.0 & 
\cellcolor{white!65!gray}\textbf{7.7} & 
\cellcolor{white!80!gray}\underline{11.5} & 
\cellcolor{white!65!gray}\textbf{6.7} & 
\cellcolor{white!65!gray}\textbf{13.9} & 
\cellcolor{white!65!gray}\textbf{14.6} & 
\cellcolor{white!65!gray}\textbf{36.4} & 
\cellcolor{white!65!gray}\textbf{22.1} & 
\cellcolor{white!65!gray}\textbf{22.8} & 
\cellcolor{white!65!gray}\textbf{22.3} & 
\cellcolor{white!80!gray}\underline{20.6} & 
\cellcolor{white!80!gray}\underline{24.5} \\

\textbf{\modelname{}} & 
30.0 & 13.0 & 11.6 & 1.5 & 7.9 & 
\cellcolor{white!51!Apricot}\textbf{19.6} & 3.5 & 4.6 & 
\cellcolor{white!51!Apricot}\textbf{22.6} & 
\cellcolor{white!80!Apricot}\underline{5.8} & 5.3 & 11.2 & 28.4 & 8.9 & 13.2 & 14.5 & 
\cellcolor{white!51!Apricot}\textbf{23.3} & 
\cellcolor{white!51!Apricot}\textbf{26.2} \\

\bottomrule
\end{tabular}
}
\label{tab:openocc}
\end{table*}

\subsection{Additional Ablations}
\label{sec:add_ablations}

In this section, we further motivate our design choices by providing additional ablation studies. All ablations are made on the 3D semantic segmentation downstream task.

\paragraph{\textbf{Loss Component Sensitivity.}}

\begin{table}[!t]
    \centering
    \caption{Sensitivity analysis of loss weights for future BEV occupancy diffusion ($\omega_{df}$) and global context distillation ($\omega_{gl}$) on NuScenes~\cite{caesar2020nuscenes}. Results show performance on downstream 3D semantic segmentation.}
    \label{tab:loss_weights}
    \setlength{\tabcolsep}{3pt}
    \begin{tabular}{l|cc}
        \toprule
        Setting & LP & 1\% \\
        \midrule
        $\omega_{df}=0.1$ & 53.95 & 57.12 \\
        $\omega_{df}=0.2$ & 55.82 & 58.65 \\
        $\omega_{df}=0.5$ & 55.98 & 58.82 \\
        $\omega_{df}=\mathbf{1.0}$ & \cellcolor{white!40!SpringGreen}\textbf{56.29} & \cellcolor{white!40!SpringGreen}\textbf{59.46} \\
        \midrule
        $\omega_{gl}=\mathbf{0.05}$ & \cellcolor{white!40!SpringGreen}\textbf{56.29} & \cellcolor{white!40!SpringGreen}\textbf{59.46} \\
        $\omega_{gl}=0.15$ & 56.12 & 59.19 \\
        $\omega_{gl}=0.2$ & 55.93 & 58.47 \\
        \bottomrule
    \end{tabular}
\end{table}

To understand the contribution of each objective during \pretraining to final performance, we independently ablate the weights of the distillation ($\omega_{gl}$) and diffusion ($\omega_{df}$) losses (\cref{tab:loss_weights}), while fixing $\omega_{ds}=1.0$ as a reference.

Consistent with findings in multi-task learning~\cite{liebel2018auxiliary}, downstream performance is sensitive to the weighting of auxiliary \pretraining objectives, although the variation we observe is moderate compared to prior reports such as CleverDistiller~\cite{govindarajan2025cleverdistiller}. Increasing the future BEV occupancy diffusion weight $\omega_{df}$ leads to a monotonic improvement in both linear probing and 1\% fine-tuning. This indicates that the diffusion objective provides a complementary supervisory signal that improves representation quality, rather than destabilizing the training. This observation is aligned with recent works combining diffusion supervision with VFM distillation~\cite{repa,li2025cuao3d, taleoftwofeatures}. Global context distillation yields the largest gains at $\omega_{gl}=0.05$, providing complementary scene-level guidance. Larger values slightly degrade performance, indicating diminishing returns for stronger distillation.


\paragraph{\textbf{Diffusion Loss Components.}}

The auxiliary $\ell_2$ reconstruction loss on $\hat{\mathbf{x}}_{\mathrm{occ}}$ is preferred over a cross-entropy alternative because the denoised estimate $\hat{\mathbf{x}}_{\mathrm{occ}}$ is continuous-valued by construction under the Gaussian forward process~\cite{ddpm}, making pixel-wise $\ell_2$ a consistent supervision signal. Binary cross-entropy is instead the natural choice under discrete-state diffusion formulations~\cite{austin2021d3pm}. Furthermore, augmenting the noise-prediction objective with a direct reconstruction term has been shown to improve training stability in diffusion models~\cite{nichol2021improved}. We balance the two loss terms using a scalar hyperparameter. As shown in~\cref{tab:diff_loss_weights}, adding the reconstruction term consistently improves downstream linear probing performance on nuScenes across all teacher sizes, indicating that adding explicit $\hat{\mathbf{x}}_{\mathrm{occ}}$ supervision helps guide representation learning during \pretraining.

\begin{table}[!t]
    \centering
    \caption{Ablation of diffusion loss terms during \pretraining and their impact on downstream linear probing performance on nuScenes~\cite{caesar2020nuscenes}. \enquote{Noise pred.} denotes using only the diffusion noise-prediction loss, while \enquote{Noise \& recon.} denotes using both noise-prediction and reconstruction losses. Results are reported for different DINOv2 teacher sizes and show downstream performance on 3D semantic segmentation.}
    \label{tab:diff_loss_weights}
    \setlength{\tabcolsep}{6pt}
    \begin{tabular}{l|cc}
        \toprule
        \textbf{Teacher} & \textbf{Noise pred.} & \textbf{Noise \& recon.} \\
        \midrule
        {\color[HTML]{2FC061}\textbf{ViT-S}} & 56.03 & \cellcolor{white!40!SpringGreen}\textbf{56.29} \\
        {\color{Apricot}\textbf{ViT-B}} & 58.16 & \cellcolor{white!51!Apricot}\textbf{58.95} \\
        {\color[HTML]{6434FC}\textbf{ViT-L}} & 59.33 & \cellcolor{white!55!Periwinkle}\textbf{60.06} \\
        \bottomrule
    \end{tabular}
\end{table}

\paragraph{\textbf{Diffusion Temporal Step.}}

We ablate the time interval $\Delta t$ between conditioning frames ($t - 2\Delta t$, $t - \Delta t$) and the target frame ($t$) to determine the optimal temporal receptive field. As shown in~\cref{tab:temporal_ablation}, a stride of $0.5\text{s}$ performs best. Small strides ($0.1\text{s}$) yield negligible point cloud differences, causing the model to collapse into a trivial identity mapping. Conversely, large strides ($1.0\text{s}$) introduce significant non-linear motion and occlusions that break local feature correspondences. Ultimately, using three frames (two conditioning, one target) with a moderate stride is optimal because it provides the minimal curvature and acceleration cues necessary for predicting complex dynamic agents~\cite{blattmann2023videoldm}.

\begin{table}[!t]
    \centering
    \setlength{\tabcolsep}{4pt}
    \renewcommand{\arraystretch}{1.1}
    \caption{Ablation of temporal stride $\Delta t$. We condition on $t{-}2\Delta t$ and $t{-}\Delta t$ to predict $t$. Results are reported on downstream 3D semantic segmentation on nuScenes.}
    \label{tab:temporal_ablation}
    \begin{tabular}{lcccc}
        \toprule
        \multirow{2}{*}{\textbf{Setting}} & \textbf{Stride} & \textbf{Cond.} & \multicolumn{2}{c}{\textbf{nuSc}} \\
        & \textbf{($\Delta t$)} & \textbf{Win.} & \textbf{LP} & \textbf{1\%} \\
        \midrule
        High Freq. & $0.1\text{s}$ & $0.2\text{s}$ & 53.15 & 55.82 \\
        \textbf{Medium} & $\mathbf{0.5\text{s}}$ & $\mathbf{1.0\text{s}}$ & \cellcolor{white!40!SpringGreen}\textbf{56.29} & \cellcolor{white!40!SpringGreen}\textbf{59.46} \\
        Long Horiz. & $1.0\text{s}$ & $2.0\text{s}$ & 54.42 & 58.10 \\
        Mixed Horiz. & $0.1\text{s}$ & $1.2\text{s}$ & 55.40 & 57.63 \\
        \bottomrule
    \end{tabular}
\end{table}

\paragraph{\textbf{Diffusion \vs Occupancy Decoders.}}
We provide an extended version of~\cref{tab:ablation_design_components_compact}, comparing downstream 3D semantic segmentation performance for occupancy-based auxiliary decoders. We report both linear-probing performance and fine-tuning with 1\% of the labels on nuScenes.

\begin{table}[t]
    \centering
    \caption{Comparison of occupancy decoder variants for downstream 3D semantic segmentation on nuScenes.}
    \label{tab:occ_decoder_ablation}
    \begin{tabular}{lcc}
        \toprule
        \textbf{Occupancy Decoder} & \multicolumn{2}{c}{\textbf{nuScenes}} \\
        \cmidrule(lr){2-3}
        & \textbf{LP} & \textbf{1\%} \\
        \midrule
        Sim. Dec.  & 49.1 & 52.9 \\
        ALSO       & 54.7 & 56.6 \\
        Diffusion  & \cellcolor{white!40!SpringGreen}\textbf{56.3}
                   & \cellcolor{white!40!SpringGreen}\textbf{59.5} \\
        \bottomrule
    \end{tabular}
\end{table}

\paragraph{\textbf{Computational Time.}}

The ablation study in~\cref{tab:ablation_components_suppl} demonstrates a consistent performance improvement on the nuScenes dataset as each component is integrated. The full configuration (e) achieves the highest mIoU for both linear probing and 1\% fine-tuning, marking a significant gain over the ScaLR~\cite{scalr} base model. However, these architectural additions introduce a non-negligible overhead where the model parameters increase from $66.4$M to $77.1$M and peak memory usage grows from $36.2$GB to $54.0$GB. This translates to a $16.11\,\%$ parameter increase, which can be compared to the $21.42\,\%$ increase in LP performance. Adding only diffusion (a) yields a $0.6\,\%$ parameter increase with $8.78\,\%$ increase in LP. Notably, diffusion incurs a training-time computational overhead comparable to that of multi-layer distillation (b), primarily because the ground-truth occupancy target is constructed online during training (this target generation could however be shifted to a pre-processing stage). It is important to note that while the reported memory and training time ($\approx$ $30$h) were measured using a batch size of 4 on high-end A100 $80$GB GPUs, the memory can be effectively scaled down by reducing the batch size. This adjustment allows for training when large-memory GPUs are unavailable, and our observations indicate that such a reduction does not result in a significant change in model performance. It is further important to emphasize that the size of the trained LiDAR encoder backbone remains \emph{constant}, and inference-time usage is thus not affected by these \pretraining additions.

\begin{table}[!t]
  \centering
  \setlength{\tabcolsep}{3pt}
  \caption{Ablation study of different components of our \pretraining method on \mbox{\minkunet}~\cite{choy20194minkunet}. Results in mIoU for 3D semantic segmentation. ScaLR~\cite{scalr} with MLP projection head is (base). Memory and time measured with a batch size of 4 on 8 NVIDIA A100 80GB GPUs.}
  \label{tab:ablation_components_suppl}
  \begin{tabular}{c|ccc|cc|ccc}
    \toprule
    \multirow{2}{*}{\#} &
    \multirow{2}{*}{\textbf{Diff}} &
    \multirow{2}{*}{\textbf{Distill}} &
    \multirow{2}{*}{\textbf{\texttt{CLS}}} &
    \multicolumn{2}{c|}{nuScenes} &
    Params. &
    Memory &
    Train Time  \\
    & & & & LP & 1\% & (M) & (GB) & (Hours) \\
    \midrule
    (base) & \textcolor{red}{\xmark}& \textcolor{red}{\xmark} & \textcolor{red}{\xmark} & 46.36 & 55.01 & 66.4 & 36.2 & $\approx$ 20h \\ 
    (a) & \textcolor{green}{\cmark} & \textcolor{red}{\xmark} & \textcolor{red}{\xmark} & 50.43 & 56.97 & 66.8 & 37.8 & $\approx$ 24h \\
    (b) & \textcolor{red}{\xmark} & \textcolor{green}{\cmark} & \textcolor{red}{\xmark} & 53.77 & 57.03 & 71.5 & 46.4 & $\approx$ 24.5h \\
    (c) & \textcolor{red}{\xmark}& \textcolor{green}{\cmark} & \textcolor{green}{\cmark} & 55.13 & 57.81 & 76.7 & 52.5 & $\approx$ 26h \\
    (d) & \textcolor{green}{\cmark}& \textcolor{green}{\cmark} & \textcolor{red}{\xmark} & 55.53 & 59.04 & 71.9 & 47.9 & $\approx$ 28.5h \\
    (e) & \textcolor{green}{\cmark}& \textcolor{green}{\cmark} & \textcolor{green}{\cmark} & \cellcolor{white!40!SpringGreen}\textbf{56.29} & \cellcolor{white!40!SpringGreen}\textbf{59.46} & 77.1 & 54.0 & $\approx$ 30h \\
    \bottomrule
  \end{tabular}
\end{table}

\paragraph{\textbf{Masking.}}


In~\cref{tab:masking_ablation}, we ablate randomly masking the inputs to the auxiliary future BEV occupancy diffusion head. We compare the default unmasked setting against masking either the LiDAR BEV features that condition the diffusion UNet or the noisy future occupancy BEV map itself. Random masking is applied at $50\%$. Masking the BEV conditioning features degrades downstream performance, indicating that depriving the diffusion head of encoder-provided spatiotemporal context weakens the gradients that encourage informative LiDAR features, yielding less transferable representations. Masking the noisy occupancy input reduces linear-probe performance but slightly improves $1\%$ fine-tuning. We hypothesize that moderate masking can increase reliance on the available conditioning signal, consistent with observations that visible-context conditioning is important in masked pretraining~\cite{wei2023diffmae}. While this appears to reduce linear separability (LP), it may provide a mild regularizing effect that benefits $1\%$ fine-tuning.

\begin{table}[!t]
    \centering
    \setlength{\tabcolsep}{5pt} 
    \caption{Ablation of diffusion input masking strategies. We either mask the LiDAR encoder BEV features (conditioning) or the noisy future BEV occupancy map (input), applying random masking at $50\%$. Masking conditioning features degrades downstream performance, indicating that full encoder context is important for learning transferable LiDAR representations. Results show performance on downstream 3D semantic segmentation.}
    \label{tab:masking_ablation}
    \begin{tabular}{l | c c}
        \toprule
        \multirow{2}{*}{\textbf{Strategy}} & \multicolumn{2}{c}{\textbf{nuScenes}} \\
        & LP & 1\% \\
        \midrule
        \textbf{no masking} & \cellcolor{white!40!SpringGreen}\textbf{56.29} & 59.46 \\
        random masking (BEV feat.) & 53.05 &  58.42\\
        random masking (noisy occ.) & 54.61 & \cellcolor{white!40!SpringGreen}\textbf{60.33} \\
        \bottomrule
    \end{tabular}
\end{table}

\paragraph{\textbf{Multi-Layer Distillation Design (Extended).}}

In this section, we ablate the design choices behind our multi-layer distillation strategy and discuss the specific challenges that arise in the cross-modal camera-to-LiDAR setting.

While multi-layer distillation has been studied in 2D~\cite{chen2021distilling}, our contribution is identifying a design that fits point--pixel cross-modal supervision. Prior 2D-VFM-to-3D-LiDAR methods distill only the final teacher layer, discarding the VFM's layer-wise abstraction hierarchy. Unlike same-modality distillation, camera-to-LiDAR supervision is constrained by calibrated point--pixel correspondences. Earlier student blocks aggregate coarser, irregular 3D neighborhoods whose points can project to pixels with different semantics, making fine-grained 2D targets less reliable. At the same time, too early teacher blocks may not match the semantic abstraction level of late student features. \modelname{} therefore matches neighboring late teacher and student blocks, \eg, ($\mathbf{Q}_L{\rightarrow}\mathbf{F}_L$, $\mathbf{Q}_{L-1}{\rightarrow}\mathbf{F}_{L-1}$), where abstraction levels and point--pixel correspondences are both more consistent. We refer to this strategy as the \enquote{\textit{Separate}} strategy.

\begin{figure}[tb]
    \centering
    \includegraphics[width=0.6\linewidth, trim={1cm 1cm 33cm 0cm}, clip]{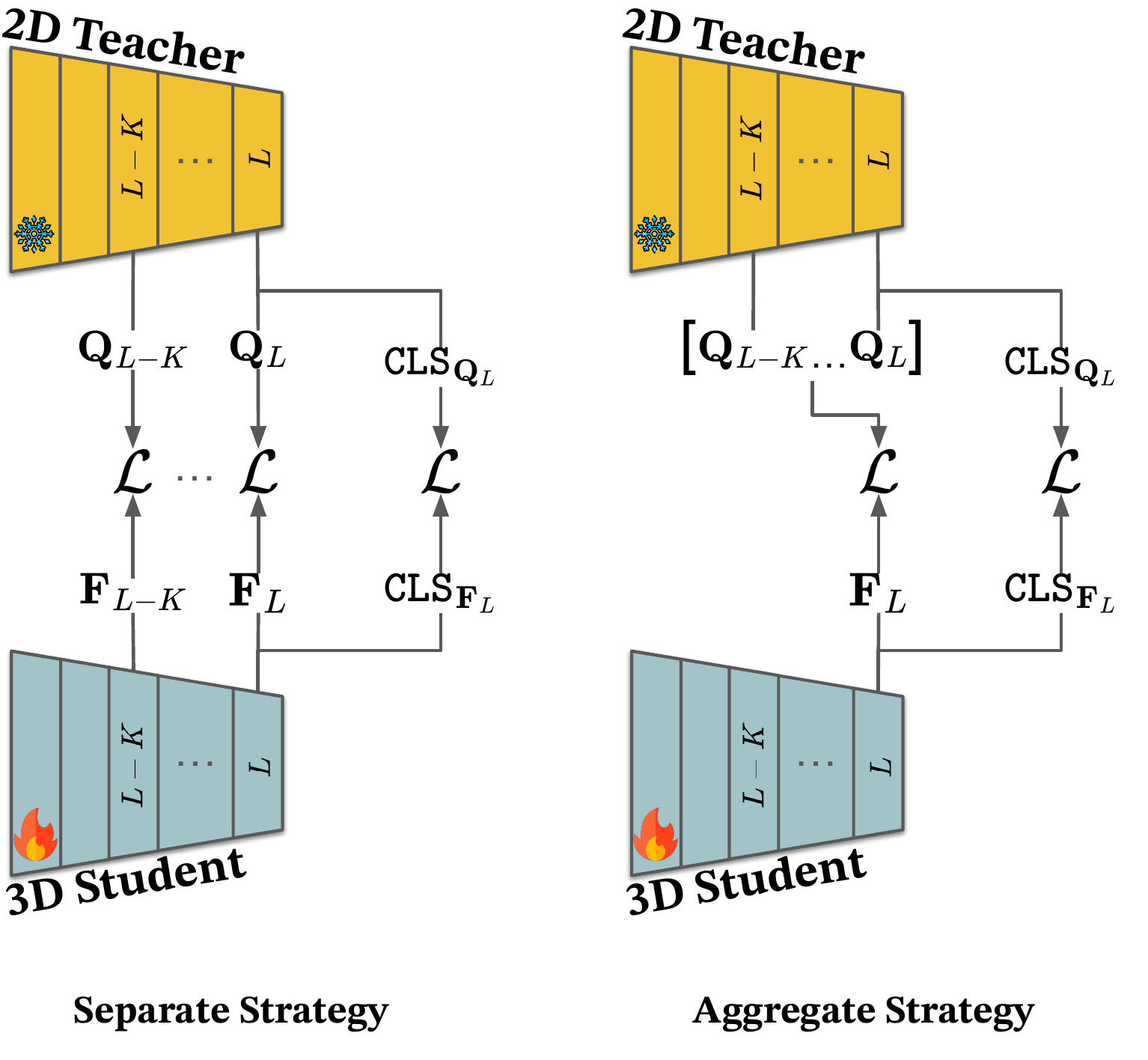}
    \caption{Illustration of the \emph{Separate} and \emph{Aggregate} multi-layer distillation strategies ablated in~\cref{sec:add_ablations}. The separate strategy is shown to produce better performance on downstream 3D semantic segmentation.}
    \label{fig:sep_vs_agg_strat}
\end{figure}

We evaluate five different multi-layer distillation designs corresponding to: (i) ablating the distillation depth for the Separate strategy; (ii) complementing the depth ablation by a variant distilling only from the penultimate layer of the teacher; (iii) the \textit{Aggregate} strategy where we concatenate features from the last $K$ teacher layers to one single target $\{\mathbf{Q}_{L-k}\}_{k=0}^{K-1}{\rightarrow}\mathbf{F}_{L}$; (iv) a \enquote{\textit{Knowledge Review}}-style~\cite{chen2021distilling} (KR-style) aggregation, where aggregated teacher features supervise the final two student layers. Specifically, for a $K$-layer aggregation, we distill $\{\mathbf{Q}_{L-k}\}_{k=j}^{K-1}{\rightarrow}\mathbf{F}_{L-j}$, $j=0,1$; and (v) the \textit{Non-Uniform} variant where student-teacher layer pairs are distilled with a skipped layer in-between ($\mathbf{Q}_{L}{\rightarrow}\mathbf{F}_L$, $\mathbf{Q}_{L-2}{\rightarrow}\mathbf{F}_{L-2}$). The Separate (\modelname{} default) and Aggregate strategies are illustrated in~\cref{fig:sep_vs_agg_strat}. Note that, in~\cref{sec:ptv3}, we also evaluate multi-layer distillation using a different PTv3~\cite{ptv3} student model architecture.



\Cref{tab:layer_ablation_merged} shows that distilling from multiple layers consistently outperforms using only the final teacher layer (\emph{Last 1}), indicating that intermediate teacher features provide complementary supervision. Furthermore, the penultimate-teacher-layer distillation results in~\cref{tab:knowledge_review} indicate that the performance gains of multi-layer distillation are not attributable to the penultimate layer alone, but rather to the use of multiple teacher layers. As seen in~\cref{tab:layer_ablation_merged}, the optimal distillation depth depends on the distillation strategy, and the trend is consistent across both studied training setups (all \pretraining loss components included vs. multi-layer distillation only). Separate outperforms Aggregate in all cases and achieves its best results by distilling the last two layers. In contrast, Aggregate is strongest when applied to the last three layers. We hypothesize that Separate is more effective because it preserves layer-wise correspondence, providing matched supervision across abstraction levels. In contrast, Aggregate compresses features from multiple teacher layers into a single target for the final student layer, which can dilute layer-specific signals. We further attribute Separate peaking at the 2-layer depth to the fact that distilling earlier layers can violate geometric correspondence, where earlier and coarser 3D (\minkunet{}) up-sampling features do not perfectly align with the VFM’s 2D patches. In contrast, Aggregate benefits from including one additional teacher layer (last three), which may enrich the final-layer target with useful mid-level cues before the aggregated representation becomes too heterogeneous to compress effectively.

\begin{table}[!t]
    \centering
    \setlength{\tabcolsep}{3pt}
    \renewcommand{\arraystretch}{1.15}
    \caption{Ablation of distillation depth and multi-layer distillation strategy on nuScenes linear probing. We report results for \pretraining with all loss components enabled, and when only multi-layer distillation is included. Results show performance on downstream 3D semantic segmentation.}
    \label{tab:layer_ablation_merged}
    \begin{tabular}{c|cc|cc}
        \toprule
        \textbf{Layers} &
        \multicolumn{2}{c|}{\textbf{All losses included}} &
        \multicolumn{2}{c}{\textbf{Multi-layer distillation}} \\
        \cmidrule(lr){2-3} \cmidrule(lr){4-5}
        & Separate & Aggregate & Separate & Aggregate \\
        \midrule
        Last 1 & 50.84 & --   & 46.36 & -- \\
        \textbf{Last 2} & \cellcolor{white!40!SpringGreen}\textbf{56.29} & 53.62 & \cellcolor{white!40!SpringGreen}\textbf{53.77} & 51.92 \\
        Last 3 & 52.59 & 55.37 & 52.02 & 52.97 \\
        Last 4 & 52.19 & 54.54 & 51.56 & 52.25 \\
        \bottomrule
    \end{tabular}
\end{table}

The results in~\cref{tab:knowledge_review} show that the proposed Separate strategy remains strongest. KR-style aggregation improves over the final-block baseline but remains below Separate matching. We attribute this to the same aggregation bottleneck we observed with the Aggregate strategy. Non-uniform matching also underperforms neighboring-block matching (Separate), suggesting that, in the cross-modal setting, preserving local semantic correspondence between late-stage blocks is more effective than distilling broader non-adjacent feature-transformation processes. Together with~\cref{tab:layer_ablation_merged}, these results indicate that \modelname{}'s multi-block distillation is a principled design effectively implemented for the camera-to-LiDAR distillation setting. Furthermore, these results stay consistent with prior observations that late teacher representations (DINOv2~\cite{dinov2}) are most aligned with semantic segmentation~\cite{perception_encoder}.

\begin{table}[!t]
    \centering
    \caption{Ablation of multi-layer matching strategies. Separate matching yields the strongest overall performance. We report results for \pretraining when only multi-layer distillation is included. Results show performance on downstream 3D semantic segmentation.}
    \label{tab:knowledge_review}
    \setlength{\tabcolsep}{4pt}
    \begin{tabular}{lcccc}
        \toprule
        \multirow{2}{*}{\textbf{Strategy}} & \multicolumn{2}{c}{\textbf{nuScenes}} & \textbf{SKITTI} & \textbf{Waymo} \\
        \cmidrule(lr){2-3}
        & LP & 1\% & 1\% & 1\% \\
        \midrule
        Penultimate & 49.9 & 55.4 & 46.5 & 48.9 \\
        2-layer KR & 50.0 & 56.8 & 47.2 & 49.2 \\
        3-layer KR & 51.4 & \cellcolor{white!40!SpringGreen}\textbf{57.0} & 47.8 & 49.7 \\
        4-layer KR & 51.1 & 56.3 & 47.5 & 49.6 \\
        Non-uniform & 52.4 & 56.1 & 50.5 & 50.2 \\
        Separate (\modelname{}) & \cellcolor{white!40!SpringGreen}\textbf{53.8} & \cellcolor{white!40!SpringGreen}\textbf{57.0} & \cellcolor{white!40!SpringGreen}\textbf{50.7} & \cellcolor{white!40!SpringGreen}\textbf{50.6} \\
        \bottomrule
    \end{tabular}
\end{table}

Furthermore, we ablate the dense point--pixel alignment loss in~\cref{tab:multi-layer-loss}. For unit-normalized features, $\ell_2$ and cosine distance are equivalent. The practical difference between the two formulations is thus whether feature magnitudes are also constrained. In our setting, the teacher (ViT) and student (\minkunet{}) features originate from different architectures and normalization schemes, so their norm distributions are not expected to be aligned. Consequently, when applied across millions of point--pixel pairs and multiple distillation layers, an $\ell_2$ objective can be dominated by scale mismatch, potentially steering optimization away from semantic feature direction. Cosine distance instead focuses the supervision on directional alignment, and yields better downstream performance in~\cref{tab:multi-layer-loss}. Notably, LiMA~\cite{lima} adopts an $\ell_2$ formulation, whereas ScaLR~\cite{scalr} and CleverDistiller~\cite{govindarajan2025cleverdistiller} use cosine. Moreover, cosine loss also outperforms the Kullback--Leibler (KL) divergence loss, which aligns predictive distributions and may be more sensitive to distribution shifts. Finally, as discussed in~\cref{sec:related_work}, contrastive losses require negatives, positive-negative sampling strategies (because an exhaustive point-pixel comparison is computationally prohibitive), semantic priors/superpixels, and temperature calibration. Recent state-of-the-art methods have moved away from contrastive formulations~\cite{govindarajan2025cleverdistiller,scalr,lima}. Based on these results and considerations, we use cosine distance as the default formulation for $\mathcal{L}_{\mathrm{distill}}$ in \modelname{}. For the global context distillation we use the $\ell_2$ loss because there we have a single pooled feature per scene, and scale mismatch is less dominant. Ablations further show the $\ell_2$ loss to be dominant for the final-layer derived $\cls$ token distillation.

\begin{table}[!t]
    \centering
    \caption{Ablation of multi-layer distillation loss formulations. Cosine distance provides the strongest downstream performance across datasets and is used as the default $\mathcal{L}_{\mathrm{distill}}$ in \modelname{}. Results show performance on downstream 3D semantic segmentation.}
    \label{tab:multi-layer-loss}
    \setlength{\tabcolsep}{4pt}
    \begin{tabular}{lcccc}
        \toprule
        \multirow{2}{*}{\textbf{Loss}} & \multicolumn{2}{c}{\textbf{nuScenes}} & \textbf{SKITTI} & \textbf{Waymo} \\
        \cmidrule(lr){2-3}
        & LP & 1\% & 1\% & 1\% \\
        \midrule
        $\ell_2$ & 53.8 & 58.8 & 52.2 & 51.5 \\
        KL & 52.1 & 57.4 & 51.0 & 50.7 \\
        Cosine (\modelname{}) & \cellcolor{white!40!SpringGreen}\textbf{56.3} & \cellcolor{white!40!SpringGreen}\textbf{59.5} & \cellcolor{white!40!SpringGreen}\textbf{52.9} & \cellcolor{white!40!SpringGreen}\textbf{52.2} \\
        \bottomrule
    \end{tabular}
\end{table}

\paragraph{\textbf{PTv3 with \modelname{}.}}
\label{sec:ptv3}

In the main paper, we use \minkunet{}~\cite{choy20194minkunet} as the LiDAR encoder during \pretraining. To assess whether our proposed cross-modal \pretraining transfers to alternative 3D backbones, we additionally instantiate the LiDAR encoder as PTv3~\cite{ptv3} and repeat the same \pretraining and downstream evaluation.~\Cref{tab:ptv3} reports linear probing and fine-tuning results on nuScenes, as well as 1\% fine-tuning on SemanticKITTI and Waymo. We observe that \modelname{} yields consistent gains over prior distillation baselines for PTv3, indicating that the benefits of our \pretraining are not specific to \minkunet{} and generalize to a substantially different LiDAR architecture. Finally, we ablate the multi-layer distillation depth by distilling multiple teacher--student layer pairs, where \enquote{3 layers} denotes using three layer pairs in the multi-layer distillation objective. In this setting, increasing the distillation depth is generally beneficial. For ViT-B, the 3-layer variant achieves the strongest results in every reported setting, while for ViT-S it gives the best results in most nuScenes settings and on Waymo, with the 2-layer variant performing best on nuScenes $25\,\%$ and SemanticKITTI $1\,\%$. This suggests that the optimal number of distilled layer pairs is backbone- and setting-dependent ($2$ layers for \minkunet{}. See~\cref{tab:layer_ablation_merged}). This may reflect differences in how \minkunet{} (a sparse backbone with an encoder--decoder hierarchy) and PTv3 (a transformer-based point backbone) organize intermediate representations, suggesting that the appropriate multi-layer distillation depth from a DINOv2~\cite{dinov2} ViT~\cite{vit} teacher should be tuned per encoder. Concretely, deeper distillation exposes the student to features at varying abstraction levels, and the most effective depth likely depends on how well these levels align with the student’s intermediate representations.


\begin{table*}[!t]
\centering

\caption{3D semantic segmentation results for cross-modal distillation methods \pretrained on nuScenes~\cite{caesar2020nuscenes}, and fine-tuned on nuScenes, SemanticKITTI~\cite{behley2019semantickitti}, and Waymo~\cite{sun2020waymo}. All methods use the PTv3~\cite{ptv3} backbone. LP denotes linear probing with a frozen backbone.}

\label{tab:ptv3}
\renewcommand{\arraystretch}{0.9}
\setlength{\tabcolsep}{4.5pt}
\resizebox{\linewidth}{!}{
\begin{tabular}{lccccccc}
\toprule
\multirow{2}{*}{\textbf{Method}} &
\multicolumn{5}{c}{\textbf{nuScenes}} &
\textbf{SKITTI} &
\textbf{Waymo} \\
\cmidrule(lr){2-6} \cmidrule(lr){7-7} \cmidrule(lr){8-8}
& \textbf{LP} & \textbf{1\%} & \textbf{5\%} & \textbf{10\%} & \textbf{25\%} & \textbf{1\%} & \textbf{1\%} \\
\hline
\arrayrulecolor{black} \midrule \arrayrulecolor{lightgray}

\multicolumn{8}{c}{{\color[HTML]{2FC061} \textbf{ViT-S}}} \\

ScaLR  &  {44.99} & {62.07} & {69.10} & {72.91} & {75.69} & 56.92 & 60.09 \\

CleverDistiller &  
\cellcolor{white!80!SpringGreen}\underline{58.49} & 
\cellcolor{white!80!SpringGreen}\underline{62.54} & \cellcolor{white!80!SpringGreen}\underline{71.09} & 
\cellcolor{white!80!SpringGreen}\underline{73.89} & \cellcolor{white!80!SpringGreen}\underline{76.64} & \cellcolor{white!80!SpringGreen}\underline{61.95} & \cellcolor{white!80!SpringGreen}\underline{62.19} \\

\textbf{\modelname{}} & 59.16 & 65.39 & 72.21 & 75.27 & \cellcolor{white!40!SpringGreen}\textbf{78.86} & \cellcolor{white!40!SpringGreen}\textbf{64.30} & 65.09 \\

\textbf{\modelname{}} 3 layers & 
\cellcolor{white!40!SpringGreen}\textbf{60.18}& 
\cellcolor{white!40!SpringGreen}\textbf{67.15} & 
\cellcolor{white!40!SpringGreen}\textbf{73.03} & 
\cellcolor{white!40!SpringGreen}\textbf{76.02} & 
77.68 & 
64.11 & 
\cellcolor{white!40!SpringGreen}\textbf{65.72}
\\

\arrayrulecolor{black} \midrule \midrule \arrayrulecolor{lightgray}

\multicolumn{8}{c}{{\color{Apricot} \textbf{ViT-B}}} \\

ScaLR &  {41.62} & {63.54} & {71.74} & {73.84} & {76.56} & 60.74 & 57.17 \\

CleverDistiller &  
\cellcolor{white!80!Apricot}\underline{60.54} & \cellcolor{white!80!Apricot}\underline{63.93} & \cellcolor{white!80!Apricot}\underline{72.27} & \cellcolor{white!80!Apricot}\underline{74.11} & \cellcolor{white!80!Apricot}\underline{76.93} &\cellcolor{white!80!Apricot}\underline{63.31} & \cellcolor{white!80!Apricot}\underline{62.20} \\

\textbf{\modelname{}} & 60.15 & 67.84 &75.59 & 76.80 & 78.14& 62.31& 62.07\\

\textbf{\modelname{}} 3 layers &
\cellcolor{white!51!Apricot}\textbf{61.49} & 
\cellcolor{white!51!Apricot}\textbf{69.80} & 
\cellcolor{white!51!Apricot}\textbf{77.33} & 
\cellcolor{white!51!Apricot}\textbf{78.44} & 
\cellcolor{white!51!Apricot}\textbf{79.00} & 
\cellcolor{white!51!Apricot}\textbf{63.80} & 
\cellcolor{white!51!Apricot}\textbf{62.93} \\







\arrayrulecolor{black} \bottomrule
\end{tabular}}
\end{table*}

\paragraph{\textbf{Global Context Distillation Design (Extended).}}

We ablate three design choices for global context distillation: the aggregation of per-view teacher \texttt{CLS} tokens to construct $\cls_{\mathbf{Q}_L}$, the aggregation of projected sparse student voxel features to construct $\cls_{\mathbf{F}_L}$, and the loss used to align global teacher and student representations. For teacher-side aggregation, we compare average and max pooling for combining the per-view teacher \cls{} token features into $\cls_{\mathbf{Q}L}$. For student-side aggregation, we compare max pooling against a learnable single-query attention pooling module and a per-view/frustum-level distillation strategy, where a separate student $\cls_{\mathbf{F}_L}$ token is constructed for each teacher-view \cls{} token and the resulting distillation losses are averaged.

As shown in~\cref{tab:pooling_teacher,tab:pooling_student}, max pooling consistently improves performance for both linear probing and 1\% fine-tuning, suggesting that scene-level alignment benefits from preserving the strongest camera view-specific and point cloud feature responses. Conversely, average pooling can attenuate salient cues by averaging them with less informative or noisier views. On the student side, learnable pooling introduces additional flexibility without a corresponding performance gain, while the per-view $\cls_{\mathbf{F}_L}$ variant also underperforms global max pooling. This indicates that a single global scene-level \texttt{CLS} distillation objective best complements the local multi-view distillation, making the hierarchical distillation design of \modelname{} the most effective configuration.

\begin{table}[!t]
    \centering
    \caption{Effect of global \texttt{CLS} token aggregation (pooling strategy) on the \textit{teacher} side ($\cls_{\mathbf{Q}_L}$) evaluated on downstream nuScenes 3D semantic segmentation. Pooling occurs over the per-view teacher CLS tokens in the multi-view input.}
    \label{tab:pooling_teacher}
    \setlength{\tabcolsep}{4pt}
    \begin{tabular}{lcc}
        \toprule
        \multirow{2}{*}{\textbf{Pooling}} & \multicolumn{2}{c}{\textbf{nuScenes}} \\
        \cmidrule(lr){2-3}
        & LP & 1\% \\
        \midrule
        Average pooling & 55.52 & 58.91 \\
        Max pooling (\modelname{}) & \cellcolor{white!40!SpringGreen}\textbf{56.29} & \cellcolor{white!40!SpringGreen}\textbf{59.46} \\
        \bottomrule
    \end{tabular}
\end{table}

\begin{table}[!t]
    \centering
    \caption{Effect of global \texttt{CLS} token aggregation (pooling strategy) on the \textit{student} side ($\cls_{\mathbf{F}_L}$) evaluated on downstream 3D semantic segmentation. Pooling occurs over the projected sparse voxel features $\mathcal{H}_\cls{}(\mathbf{F}_L)$ produced from the point cloud $\mathcal{P}$.}
    \label{tab:pooling_student}
    \setlength{\tabcolsep}{4pt}
    \begin{tabular}{lcccc}
        \toprule
        \multirow{2}{*}{\textbf{Pooling}} & \multicolumn{2}{c}{\textbf{nuScenes}} & \textbf{SKITTI} & \textbf{Waymo} \\
        \cmidrule(lr){2-3}
        & LP & 1\% & 1\% & 1\% \\
        \midrule
        Learnable pooling & 55.9 & 58.9 & 52.2 & 51.4 \\
        Per-image $\cls_{\mathbf{F}_L}$ & 55.5 & 57.7 & 50.4 & 50.6 \\
        Max pooling (\modelname{}) & \cellcolor{white!40!SpringGreen}\textbf{56.3} & \cellcolor{white!40!SpringGreen}\textbf{59.5} & \cellcolor{white!40!SpringGreen}\textbf{52.9} & \cellcolor{white!40!SpringGreen}\textbf{52.2} \\
        \bottomrule
    \end{tabular}
\end{table}

Finally,~\cref{tab:global_loss_formulation} shows that $\ell_2$ alignment slightly outperforms cosine similarity in the low-label fine-tuning setting, while matching it for linear probing on nuScenes. We therefore use $\ell_2$ loss for global context distillation, together with max pooling on both the teacher and student sides.

\begin{table}[!t]
    \centering
    \caption{Effect of global context distillation loss formulation on downstream 3D semantic segmentation.}
    \label{tab:global_loss_formulation}
    \setlength{\tabcolsep}{4pt}
    \begin{tabular}{lcccc}
        \toprule
        \multirow{2}{*}{\textbf{Loss}} & \multicolumn{2}{c}{\textbf{nuScenes}} & \textbf{SKITTI} & \textbf{Waymo} \\
        \cmidrule(lr){2-3}
        & LP & 1\% & 1\% & 1\% \\
        \midrule
        Cosine similarity & \cellcolor{white!40!SpringGreen}\textbf{56.3} & 59.1 & 52.5 & 52.1 \\
        $\ell_2$ (\modelname{}) & \cellcolor{white!40!SpringGreen}\textbf{56.3} & \cellcolor{white!40!SpringGreen}\textbf{59.5} & \cellcolor{white!40!SpringGreen}\textbf{52.9} & \cellcolor{white!40!SpringGreen}\textbf{52.2} \\
        \bottomrule
    \end{tabular}
\end{table}

\section{Additional Qualitative Results}
\label{sec:add_qualitative}

In this section, we provide additional qualitative visualizations to complement the quantitative results in the main paper and to better illustrate the behavior of our method across different scenarios.

\subsection{3D Semantic Segmentation}

\begin{figure}[h]
    \centering
    \centering 
  \includegraphics[width=1\linewidth, trim=0 3cm 12cm 0, clip]{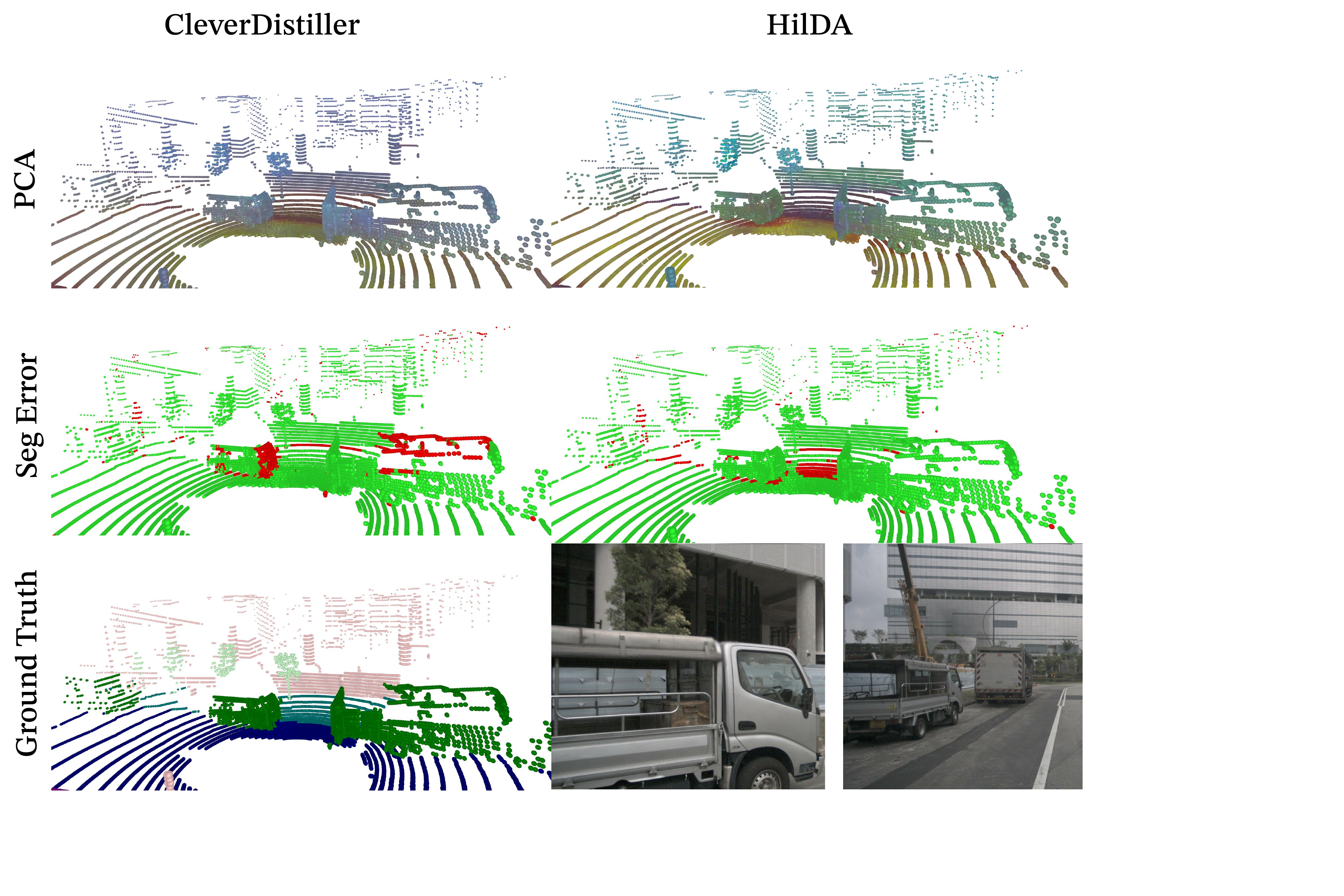}
    \caption{Qualitative comparison for 3D semantic segmentation. We show PCA projections of \modelname{} and CleverDistiller~\cite{govindarajan2025cleverdistiller} feature embeddings, and the corresponding error maps.}
    \label{fig:pca1}
\end{figure}

\begin{figure}[h]
    \centering
    \includegraphics[width=1\linewidth, trim=0 2cm 12cm 0, clip]{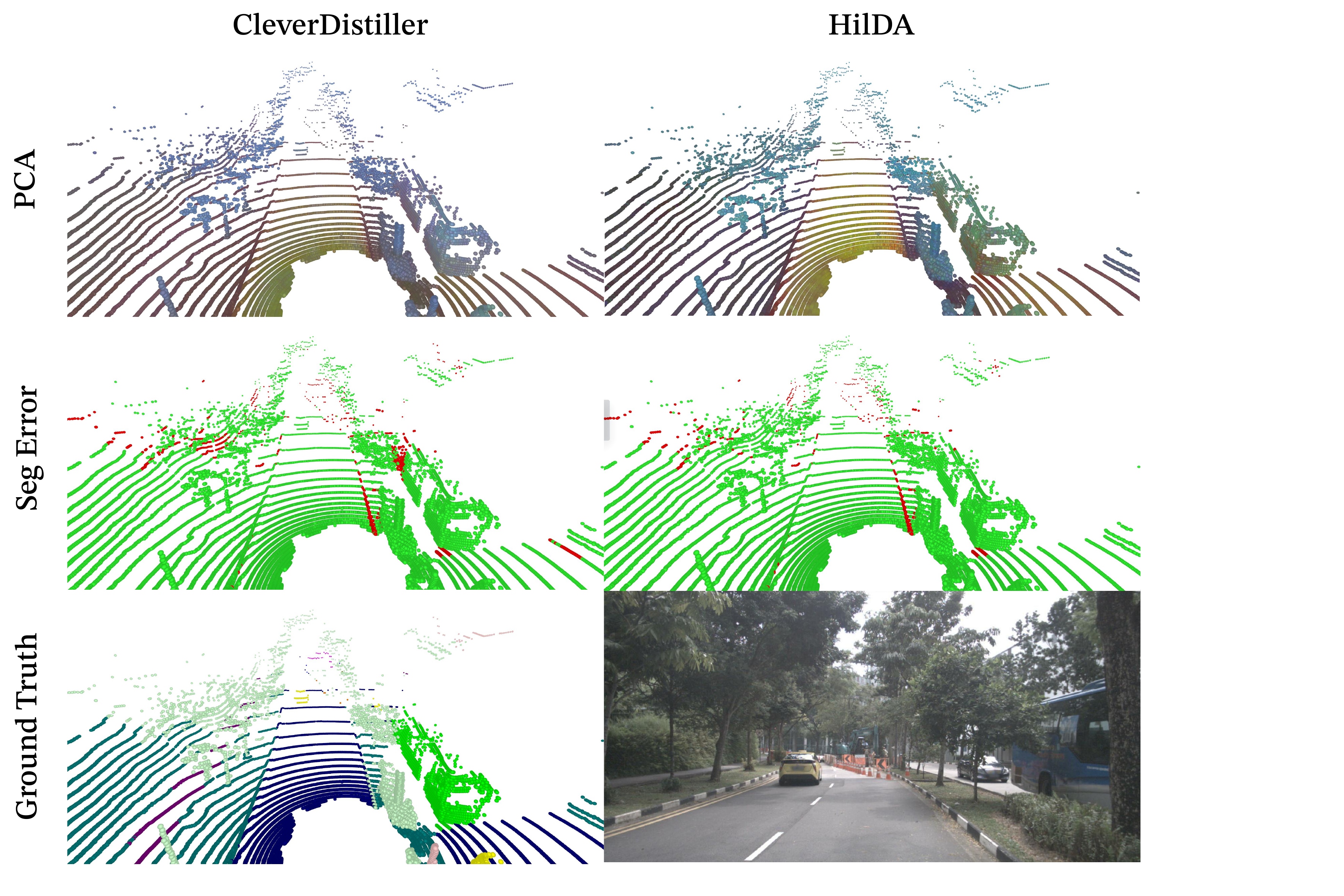}
    \caption{Qualitative comparison for 3D semantic segmentation covering a different scene. We show PCA projections of \modelname{} and CleverDistiller~\cite{govindarajan2025cleverdistiller} feature embeddings, and the corresponding error maps.}
    \label{fig:pca2}
\end{figure}

\begin{figure}[!t]
  \centering 
  \includegraphics[width=1.0\linewidth, trim=25 80 25 150, clip]{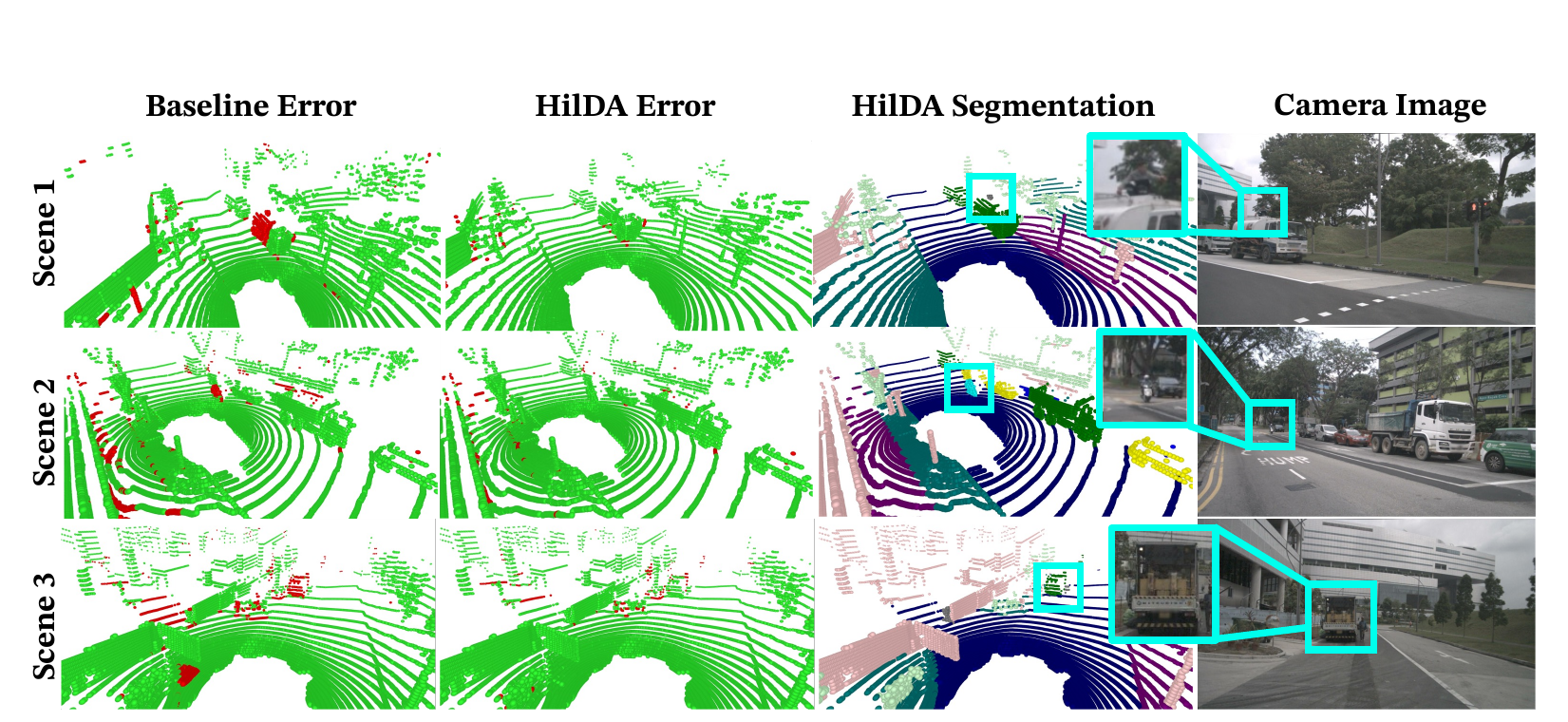}
  \caption{Qualitative comparison for 3D semantic segmentation highlighting difficult examples for three different scenes. We show segmentation errors for both a baseline model (ScaLR~\cite{scalr}) and \modelname{}, the semantic prediction made by \modelname{}, and a camera image for clarity. The scenes show cases where \modelname{} outperforms the baseline on classifying: a pedestrian on top of truck, a scooter, and a truck with opened rear cargo doors.}
  \label{fig:sem_seg_suppl_hard}
\end{figure}

In~\cref{fig:pca1,fig:pca2}, we provide a qualitative comparison on two representative nuScenes scenes. We visualize a PCA projection of the learned point embeddings, the predicted semantic segmentation, and per-point error maps for \modelname{} and CleverDistiller~\cite{govindarajan2025cleverdistiller}. The PCA views serve as a proxy for feature separability, where more coherent and class-consistent clusters typically indicate a representation that is easier to linearly separate. Notably, the PCA separation of \modelname{} bears resemblance to its segmentation masks. Across both scenes, \modelname{} exhibits fewer and more spatially localized errors than CleverDistiller, aligning with our quantitative gains and suggesting that the proposed \pretraining improves the semantic prediction capability of our \pretrained features, leading to cleaner decision boundaries. At the same time, \modelname{} can still struggle with geometrically ambiguous surface regions, such as the ground underneath vehicles. This is consistent with the BEV diffusion limitation discussed in~\cref{sec:multi-task}, since the auxiliary occupancy target emphasizes 2D object-centric geometric structure rather than fine-grained surface semantics. Moreover,~\cref{fig:sem_seg_suppl_hard} highlights challenging classes and scenes where \modelname{} also outperforms ScaLR.

\subsection{3D Object Detection}

\begin{figure}[!t]
  \centering 
  \includegraphics[width=1.0\linewidth, trim=22 78 10 35, clip]{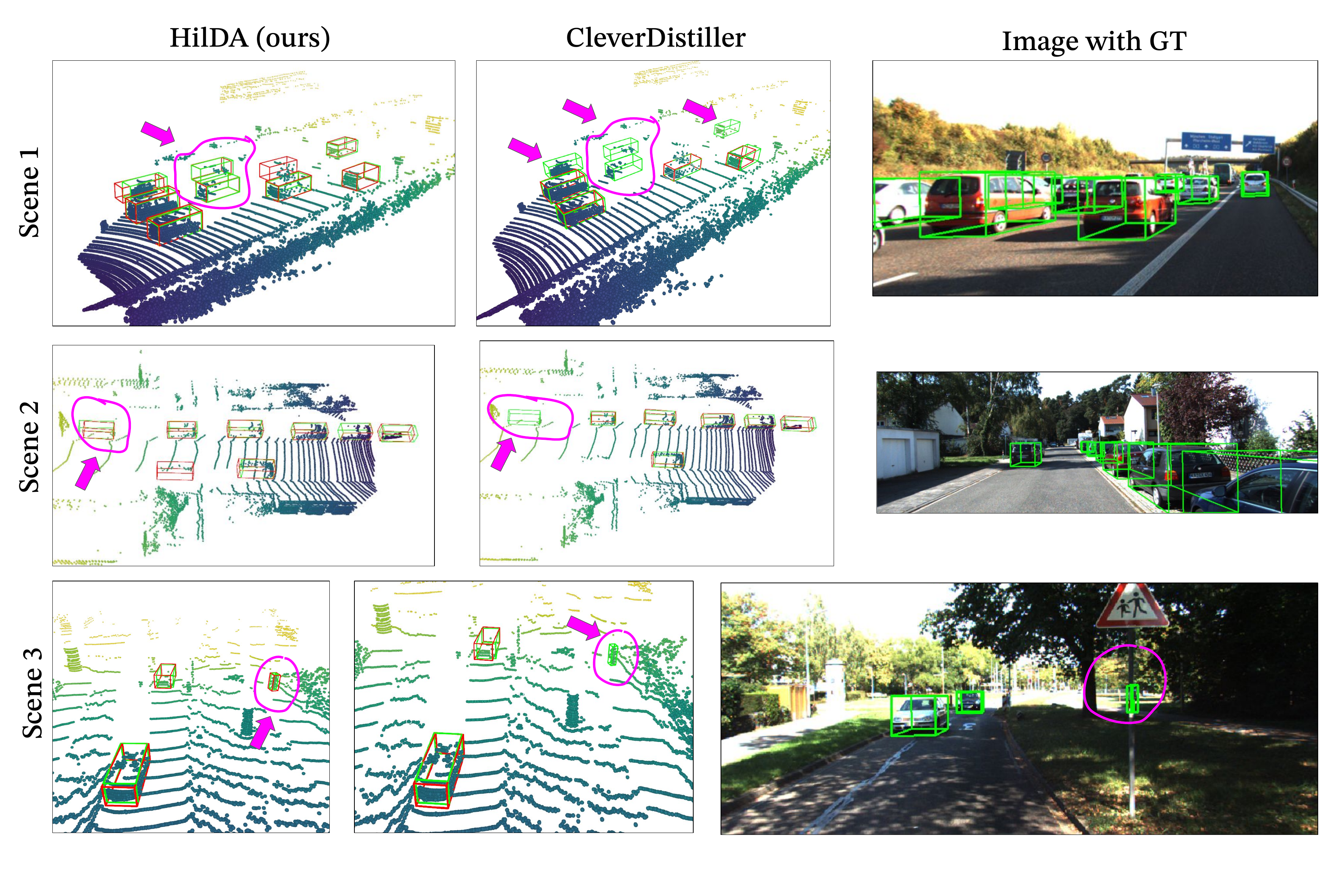}
  \caption{Qualitative comparison of 3D object detection across three scenes, comparing CleverDistiller with \modelname{}. Green boxes denote ground truth and red boxes denote predictions. The examples show cases where \modelname{} better detects object instances, particularly at far distances.}
  \label{fig:3dod_suppl_1}
\end{figure}

\begin{figure}[!t]
  \centering 
  \includegraphics[width=1.0\linewidth, trim=45 84 64 240, clip]{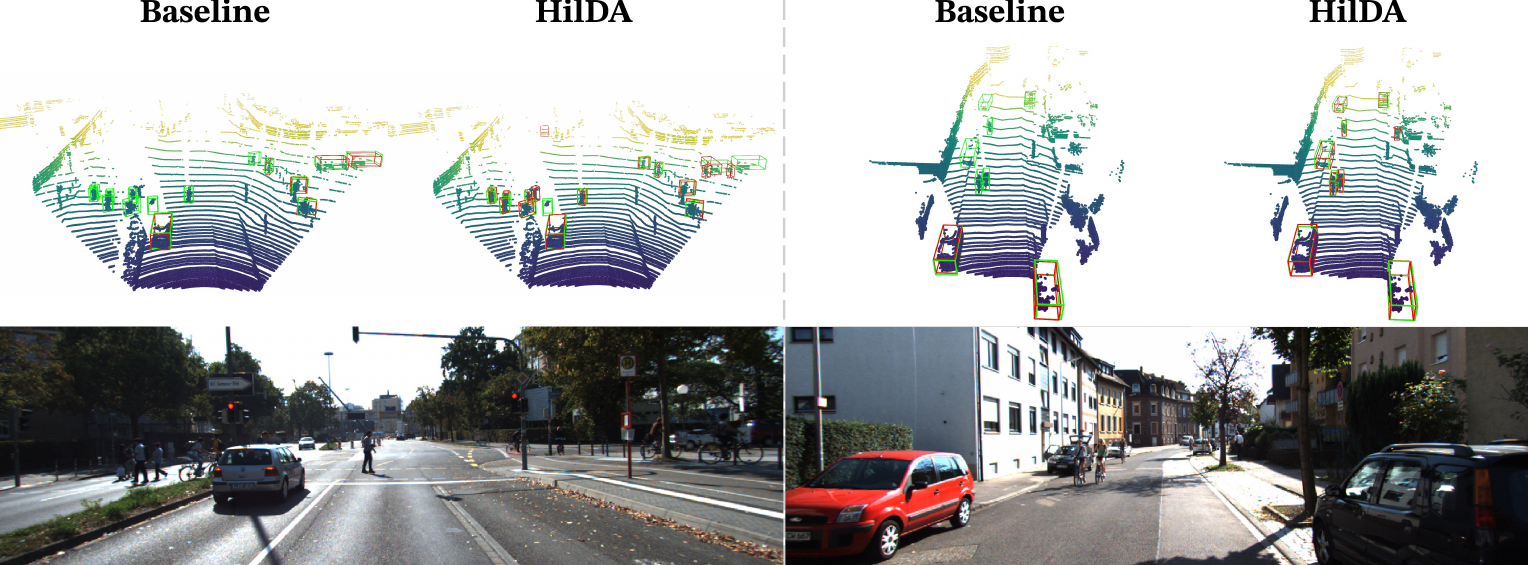}
  \caption{Additional qualitative 3D object detection scenes where \modelname{} outperforms CleverDistiller, exhibiting fewer false negatives across various object categories. Best viewed zoomed \faSearch.}
  \label{fig:3dod_suppl_2}
\end{figure}

In~\cref{fig:3dod_suppl_1,fig:3dod_suppl_2}, we provide additional qualitative results for 3D object detection, comparing CleverDistiller against \modelname{} across multiple nuScenes scenes. Across these examples, \modelname{} yields more complete detections with noticeably fewer false negatives over various object categories. The improvements are particularly apparent for long-range instances, where \modelname{} more reliably recovers objects that are missed by CleverDistiller, consistent with the quantitative gains reported in the main paper.

\subsection{Semantic Occupancy Prediction}

\begin{figure}[!t]
  \centering
  \includegraphics[width=0.77\linewidth, trim=45 84 64 83, clip]{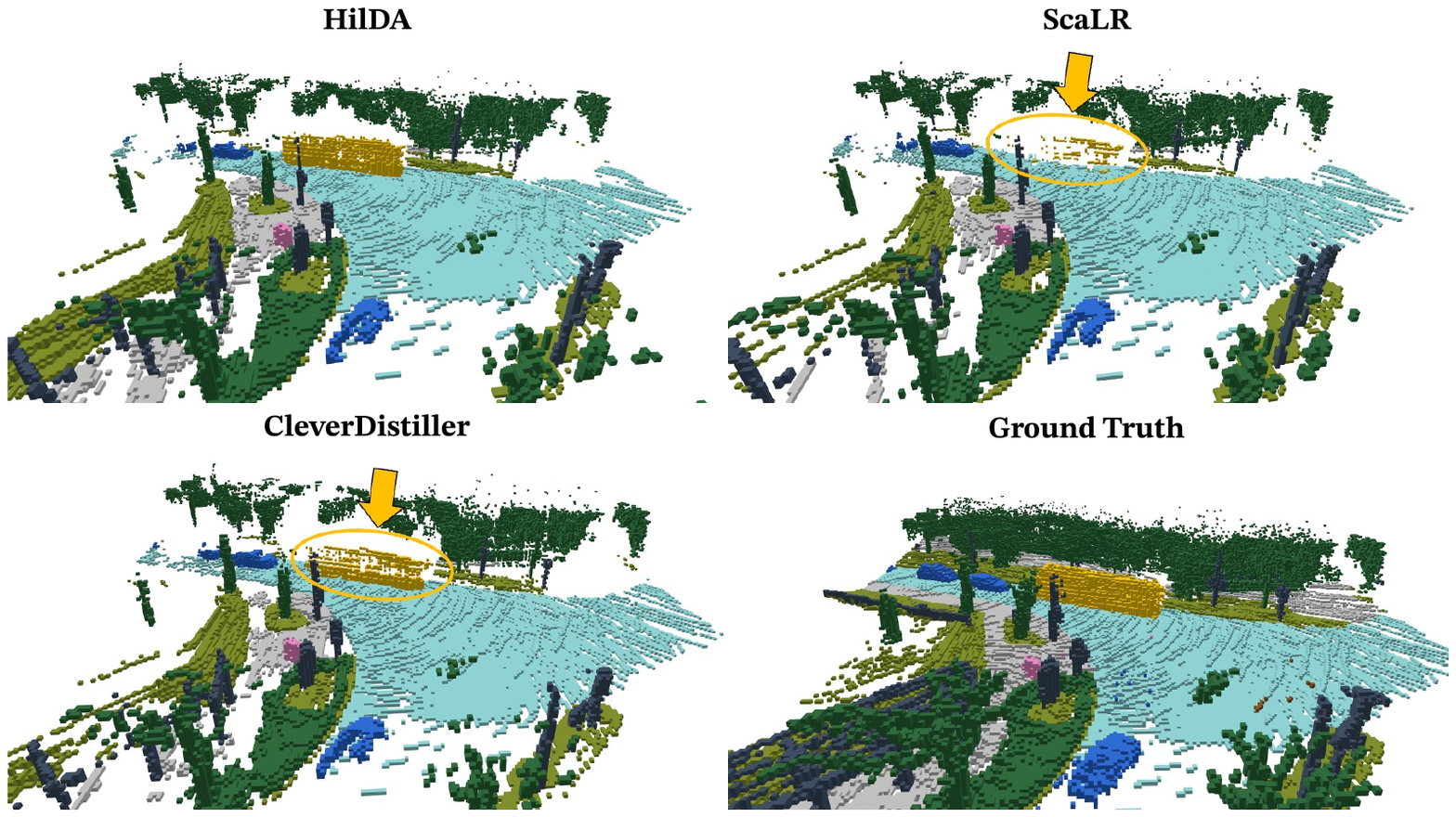}

  \includegraphics[width=0.77\linewidth, trim=45 84 64 83, clip]{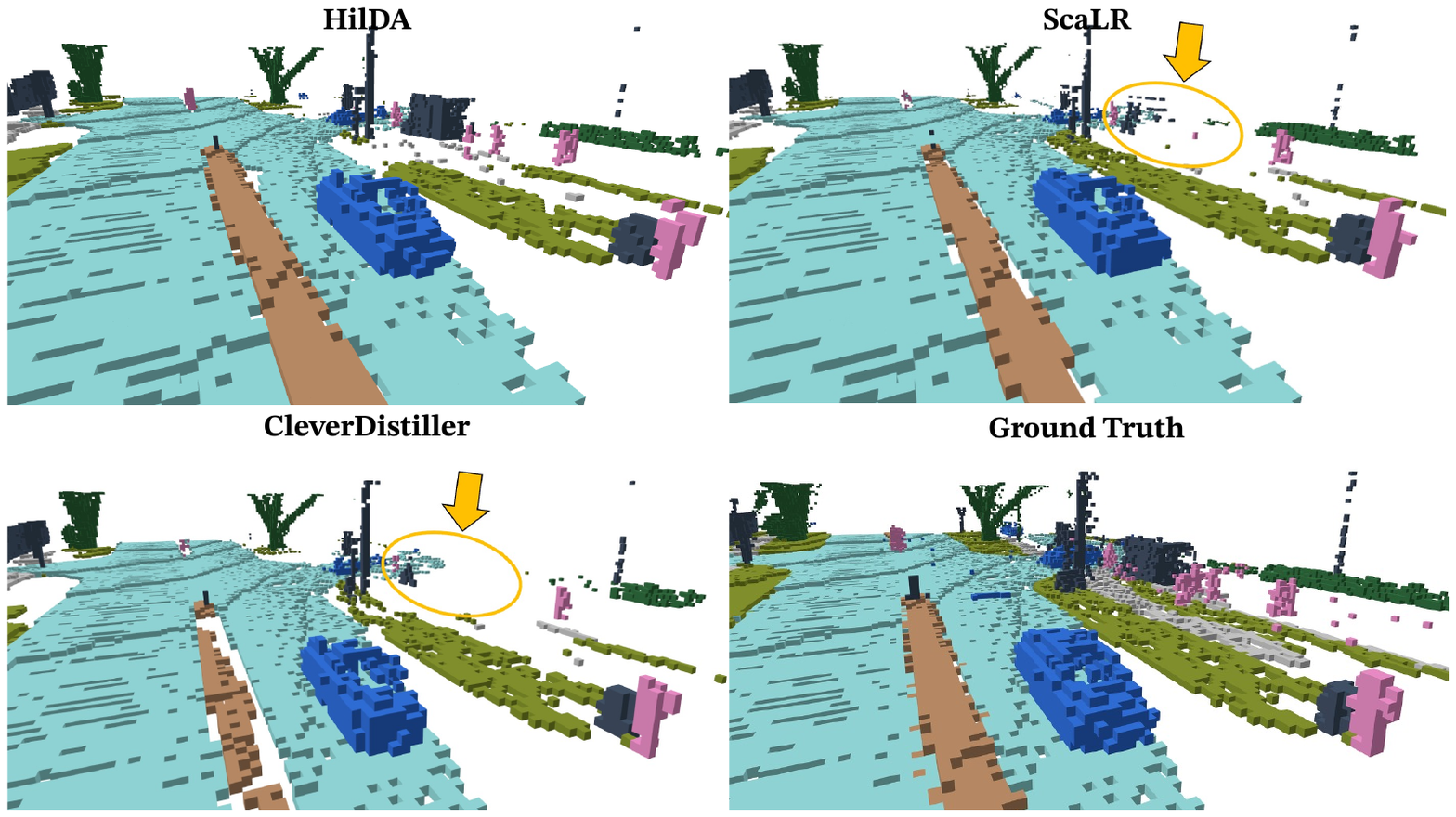}

  \includegraphics[width=0.77\linewidth, trim=45 84 64 83, clip]
  {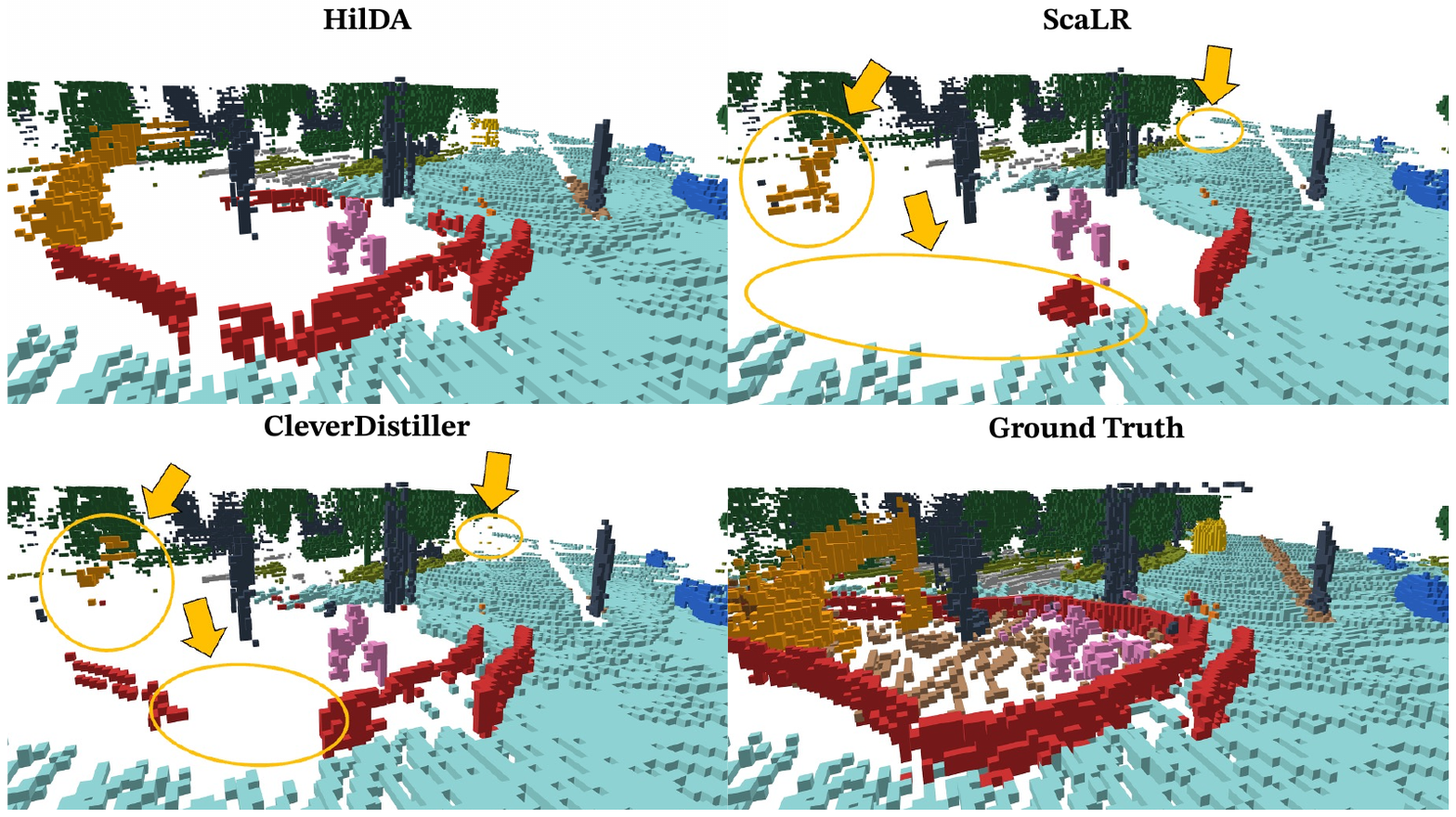}

  \caption{Semantic occupancy prediction. Illustrating higher \textit{misclassification} error of various semantic classes from baselines compared with \modelname{}. Figure shows correctly predicted occupied voxels with correct class label. Class labels colored according to Tab.~\ref{tab:4docc_sem_temp}.}
  \label{fig:sem_occ_qual_1}
\end{figure}

\Cref{fig:sem_occ_qual_1} qualitatively supports the quantitative gains in Table~\ref{tab:4docc_sem_temp}. Using \modelname{} as a frozen backbone yields cleaner semantic occupancy. In the shown scenes, the baseline backbones \emph{misclassifies} various semantic classes (\eg, excavator/construction vehicle, cars, buses, and pedestrians) to a higher degree than \modelname{}, illustrating improved class discrimination and completion.

\subsection{Scene Flow}

\begin{figure}[!t]
  \centering 
  \includegraphics[width=0.8\linewidth, trim=15 350 910 20, clip]{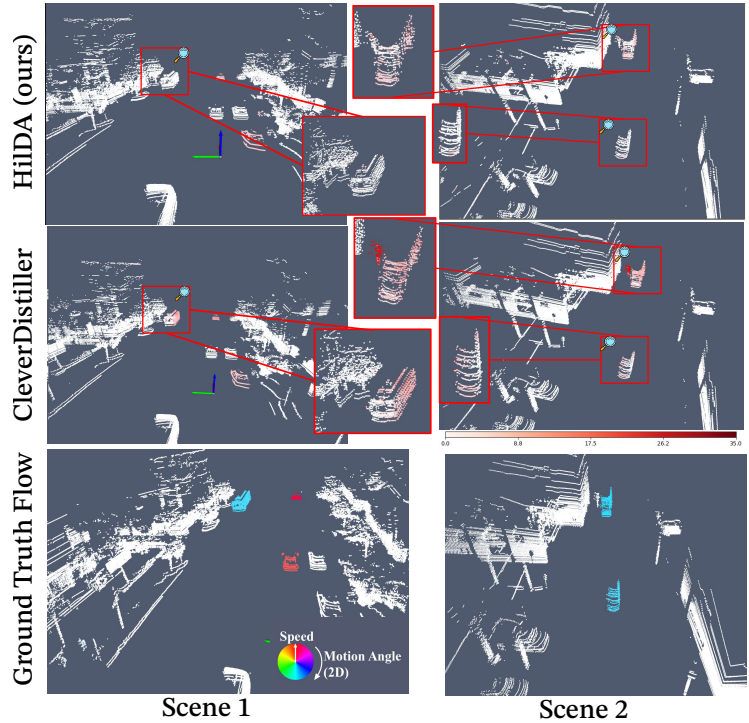}
   \caption{\textbf{Qualitative scene flow comparison.} Scene flow predictions for two example scenes. From top to bottom: \modelname{}, CleverDistiller, and ground-truth flow. For the top two rows (\modelname{} and CleverDistiller), color denotes motion error, with redder regions indicating larger errors. Ground-truth motion is visualized with a color wheel, where hue encodes direction and color intensity encodes speed. Red boxes highlight regions where \modelname{} produces more accurate and spatially consistent flow predictions.}
  \label{fig:scene_flow_suppl}
\end{figure}

\Cref{fig:scene_flow_suppl} provides an enlarged view of the qualitative LiDAR scene flow results presented in the main paper.
We compare the 3D motion predictions of \modelname{} and CleverDistiller with the ground-truth flow for two representative scenes. In the prediction rows, color denotes motion error, with redder regions indicating larger errors. In the ground-truth row, motion is visualized using a flywheel, where hue represents direction and color intensity represents speed. In both scenes, \modelname{} produces motion estimates that are closer to the ground truth, particularly in maintaining coherent flow over the truck and bus. The zoomed-in views further illustrate that CleverDistiller fails to capture the pedestrian motion, while \modelname{} recovers it and yields a clearer separation between moving objects and the static background.

\subsection{Feature Similarity}
\begin{figure}[!h]
  \centering 
  \includegraphics[width=1.0\linewidth, trim=45 84 64 83, clip]{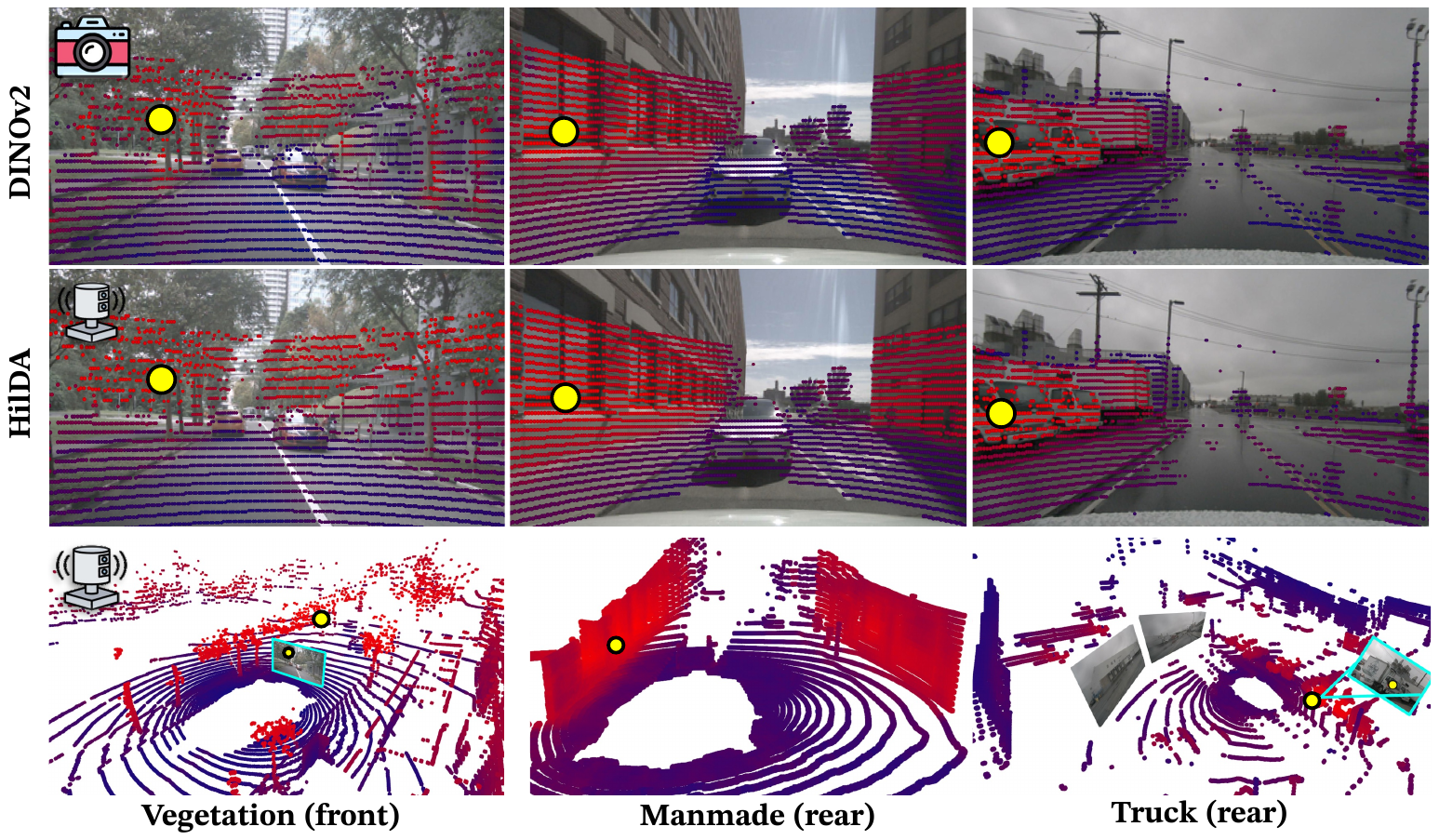}
  \caption{Feature cosine similarity. Three scenes (columns) show cosine-similarity maps for an anchor LiDAR point–pixel correspondence (\textcolor{yellow}{\textbf{yellow}} dot). The anchor pair is selected from LiDAR–image correspondences obtained by projecting the point cloud into the camera view and matching points to pixels. For \modelname{}, we compute similarity between the 3D anchor feature and all 3D point features. For DINOv2, we compute similarity between the anchor pixel feature and 2D features at point-projected pixels. The similar patterns across modalities indicate strong cross-modal alignment across environments and semantic classes. Bottom: full point cloud with the same 3D anchor. Best viewed zoomed \faSearch.}
  
  \label{fig:feat_sim_3scenes}
\end{figure}

\begin{figure}[!t]
  \centering 
  \includegraphics[width=1.0\linewidth, trim=45 84 64 83, clip]{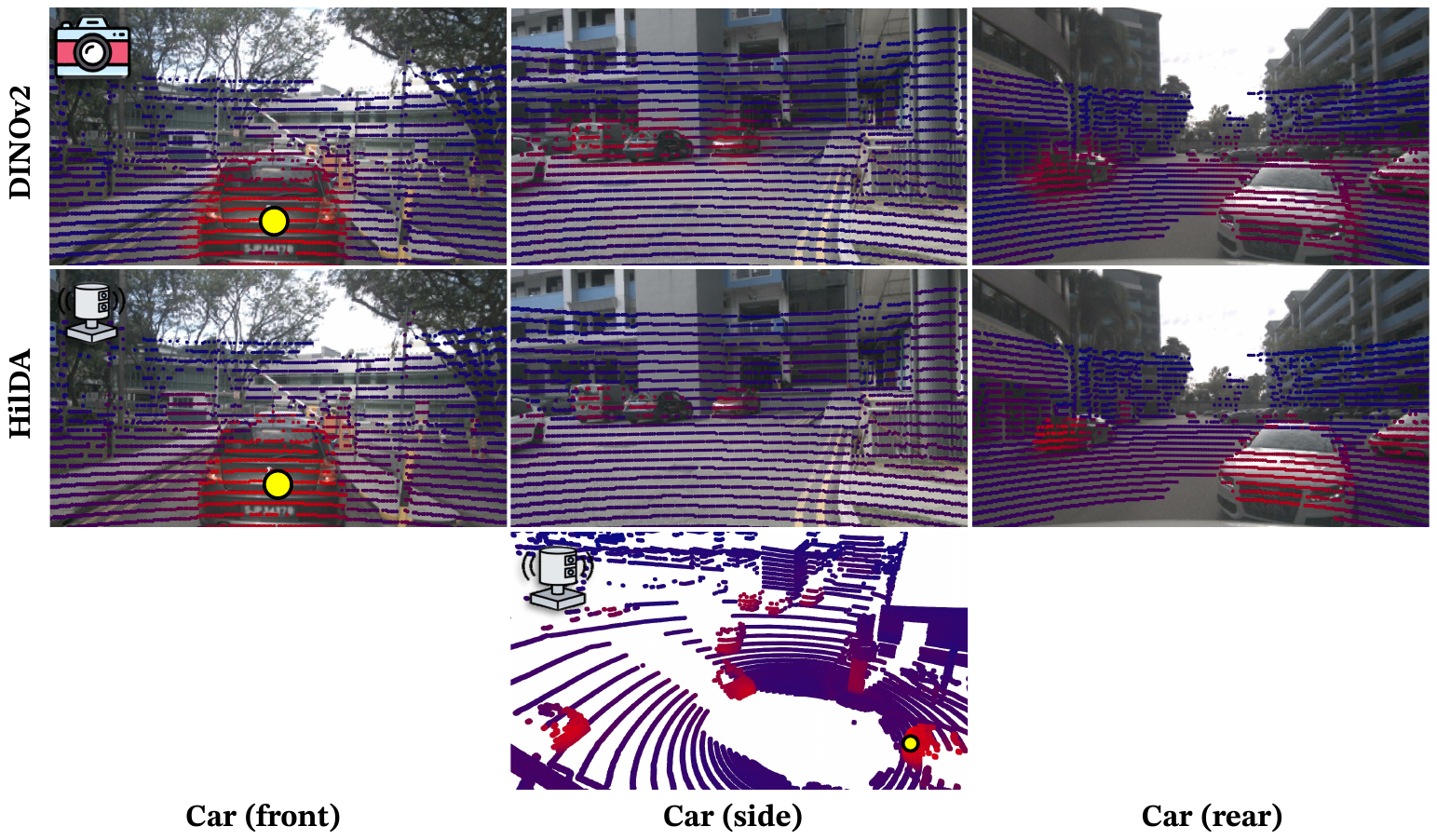}
  \caption{Cross-view feature cosine similarity (car anchor). Within a single scene, we compare a single 2D anchor (\textcolor{yellow}{\textbf{yellow}} dot) pixel feature against DINOv2 features at the point-projected pixels across \emph{all} surround-view cameras. Similarly, we compare the corresponding 3D anchor (same dot) against all points in the LiDAR sweep. Consistent high similarity on other cars across views indicates category-level cross-modal and cross-view alignment. Best viewed zoomed \faSearch.}
  \label{fig:feat_sim_car}
\end{figure}

\begin{figure}[!t]
  \centering 
  \includegraphics[width=1.0\linewidth, trim=45 84 64 83, clip]{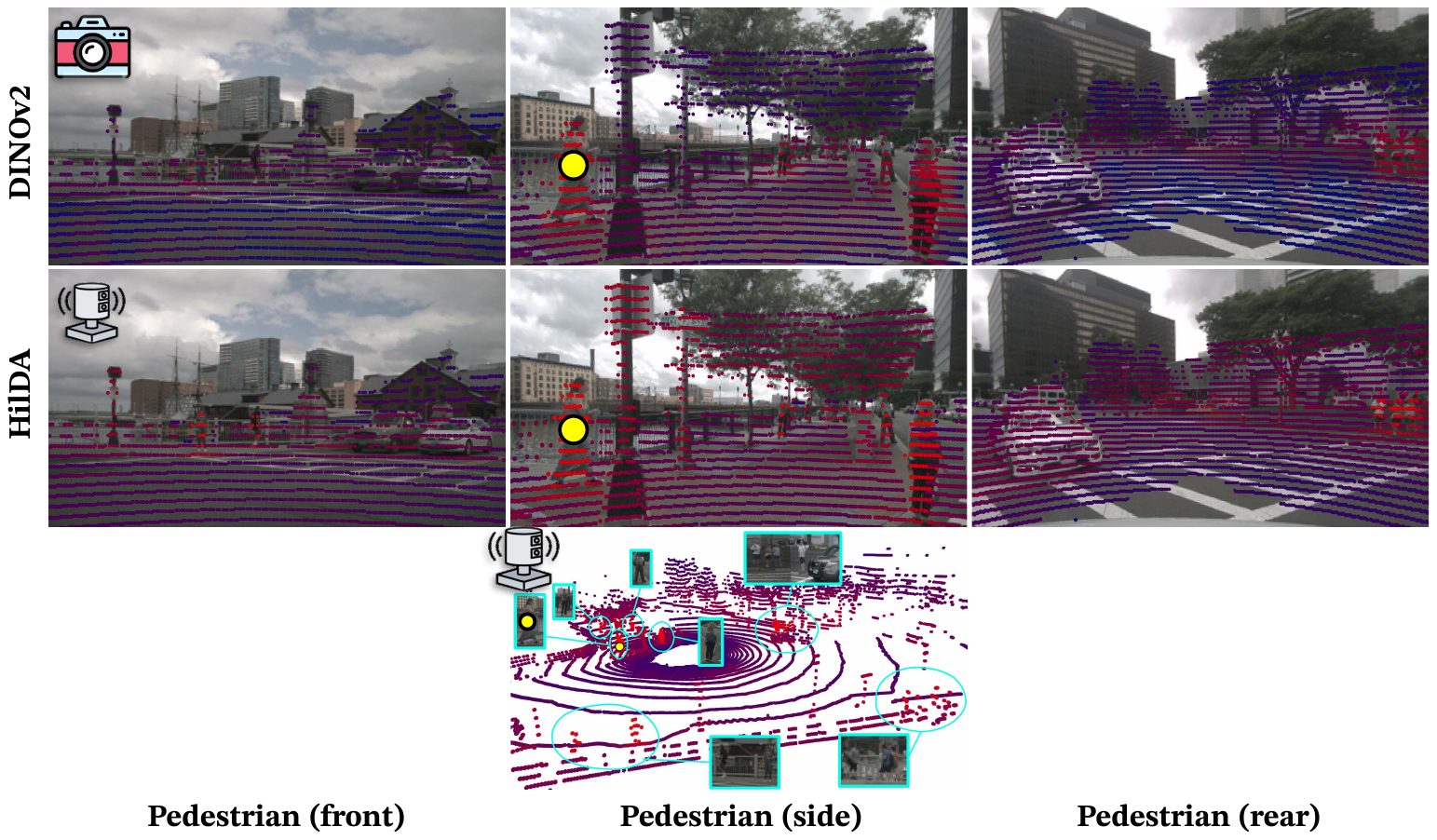}
  \caption{Cross-view feature cosine similarity (pedestrian anchor). Same setup as for~\cref{fig:feat_sim_car}, but with the anchor placed on a pedestrian (\textcolor{yellow}{\textbf{yellow}} dot) for a different scene. \modelname{} yields sharper pedestrian class boundaries and stronger similarity on more pedestrian points than DINOv2. Best viewed zoomed \faSearch.}
  \label{fig:feat_sim_ped}
\end{figure}

\begin{figure}[!t]
  \centering 
  \includegraphics[width=1.0\linewidth, trim=45 255 45 85, clip]{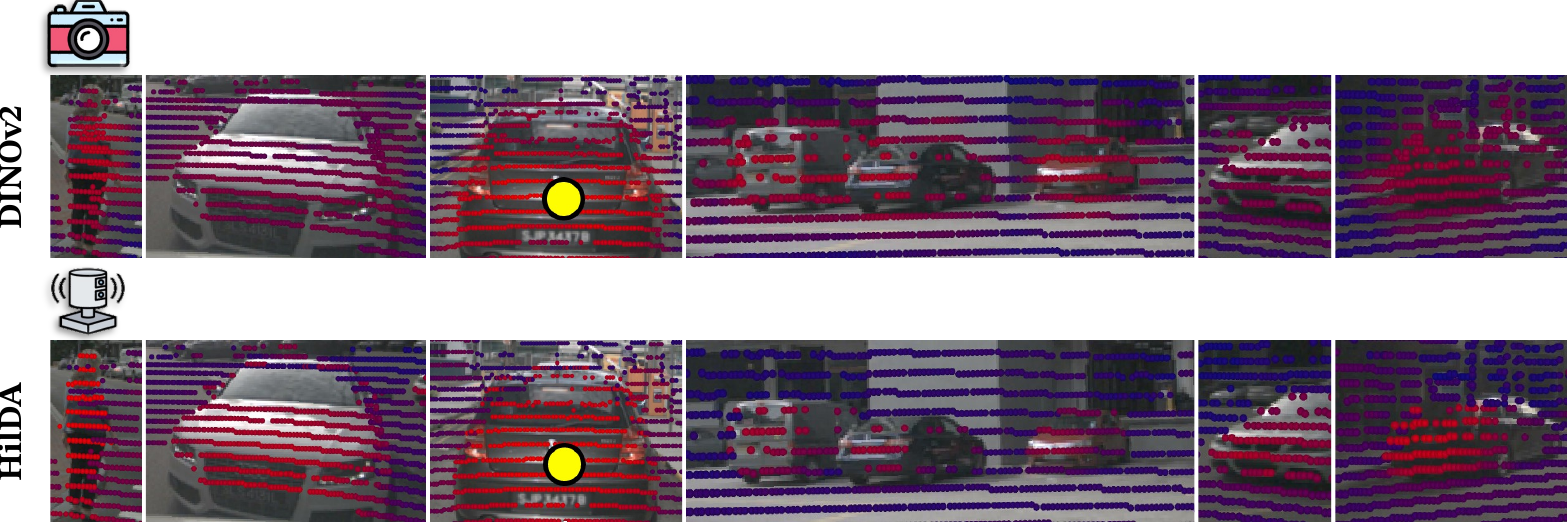}
  \caption{Feature cosine similarity close-up. Close-up from various scenes showing \modelname{}'s superior ability to respect semantic object boundaries compared to DINOv2. In this figure, the displayed anchor point is shown for reference only and is not important to the point being made. Best viewed zoomed \faSearch.}
  \label{fig:feat_sim_closeup}
\end{figure}

We qualitatively assess cross-modal alignment using cosine-similarity maps. For each scene, we choose an anchor pair (a LiDAR point and its corresponding image pixel). We compare the anchor-point feature from \modelname{} to the features of all 3D points in the scene produced by \modelname{}, and the anchor-pixel feature from DINOv2 to the DINOv2 features at point-projected pixels in 2D. That is, similarity is visualized on this sparse set rather than the full $H\times W$ grid to emphasize point-aligned evaluation. As shown in Fig.~\ref{fig:feat_sim_3scenes}, the produced similarity neighborhoods are highly consistent across modalities, indicating that \modelname{} preserves DINOv2-like semantic arrangement while operating in 3D. We also observe that DINOv2 shows higher-contrast similarity maps, while \modelname{} keeps the background more moderate.

We further probe \emph{cross-view} consistency by comparing a single 2D anchor pixel feature against DINOv2 features at the point-projected pixels across \emph{all} surround-view cameras. Similarly, we compare the corresponding 3D anchor against all points in the LiDAR sweep. In Fig.~\ref{fig:feat_sim_car} (car anchor) and Fig.~\ref{fig:feat_sim_ped} (pedestrian anchor), both methods exhibit strong intra-class responses across views, while \modelname{} is slightly stronger and more localized, producing sharper object extents.

Finally, Fig.~\ref{fig:feat_sim_closeup} highlights a possible artifact of patch/token-based 2D features. Because DINOv2 features are defined on a coarse token grid and visualized in pixel space via bi-linear up-sampling, similarity responses can \enquote{bleed} across boundaries, especially around thin structures and depth discontinuities. In contrast, we observe that \modelname{} better respects semantic and geometric boundaries, which we attribute to its voxel-based metric representation providing an explicit spatial prior. While distillation can potentially inherit these teacher-side projection artifacts, our auxiliary temporal occupancy diffusion objective may act as a structural regularizer. The denoising task emphasizes reconstructing spatial occupancy over time, encouraging sharper geometric transitions.

These observations align with the complementary strengths of discriminative and generative representations noted in \emph{A Tale of Two Features}~\cite{taleoftwofeatures}. Specifically, it reports that DINO-style discriminative features yield sparse but semantically accurate matches, whereas diffusion features often induce stronger spatial layout and coherence. By co-optimizing DINOv2 distillation with a generative temporal occupancy diffusion task, \modelname{} yields representations that remain semantically expressive while exhibiting improved geometric and boundary fidelity.

\subsection{Remarks on Annotation Noise}

\begin{figure}[!t]
  \centering 
  \includegraphics[width=1.0\linewidth, trim=40 80 100 85, clip]{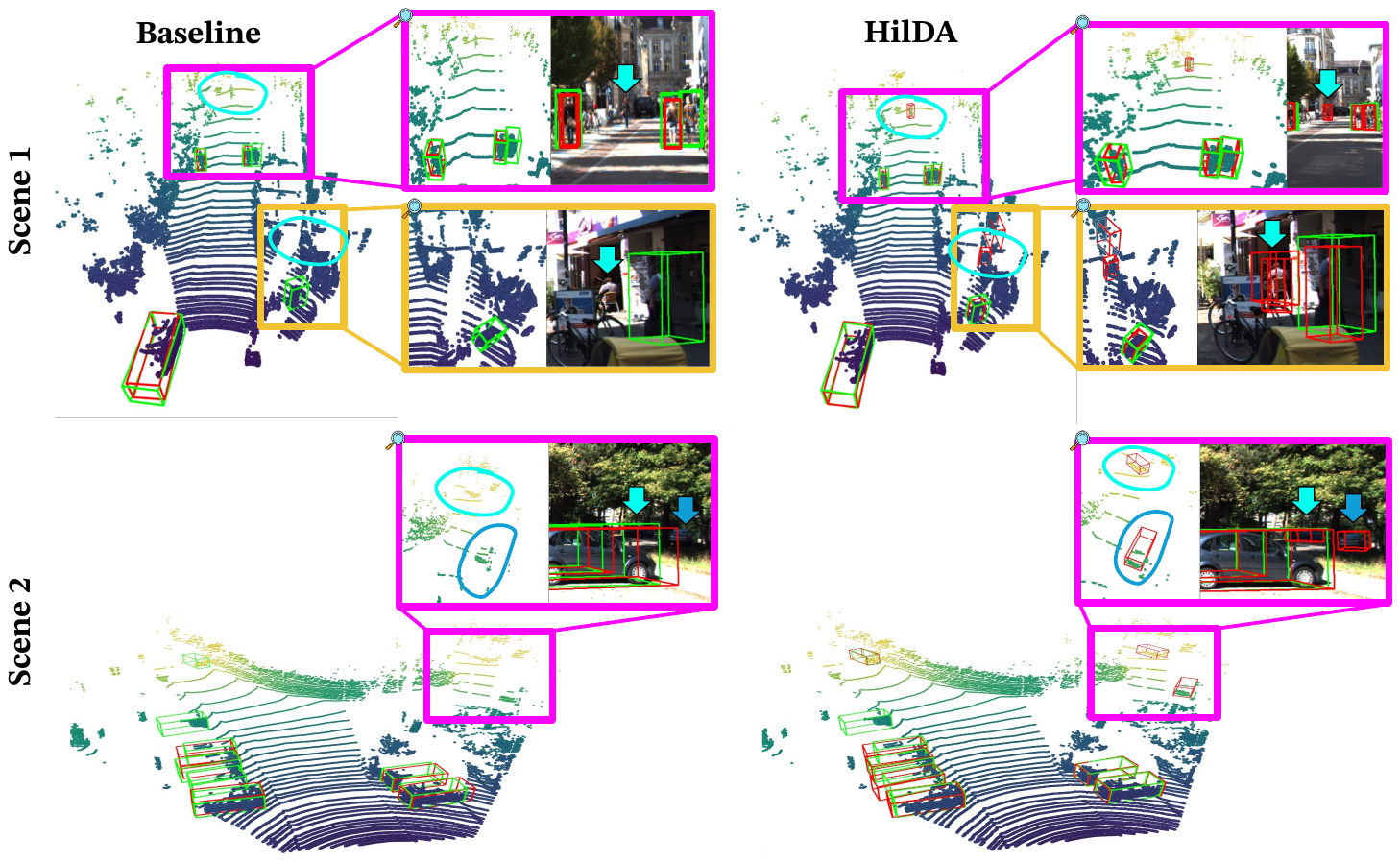}
  \caption{Annotation false negatives (3D object detection). False-negative bounding boxes are highlighted in two scenes. While CleverDistiller~\cite{govindarajan2025cleverdistiller} fails to capture these missed objects, \modelname{} recovers the missing detections, including pedestrians (Scene 1) and parked cars (Scene 2). Best viewed zoomed \faSearch.}
  \label{fig:annot_err_3}
\end{figure}

\begin{figure}[!t]
  \centering 
  \includegraphics[width=1.0\linewidth, trim=45 70 40 155, clip]{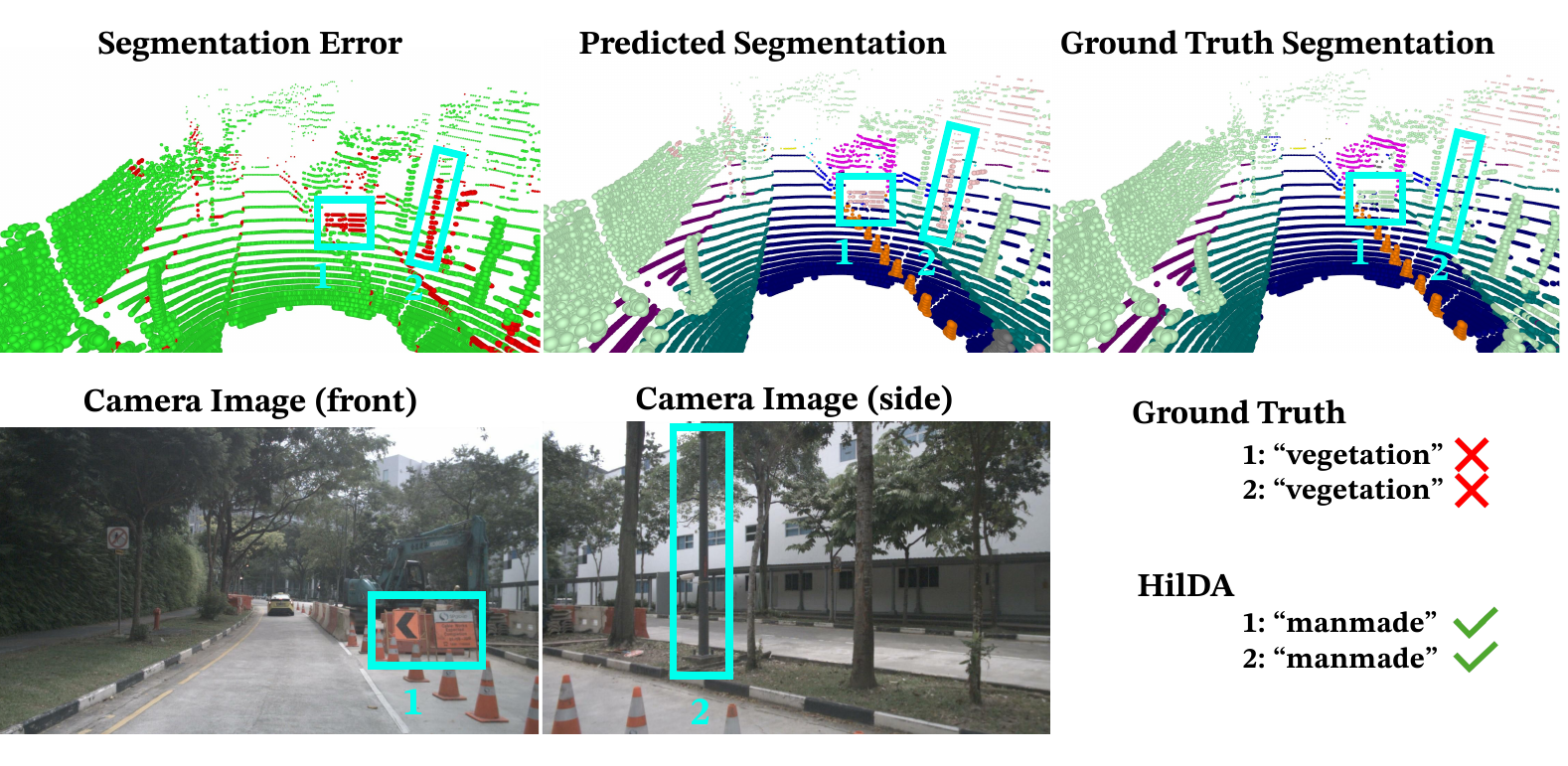}
  \caption{Annotation errors (3D semantic segmentation). We highlight a mislabeled instance where the ground truth classifies a light pole and two construction signs as \enquote{vegetation}, while \modelname{} correctly predicts them as \enquote{manmade}.}
  \label{fig:annot_err_1}
\end{figure}

\begin{figure}[!t]
  \centering 
  \includegraphics[width=1.0\linewidth, trim=45 70 45 185, clip]{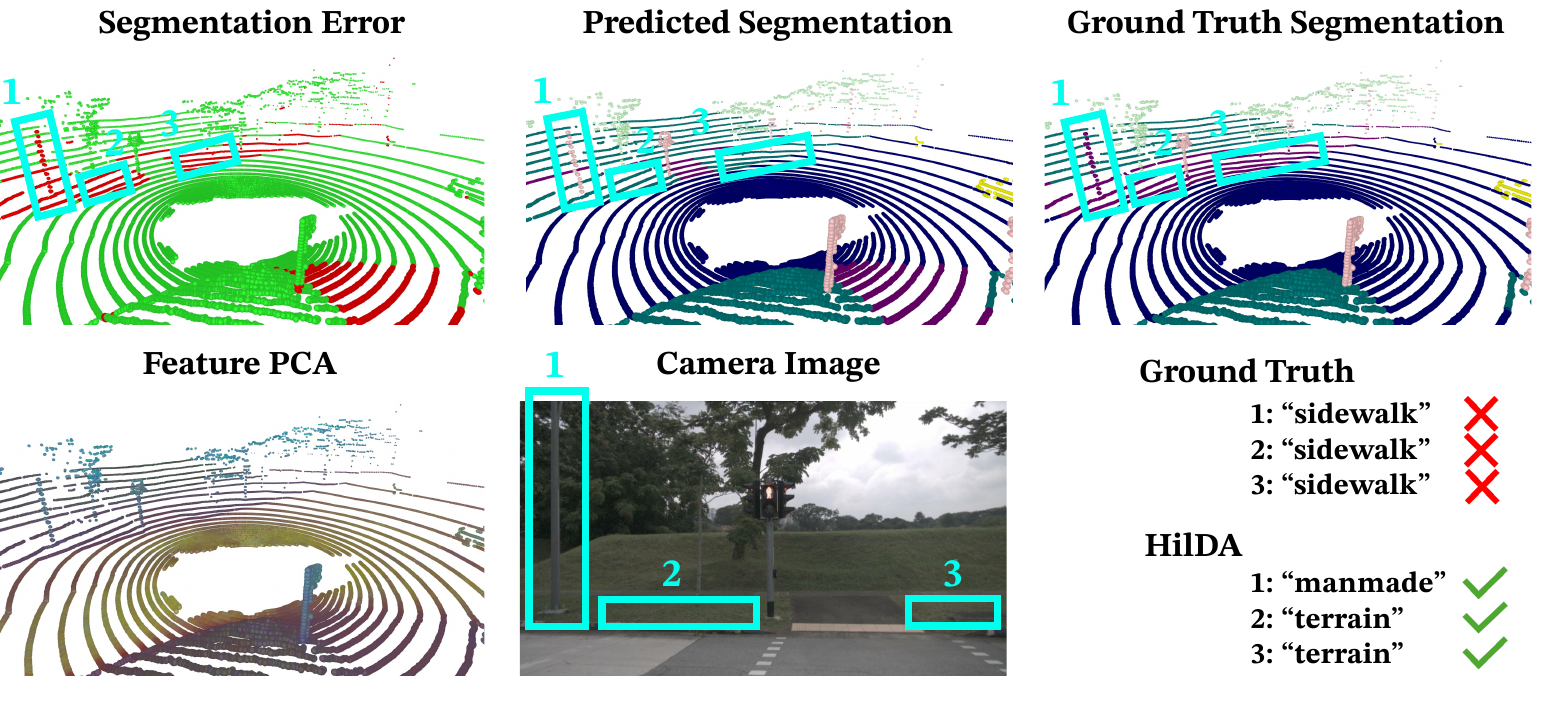}
  \caption{Annotation errors (3D semantic segmentation). We show another mislabeled example where the ground truth assigns a light pole and patches of grass to \enquote{sidewalk}, while \modelname{} correctly predicts \enquote{manmade} and \enquote{terrain}.}
  \label{fig:annot_err_2}
\end{figure}

We further observe that annotation quality can become a limiting factor once representations are sufficiently strong. In the qualitative examples (\cref{fig:annot_err_1,fig:annot_err_2,fig:annot_err_3}), \modelname{} produces predictions that disagree with the ground truth in ways that are visually attributable to missing or inconsistent annotations. We link this behavior to \modelname{}'s self-supervised (cross-modal) \pretraining. Since self-supervised learning does not optimize directly against potentially flawed human labels, it can learn label-agnostic structure from geometry and appearance, and thus be less susceptible to annotation artifacts. Compared to prior distillation baselines, \modelname{} likely benefits from the hierarchical distillation strategy (comprising multi-layer distillation and global context distillation), which improves the transfer of VFM representations into the LiDAR encoder and yields stronger, more semantically grounded 3D features.

These cases also highlight an \enquote{evaluation ceiling} induced by imperfect annotations. When test labels contain non-trivial error rates, benchmark metrics may under-estimate true progress and conflate genuine model mistakes with annotation artifacts. This effect is evident, for example, in metrics such as Average Precision (AP), where correct detections may be counted as false positives when the corresponding ground-truth instances are missing or mislabeled. Such issues have been documented~\cite{northcutt2021labelerrors,chan2024inconvenienttruth,JustoMiro_2026_WACV,zhang2025himo,Khoche2024annot,Llerena2022annot} in common benchmarks, where pervasive test-set label errors can destabilize model comparisons~\cite{northcutt2021labelerrors}. Consequently, as performance improves, qualitative inspection becomes increasingly important for separating model failure modes from limitations in annotation.

\section{Implementation Details}
\label{sec:implementationdetails}

In this section, we provide additional implementation details, including the \pretraining architecture, training protocol, and key hyperparameter settings used in our experiments. If anything is unclear, please refer to our codebase (link to be provided upon publication), open a GitHub issue, or contact the authors directly (contact information anonymized).

\subsection{Pre-Training Architecture}

Our framework is a multi-task learning system designed to learn robust 3D representations through cross-modal distillation and generative occupancy forecasting. The architecture consists of three primary components: a sparse 3D LiDAR encoder (\minkunet{}), a Vision Foundation Model (DINOv2) for distillation, and a BEV diffusion UNet head for future prediction. 

\paragraph{\textbf{LiDAR Encoder.}}

While our method is compatible with various 3D backbones, we select \minkunet{} to align with benchmarks established by previous methods. Following previous methods~\cite{choy20194minkunet, pointcept2023}, our \minkunet{} backbone is designed as a sparse ResNet-based~\cite{resnet} U-Net~\cite{ronneberger2015unet}. The model takes a voxelized point cloud as input with a grid size of $0.05\text{m}$, using relative coordinates and intensity as input features. The network follows a symmetric encoder--decoder design with skip connections, comprising four down-sampling and four up-sampling stages (called planes in many implementations). The channel dimensions for the down-sampling stages are $C=(32,64,128,256)$, with the up-sampling stages mirroring these dimensions to yield a final feature map of $96$ channels. The network depth is governed by $L_{enc}=(2,3,4,6)$ blocks for the encoder stages and $L_{dec}=(2,2,2,2)$ for the decoder, utilizing Batch Normalization with momentum $0.1$.


\paragraph{\textbf{Cross-Modal Distillation.}}

To inject 2D semantic inductive bias into the 3D encoder, we distill features from a frozen vision teacher. We use a \pretrained DINOv2~\cite{dinov2} model that processes multi-view camera images to produce dense feature maps and global \texttt{CLS} tokens. For 2D--3D alignment, we project the $96$-channel \minkunet{} features into the teacher embedding space ($384$-D) using a 3-layer MLP projection head $\mathcal{H}_\ell$ with hidden dimension $2048$, GELU activations, and LayerNorm. Distillation is performed across three temporal LiDAR sweeps by optimizing a point-wise alignment loss that maximizes cosine similarity between projected 3D points and their corresponding image-plane features, together with a global context \texttt{CLS}-token distillation loss.


\paragraph{\textbf{Spatiotemporal Self-Supervised Learning as an Auxiliary Task (Extended).}}
\label{sec:diffusion_long}

While distillation provides semantic context, it does not explicitly train the model on scene dynamics, occlusion handling, or spatial structure. To address this gap, we introduce a generative auxiliary task consisting of predicting future Bird's-Eye-View (BEV) occupancy using diffusion to steer the 3D student $S_{\theta}$ toward learning robust, predictive spatiotemporal features~\cite{hudson2023soda,diffae2022}.

Concretely, let $S_{\theta}(\cdot)$ sequentially encode two LiDAR sweeps $\mathcal{P}^{t_{-1}}$ (\emph{past}) and $\mathcal{P}^{t_0}$ (\emph{present}) into sparse features $\mathbf{F}^{t_{-1}}_L$ and $\mathbf{F}^{t_{0}}_L$ at the final student output layer. To condition the diffusion model, we must extract dense spatiotemporal cues from the sparse 3D backbone. We collapse the vertical dimension of both $\mathbf{F}^{t_{-1}}_L$ and $\mathbf{F}^{t_0}_L$ via average pooling, followed by zero-padding, to form dense BEV representations. To explicitly capture motion dynamics, these multi-frame BEV maps are channel-wise concatenated and projected through a point-wise convolution block. The resulting tensor $\mathbf{C}_\mathrm{history}$ serves as the conditioning for the diffusion process.

Guided by this condition, the diffusion model denoises the future BEV occupancy at $t_1$. To supervise this process, we utilize the future LiDAR sweep $\mathcal{P}^{t_1}$ to construct the target ground truth $\mathbf{x}_{\mathrm{occ}} \in \{0, 1\}^{H_{\mathrm{BEV}} \times W_{\mathrm{BEV}}}$. We first transform $\mathcal{P}^{t_1}$ into the current coordinate frame corresponding to $t_0$, effectively stabilizing the static background while retaining object motion. After removing the ground plane, we project the remaining points onto the BEV grid. A cell is marked as occupied ($1$) if any point from $\mathcal{P}^{t_1}$ falls within it, and zero otherwise.

We formulate future occupancy prediction as a conditional Denoising Diffusion Probabilistic Model (DDPM)~\cite{ddpm}. In the forward process, the ground-truth future occupancy map $\mathbf{x}_{\mathrm{occ}}$ is progressively corrupted with Gaussian noise $\epsilon \sim \mathcal{N}(0, \mathbf{I})$ over $T$ diffusion steps according to a predefined variance schedule $\{\beta_\tau\}_{\tau=1}^T$: $\mathbf{x}_\tau = \sqrt{\bar{\alpha}_\tau}\,\mathbf{x}_{\mathrm{occ}} + \sqrt{1 - \bar{\alpha}_\tau}\,\epsilon$, where $\bar{\alpha}_\tau = \prod_{s=1}^{\tau} (1 - \beta_s)$. The reverse denoising process, which predicts the added noise, is parameterized by a neural network $\epsilon_\theta(\mathbf{x}_\tau, \tau, \mathbf{C}_{\mathrm{history}})$, as shown in~\cref{fig:architecture}. It is implemented as a 2D UNet~\cite{ronneberger2015unet}, which predicts the injected noise $\epsilon$ given the noisy future occupancy $\mathbf{x}_\tau$, conditioned on the historical BEV features $\mathbf{C}_{\mathrm{history}}$ and diffusion timestep $\tau$. Conditioning is implemented via channel-wise concatenation of $\mathbf{x}_\tau$ and $\mathbf{C}_{\mathrm{history}}$, following~\cite{Li_2025_CVPR}. The reconstructed occupancy $\hat{\mathbf{x}}_{\mathrm{occ}}$ is recovered from the noisy sample by inverting the forward process using the predicted noise $\hat{\mathbf{x}}_{\mathrm{occ}} = (\mathbf{x}_\tau - \sqrt{1 - \bar{\alpha}_\tau}\,\epsilon_\theta(\mathbf{x}_\tau, \tau, \mathbf{C}_{\mathrm{history}}))/{\sqrt{\bar{\alpha}_\tau}}$. Training across varying $\tau$ exposes the model to denoising tasks at different noise levels, encouraging a coarse-to-fine behavior from global structure to local detail. We optimize the hybrid objective given in~\cref{eq:diffusion_loss}.

\paragraph{\textbf{BEV Diffusion Auxiliary Head.}}

We introduce a future BEV occupancy diffusion auxiliary head. Sparse 3D features from the \minkunet{} are projected onto a $512\times512$ BEV grid with an effective stride of $8$ relative to the voxel grid by mean-pooling features within each BEV pillar using \texttt{torch\_scatter}. The diffusion model is conditioned on fused BEV context from the current sweep ($t_0$) and the previous sweep ($t_{-1}$), which are concatenated and compressed with a convolutional grouping block ($192\!\to\!32$ channels) using GroupNorm and SiLU activations. The denoising network is a lightweight 2D Unet that takes the noisy future state (targeting occupancy at $t_1$) together with the history-conditioned context $C_{\mathrm{history}}$, and uses time embeddings ($\texttt{dim}=16$) and ResBlocks. We adopt a cosine $\beta$-schedule with $T=1000$ diffusion steps. Training uses a composite objective consisting of the standard diffusion noise prediction MSE and a reconstruction $\ell$-2 norm loss on the denoised $\hat{\mathbf{x}}_{\mathrm{occ}}$ estimate. The reconstruction loss term is weighted by $\lambda=0.5$.

\paragraph{\textbf{Training.}}

We implement our framework using PyTorch and conduct training on a cluster of 8 NVIDIA A100 GPUs. The model is trained for 100 epochs with a global batch size of 32 (4 samples per GPU), utilizing Automatic Mixed Precision (AMP). We optimize the network using AdamW~\cite{loshchilov2017decoupled} with a weight decay of $0.005$. Learning rate scheduling is managed via the OneCycle policy with a peak learning rate of $1 \times 10^{-3}$. This schedule comprises a linear warmup for the initial $4\%$ of iterations (starting from $\frac{\text{lr}_{max}}{10}$) followed by cosine annealing that decays to $\frac{\text{lr}_{max}}{1000}$. Input point clouds are voxelized with a grid size of $0.05\text{m}$.

To mitigate overfitting, we apply standard augmentations: (1) random rotation around the $z$-axis in the range $\theta \in [-1, 1]$ radians; (2) random uniform scaling $s \in [0.9, 1.1]$; (3) random horizontal flipping ($p=0.5$); and (4) Gaussian point jittering ($\sigma=0.005$) clipped at $0.02$.


\subsection{3D Semantic Segmentation}

For the downstream semantic segmentation experiments, we utilize a specialized decoding module to evaluate the quality of the learned representations. The architecture consists of the \pretrained LiDAR backbone followed by a lightweight segmentation head. As shown in the implementation, this head is composed of a single 1D Batch Normalization layer followed by a linear projection layer that maps the feature dimension $C^L$ to the number of semantic classes $N_{cls}$.

In the linear probing configuration, we explicitly freeze the parameters of the \pretrained backbone to assess the expressiveness of the fixed feature representations. During training, gradients are not computed for the backbone. Optimization is restricted strictly to the parameters of the segmentation head (the batch norm and linear layers). For the fine-tuning configuration, we initialize the backbone with the \pretrained weights and allow all model parameters to be updated. The entire network (backbone and segmentation head) is trained end-to-end using the standard cross-entropy loss.

\paragraph{\textbf{Metrics.}}

We report the standard Intersection-over-Union (IoU) for individual categories and the mean IoU (mIoU) for overall performance. The IoU for category $i$ is defined as:
\begin{equation}
    \text{IoU}_i = \frac{\text{TP}_i}{\text{TP}_i + \text{FP}_i + \text{FN}_i},
\end{equation}
where $\text{TP}_i$, $\text{FP}_i$, and $\text{FN}_i$ denote true positives, false positives, and false negatives, respectively.

For the robustness analysis on Robo3D~\cite{kong2023robo3d}, we adopt the Corruption Error (CE) and Resilience Rate (RR) metrics, defined as:
\begin{equation}
    \text{CE}_i = \frac{\sum_{j=1}^{3}(1 - \text{mIoU}_{i,j})}{\sum_{j=1}^{3}(1 - \text{mIoU}_{i,j}^{\text{base}})}, \quad 
    \text{RR}_i = \frac{\sum_{j=1}^{3} \text{mIoU}_{i,j}}{3 \times \text{mIoU}_{\text{clean}}},
\end{equation}
where $\text{mIoU}_{i,j}$ represents the score for scenario $i$ at severity level $j$. $\text{mIoU}_{i,j}^{\text{base}}$ refers to the baseline model's performance, and $\text{mIoU}_{\text{clean}}$ denotes performance on the clean validation set. Consistent with prior work, all evaluations are performed without test-time augmentation or ensembling.

\subsection{3D Object Detection}


The model architecture is implemented within the open-source MMDetection3D~\cite{mmdet3d2020} framework and follows a two-stage 3D object detection paradigm based on PointRCNN. For point cloud feature extraction, the standard PointNet++ backbone is replaced with a \minkunet{} module. The input point clouds, possessing a dimensionality of $4$ ($x, y, z, \text{intensity}$), are processed to yield dense point-wise features in $\mathbb{R}^{96}$. Notably, we use \minkunet{} weights pretrained with ours and baseline methods.

\paragraph{\textbf{Detection Model.}}

In the first stage, the Region Proposal Network maps the 128-dimensional point features through linear layers of dimension $(256, 256)$ to generate initial 3D bounding box proposals. The bounding boxes are encoded as 8-dimensional vectors (center residual $\in \mathbb{R}^3$, size regression $\in \mathbb{R}^3$, and orientation $[\cos(\theta), \sin(\theta)] \in \mathbb{R}^2$). This stage is optimized via Focal Loss for foreground/background classification across three categories (Pedestrian, Cyclist, Car) and Smooth L1 Loss for bounding box regression.

In the second stage, a Point Head refines the initial proposals. It pools features from 512 sampled points per Region of Interest (RoI). The pooled points, initially represented by 5-dimensional vectors (spatial coordinates, classification score, and depth), are processed through successive Set Abstraction layers that expand the feature dimensionality to $512$. Finally, Multi-Layer Perceptrons (MLPs) with hidden channels of $(256, 256)$ perform the final classification and bounding box parameter refinement.

\paragraph{\textbf{Metrics.}}

Performance is evaluated using the standard mean Average Precision (mAP) metric. We calculate mAP across different difficulty levels (\emph{Easy}, \emph{Moderate}, \emph{Hard}) with thresholds specific to each class, following standard benchmark protocols~\cite{kitti}.

\subsection{Semantic Occupancy Prediction}

For this task, we inherit the data pre-processing pipeline of Occ4cast~\cite{Liu2023occ4cast} to produce ground truth semantic occupancy in 3D and 4D using the nuScenes \cite{caesar2020nuscenes} lidarseg dataset. 

\paragraph{\textbf{Semantic Occupancy Decoder.}}

All semantic occupancy experiments are run with a \pretrained and frozen sparse 3D backbone and a trainable, lightweight occupancy decoder head, which comprises $\sim1\,\%$ of the total model parameters. The backbone takes point cloud data carrying spatial position $(x,y,z)$ and intensity as input, voxelized at $0.05\,\mathrm{m}$, and produces sparse 3D voxel features with a channel dimension of $96$. For decoding, sparse features are pooled to the $0.2\,\mathrm{m}$ output grid cells and projected to a BEV feature map via scatter-mean aggregation over $(x,y)$ bins. The BEV features are densified into a 2D grid by zero-filling empty cells, and projected to a $64$-channel latent space. We then apply two stride-2 convolutional blocks to down-sample the BEV map, followed by a single transformer encoder block layer for global aggregation (one self-attention head with dimension $64$). We add 2D sine--cosine positional embeddings to the BEV tokens before attention. The attention layer provides global context over the backbone features, enabling spatial propagation of information used to infer structural connectivity between sparse points, as demonstrated by \cite{diehl2025dio}. The features are then up-sampled back to the original BEV resolution using two transposed-convolution blocks. To lift the BEV representation to 3D, we use a learned $1\times1$ projection that expands the channel dimension to $64\times H$ (with $H$ being the height dimension) and reshapes the resulting tensor into a dense 3D volume. This volume is fused with a densified 3D skip branch from the sparse backbone features aligned on the same target grid. Fusion is performed by channel-wise concatenation followed by a voxel-wise convolution, and the result is mapped to semantic logits via a final per-voxel linear projection. Predictions are produced on a dense 3D grid with $0.2\,\mathrm{m}$ resolution, within $[-51.2,51.2]\times[-25.6,25.6]\times[-2.0,4.4]\,\mathrm{m}$ ($512\times 256 \times 32$ voxels).

\paragraph{\textbf{Loss Function.}}

We train the semantic occupancy decoder using a multi-class cross-entropy loss over voxel labels, which include \emph{free space} and an \emph{unlabeled occupied} category in addition to the $16$ semantic classes shown in~\cref{tab:4docc_sem_temp}. Invalid voxels are ignored, following \cite{Liu2023occ4cast}.

\paragraph{\textbf{Training.}}

To mitigate the dominance of free-space voxels, we include all voxels labeled as occupied and uniformly subsample free voxels when forming the loss. We found a $5{:}1$ free-to-occupied sampling ratio to work best and use it for all results. Importantly, we retain free-space supervision in the loss to explicitly learn the occupancy boundary and prevent semantic labels from bleeding into empty space. We train the decoder for $15$ epochs using Adam~\cite{kingma2014adam} with initial learning rate $5\times 10^{-4}$ and StepLR decay (step size $5$ epochs, decay factor $0.1$). We use a per-GPU batch size of $4$ (global batch size $16$) over $4$ NVIDIA A100 GPUs, and automatic mixed precision. The semantic occupancy prediction uses either a single input sweep at \(t_0\) or two sweeps at \((t_{-1}, t_0)\), where each sweep is encoded independently by the frozen backbone and the resulting features are channel-wise concatenated and projected into the decoder latent space. The max forecasting horizon is set to either $t_0$, $t_1$, $t_5$, or $t_{10}$ (inclusive).

\paragraph{\textbf{Metrics.}}

During evaluation, we report semantic IoU over labeled occupied semantic classes, excluding \emph{free space} and \emph{unlabeled occupied}. While \emph{unlabeled occupied} is included during training to account for uncertain occupancy, it is omitted from IoU reporting to focus on semantic performance on labeled classes. For multi-step forecasting, metrics are computed separately for each prediction horizon ($t_0,\dots,t_N$), where $N$ ranges from $1$ to $10$. For each horizon, we average class-wise IoUs to obtain a per-horizon mIoU. We then report a single summary score by averaging mIoU across horizons.

\subsection{Scene Flow}

We implement our experiments using the OpenSceneFlow~\cite{opensceneflow2026} framework. For the scene flow estimation task, we adopt the SSF method~\cite{khoche2025ssf} as our baseline. We replace the default SSF sparse backbone~\cite{khoche2025ssf} with a \minkunet{} backbone~\cite{choy20194minkunet} and use our \pretrained model weights. 

\paragraph{\textbf{Data and Voxelization.}}

We conduct training and evaluation on the ArgoverseV2~\cite{Argoverse2} dataset. The input point clouds are clipped to a spatial range of $[-51.2, 51.2]\,\text{m}$ along the $x$ and $y$ axes, and $[-3, 3]\,\text{m}$ along the $z$ axis. We employ a voxel grid size of $0.05\,\text{m}$ for fine-grained geometric processing.

\paragraph{\textbf{Training Configuration.}}

We set the initial learning rate to $2 \times 10^{-4}$. To preserve the \pretrained representations of the backbone while adapting to the task, the backbone learning rate is scaled down by a factor of $0.1$ (resulting in an effective rate of $2 \times 10^{-5}$). The network is supervised using the DeFlow~\cite{zhang2024deflow} loss function, following the configuration in~\cite{khoche2025ssf}.


\paragraph{\textbf{Metrics.}}

We evaluate scene flow using two primary metrics: three-way End Point Error (EPE)~\cite{Chodosh_2024_WACV} and Dynamic Bucket-Normalized EPE~\cite{khatri2024trackflow}. The End Point Error (EPE) is defined as the $\ell_2$ norm of the difference between the predicted flow vector $\hat{\mathbf{f}}$ and the ground truth flow vector $\mathbf{f}_{gt}$, measured in centimeters:
\begin{equation}
    \text{EPE} = \| \hat{\mathbf{f}} - \mathbf{f}_{gt} \|_2.
\end{equation}
The three-way EPE computes the unweighted average EPE over three distinct regions: Foreground Dynamic (FD), Foreground Static (FS), and Background Static (BS). The regions are classified based on two criteria: (1) a point is classified as \emph{dynamic} if its ground truth velocity satisfies $\| \mathbf{v}_{gt} \| > 0.5$ m/s, (2) a point is considered \emph{foreground} if it lies within the bounding box of any tracked object. Dynamic Bucket-Normalized EPE groups points into predefined motion buckets based on their speeds. This metric normalizes the error relative to the magnitude of motion:
\begin{equation}
    \text{Normalized EPE} = \frac{\text{Mean EPE}}{\text{Mean Speed}}.
\end{equation}


\section{Datasets}

We assess the performance and generalization capability of our method through three primary experimental setups: linear probing, data-efficient fine-tuning, and cross-domain transfer learning.

\subsection{NuScenes Experiments}

We utilize the nuScenes dataset~\cite{caesar2020nuscenes} as our primary testbed, employing standard splits of $700$ scenes for training ($29,130$ frames) and $150$ scenes for validation ($6,019$ frames). Performance is measured across $16$ semantic classes. To evaluate the quality of the frozen representations, we train only a linear head on top of the fixed backbone using the full training set. We further assess the adaptability of the \pretrained weights by fine-tuning the entire network. To simulate varying data regimes, we train on subsets containing $1\%$, $5\%$, $10\%$, $25\%$, and $100\%$ of the available annotated data.

\subsection{Cross-Domain Transfer Learning}

To verify the generalization of our \pretraining, we transfer the model to a diverse set of datasets. 

\paragraph{\textbf{Large-Scale Real-World Data.}} 

We utilize SemanticKITTI~\cite{behley2019semantickitti} and Waymo Open~\cite{sun2020waymo}, both captured with $64$-beam LiDARs. For data-efficient transfer, we construct $1\%$ subsets by sampling every $100^{\text{th}}$ frame. 

\paragraph{\textbf{Weak and Sparse Annotations.}} 

We employ ScribbleKITTI~\cite{unal2022scribble}, which shares scans with SemanticKITTI but uses line-scribble annotations (covering only $8.06\%$ of points) to test performance under weak supervision. 

\paragraph{\textbf{Specialized Environments.}} 

We include RELLIS-3D~\cite{jiang2021rellis}, an off-road multi-modal dataset which is challenging for its class imbalance and unstructured topography.

\paragraph{\textbf{Adverse Conditions.}} 

Robustness to weather is tested on SemanticSTF~\cite{xiao20233d}, which contains $2,076$ scans spanning rainy, snowy, and foggy conditions.

\paragraph{\textbf{Synthetic Data.}} 

We further evaluate on simulated environments using the large-scale semi-synthetic DAPS-3D~\cite{klokov2023daps3d} (DAPS-1 subset).


\subsection{Robustness and Adverse Conditions}

We utilize the nuScenes-C dataset from the Robo3D benchmark~\cite{kong2023robo3d} to evaluate robustness against sensor corruption. This benchmark introduces eight corruption types (\eg, motion blur, crosstalk, beam missing) across three severity levels (light, moderate, heavy). Models are evaluated on all levels, and we report the average performance.

\section{Limitations and Future Work}

We summarize the main limitations of \modelname{} and identify several avenues for extending the proposed framework.

\paragraph{\textbf{Calibration.}}



Although our method is robust to common data corruptions (as seen in~\cref{tab:robust_full}), distillation is performed on well-calibrated data and relies on calibration parameters to project points onto image pixels. Small misalignments are often tolerated due to the ViT~\cite{vit} patch-based representation, whereas larger offsets can lead to incorrect point–pixel correspondences and degraded performance. In the full \pretraining setup of \modelname{}, global context distillation and temporal diffusion, neither of which depends on strict pairwise alignment, may help reduce this sensitivity. We further reduce sensitivity to small 2D–3D random misalignments during distillation using point-cloud jittering at \pretraining.

\paragraph{\textbf{Multi-Layer Distillation.}}

We report extensive experiments with a DINOv2~\cite{dinov2} teacher and two LiDAR backbones, \minkunet{}\cite{choy20194minkunet} and PTv3\cite{ptv3}. \minkunet{} is widely used in prior cross-modal distillation benchmarks, which facilitates comparison to existing work. Our results with PTv3 indicate that the benefits of \modelname{} transfer to a substantially different LiDAR architecture, and that the optimal multi-layer distillation depth can be backbone-dependent. Beyond the fixed layer-matching strategy used here, an interesting direction is to optimize teacher–student layer pairing via matching, learned correspondences, or dynamic loss weighting. Moreover, we only evaluate how the proposed multi-layer distillation performs for a single \emph{teacher} family (varying sizes of DINOv2). Since different vision foundation models produce features with distinct inductive biases and semantic content, both the optimal layer correspondences and resulting distilled LiDAR representations may be teacher-dependent. Characterizing this dependence, and extending the framework to distill from multiple complementary teachers, are promising directions for future work within the cross-modal distillation field.

\paragraph{\textbf{Auxiliary Diffusion Objective.}}

Finally, the current diffusion target operates in BEV, which offers a compact and efficient representation but inevitably compresses 3D structure into a 2D plane. Extending the diffusion target to more expressive 3D-aware representations (\eg, tri-plane, or other volumetric or implicit formulations) could better preserve fine-grained geometry and improve modeling of temporal evolution. A complementary direction is to move the diffusion process to a latent space (\eg, latent diffusion~\cite{rombach2022high}), which may retain richer scene information while avoiding an explicit 3D-to-2D bottleneck. We do not pursue this here because it would introduce an additional \pretrained variational auto-encoder pathway and associated distillation, potentially confounding the analysis of what is learned from the diffusion loss itself. Nonetheless, latent (and more generally fully 3D-aware) diffusion targets remain a promising avenue for future work.

\section{Hyperparameter Settings}
\label{sec:hyperparam}

For implementation, we reuse the Pointcept~\cite{pointcept2023} framework. We use the standard training configuration and \minkunet{}~\cite{choy20194minkunet} from the available configurations in the repository.~\Cref{tab:training_hyperparams,tab:model_hyperparams,tab:data_hyperparams} summarize the hyperparameter settings used in our experiments.

\begin{table}[h]
\centering
\caption{Hyperparameters for training and optimization settings.}
\label{tab:training_hyperparams}
\begin{tabular}{@{}ll@{}}
\toprule
\textbf{Parameter} & \textbf{Value} \\
\midrule
Epochs & 100 \\
Batch Size & 32 (Total across all GPUs) \\
Optimizer & AdamW \\
Learning Rate (lr) & 0.001 \\
Weight Decay & 0.005 \\
Scheduler & OneCycleLR \\
Max LR & 0.001 \\
Pct Start & 0.04 \\
Anneal Strategy & Cosine (\texttt{cos}) \\
Div Factor / Final Div Factor & 10.0 / 100.0 \\
Mixed Precision (AMP) & Enabled \\
\bottomrule
\end{tabular}
\end{table}

\begin{table}[h]
\centering
\caption{Model Architecture configurations for the 3D student (\minkunet{} backbone).}
\label{tab:model_hyperparams}
\begin{tabular}{@{}ll@{}}
\toprule
\textbf{Parameter} & \textbf{Value} \\
\midrule
Backbone Out Channels & 96 \\
Input Channels & 4 \\
Base Channels & 32 \\
Planes & (32, 64, 128, 256, 256, 128, 96, 96) \\
Layer (per plane) & (2, 3, 4, 6, 2, 2, 2, 2) \\
BatchNorm Momentum & 0.1 \\
3D Head Type (all) & MLP (3 layers, In: 96, Out: 384, Norm: True) \\
\bottomrule
\end{tabular}
\end{table}

\begin{table}[h]
\centering
\caption{Data augmentation and processing parameters for distillation.}
\label{tab:data_hyperparams}
\begin{tabular}{@{}ll@{}}
\toprule
\textbf{Parameter} & \textbf{Value} \\
\midrule
Grid Size (Train) & 0.05 \\
Random Rotate (z-axis) & Angle: $[-1, 1]$ radians, $p=0.5$ \\
Random Scale & $[0.9, 1.1]$ \\
Random Flip & $p=0.5$ \\
Random Jitter & $\sigma = 0.005$, Clip: $0.02$ \\
\bottomrule
\end{tabular}
\end{table}

\end{document}